\documentclass{article} 
\usepackage{iclr2027_conference,times}

\usepackage{amsmath,amsfonts,bm}

\def\eqref#1{equation~\ref{#1}}

\def\1{\bm{1}}

\DeclareMathAlphabet{\mathsfit}{\encodingdefault}{\sfdefault}{m}{sl}
\SetMathAlphabet{\mathsfit}{bold}{\encodingdefault}{\sfdefault}{bx}{n}

\usepackage{hyperref}
\usepackage{url}
\usepackage[utf8]{inputenc} 
\usepackage[T1]{fontenc}    
\usepackage{booktabs}       
\usepackage{amsfonts}       
\usepackage{nicefrac}       
\usepackage{microtype}      
\usepackage{xcolor}         
\usepackage{amsmath}
\usepackage{algorithm}
\usepackage[noend]{algpseudocode}
\usepackage{amssymb}
\usepackage{graphicx}
\usepackage{booktabs}
\usepackage{makecell}
\usepackage{adjustbox}
\usepackage{amsmath, amssymb}
\usepackage{multirow}
\usepackage{adjustbox}
\usepackage{tabularx}
\usepackage{enumitem}
\usepackage{mdframed}
\usepackage[table]{xcolor}
\usepackage[most]{tcolorbox}
\usepackage{minted} 
\usepackage{subcaption}
\usepackage{arydshln}

\definecolor{figgreen}{RGB}{59, 125, 35}
\definecolor{figorange}{RGB}{215, 137, 0}
\definecolor{figred}{RGB}{192, 0, 0}
\definecolor{figblue}{RGB}{48, 132, 194}
\definecolor{mygray}{gray}{0.6}
\definecolor{Red}{rgb}{0.8,0,0}
\definecolor{Blue}{rgb}{0,0,0.8}
\definecolor{gg}{gray}{0.92}
\definecolor{tabred}{RGB}{230, 10, 10}
\definecolor{tabblue}{RGB}{10, 10, 230}
\definecolor{citeblue}{rgb}{0.15,0.35,0.55}
\definecolor{linkred}{rgb}{0.9,0,0}
\definecolor{Blue9}{rgb}{0.1,0.3,0.95}
\definecolor{myurl}{RGB}{240, 82, 156}
\hypersetup{
    colorlinks = true,
    citecolor = teal,
    linkcolor = linkred,
    urlcolor = myurl
}

\newcolumntype{a}{>{\columncolor{gg}}c}

\title{Principia: Test-Time Hypotheses to Principle Construction for Scientific Discovery}

\author{\mbox{Dongki Kim$^{1}$\thanks{denotes equal contribution.}\hphantom{$^{*}$}}\hspace{0.6em}\mbox{Sangwoo Park$^{1}$\footnotemark[1]\hphantom{$^{*}$}}\hspace{0.6em}Jinheon Baek$^{1}$\hspace{0.6em}Soyeong Jeong$^{1}$\hspace{0.6em}Sung Ju Hwang$^{1,2}$ \\[0.25em]
$^{1}$KAIST \quad $^{2}$DeepAuto.ai \\[0.25em]
\texttt{\fontsize{8.5pt}{8.5pt}\selectfont \{cleverki, jackson0530, jinheon.baek, starsuzi, sungju.hwang\}@kaist.ac.kr}}

\iclrfinalcopy 
\begin{document}

\maketitle
\lhead{Preprint}
\begin{abstract}
Scientific discovery requires identifying the scientific principles underlying observed phenomena to guide further investigation. As these empirical principles are often unknown in complex discovery processes, they are constructed via a hypotheses-to-principle pipeline to obtain recurring causalities from progressively accumulated observations during test-time discovery. However, constructing this knowledge requires substantial domain expertise, time, and effort, making this translation stage a bottleneck for scientific discovery. To address this limitation, we propose \textbf{\textsc{Principia}}, an LLM-based framework that constructs and continually refines scientific principles at test-time. Drawing on how human researchers develop principles, \textsc{Principia} forms comparative causal hypotheses from related inputs and their observed outcomes, then synthesizes recurring relationships across these hypotheses into inductive principles. By applying the constructed hypotheses and inductive principles to the task, it obtains new observations, revises the used hypotheses based on these new observations, and reconstructs principles from the updated hypotheses, allowing scientific knowledge to continuously evolve throughout test-time discovery. Across benchmarks spanning molecular optimization, protein engineering, gene perturbation, and molecular property prediction, \textsc{Principia} outperforms baselines relying on internal knowledge, external sources, or experience summaries, while generating principles that are plausible and consistent to scientific literature, demonstrating its effectiveness in constructing and refining task-specific hypotheses and principles at test-time for scientific discovery. Our code is available at \url{https://github.com/DongkiKim95/Principia}.
\end{abstract}
\section{Introduction}
\label{sec:introduction}

Scientific discovery has long driven human progress, enabling advances across chemistry, biology, and medicine. Central to this process is identifying scientific principles that govern the systems of interest. In complex scientific discovery processes such as drug discovery, the scientific principles are usually unknown due to the complexity of the underlying systems. Therefore, empirical principles are usually constructed from progressively accumulated experimental observations during test-time discovery to explain causality between input characteristics and experimental outcomes~\citep{scheiner2008general,hone2015prescriptive,pearlstein2021toward}. Specifically, hypotheses that explain a specific scope of individual observations are formed and then generalized into principles that cover a broader scope of observations, being formulated as a hypotheses-to-principle construction pipeline~\citep{lamma2000cooperation,kakas2020abduction,engelschalt2023abductive}.

\begin{figure}[t]
    \centering
    \includegraphics[width=\linewidth]{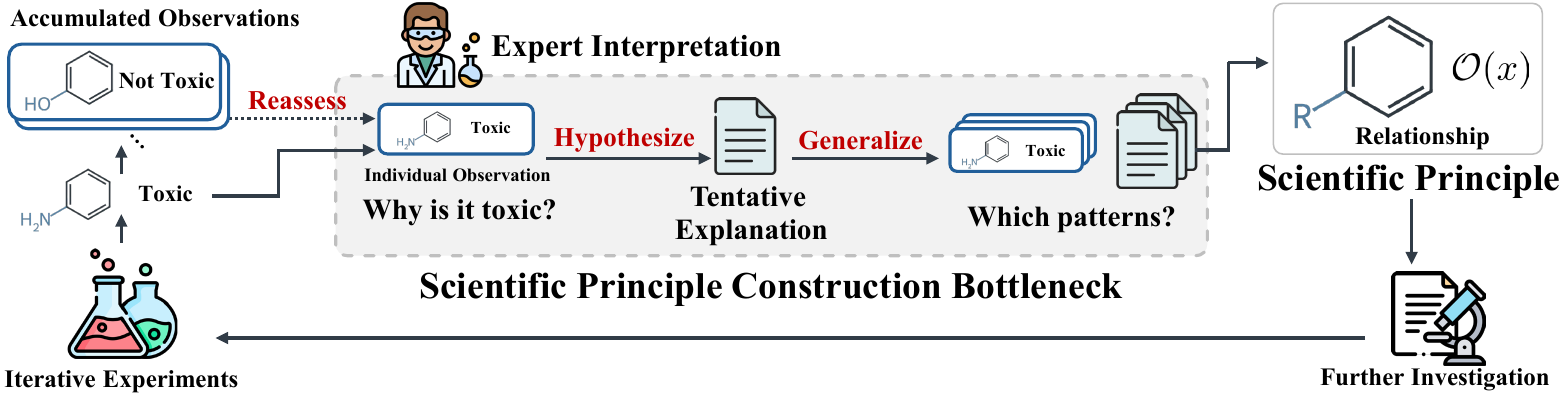}
    \caption{\textbf{Scientific principle construction as a bottleneck in scientific discovery.} As experimental observations accumulate, researchers formulate tentative explanations of individual outcomes, generalize recurring relationships across observations into scientific principles, and reassess these explanations in light of new observations. The resulting principles guide further investigation, forming an iterative discovery cycle. However, this process requires substantial domain expertise, time, labor, and financial resources, limiting the pace of scientific discovery.}
    \vspace{-0.3in}
    \label{fig:scientific_discovery_bottleneck}
\end{figure}

However, constructing such principles by human researchers requires substantial domain expertise, time, and effort, creating a bottleneck in scientific discovery. As illustrated in Fig.~\ref{fig:scientific_discovery_bottleneck}, human researchers also follow the hypotheses-to-principle construction pipeline, while leveraging an extensive body of prior knowledge. They interpret the experimental observations, hypothesize the plausible causal relationships, generalize them by analyzing the recurring patterns, and revise hypotheses when new observations contradict prior explanations. This continued reliance on expert interpretation limits the pace of scientific discovery by slowing the translation of accumulated observations into scientific principles. Recent advances in large language models (LLMs)~\citep{claude_mythos, openai_gpt56, openai_rosalind, xu2026deepseek} have motivated growing efforts to automate scientific research~\citep{ai4science2023impact}, including idea generation~\citep{wang2024scimon, li2025chain, baek2025researchagent}, experimental design~\citep{gottweis2026accelerating, schmidgall2026acceleratingscientificresearchgemini}, and candidate optimization~\citep{wang2025efficient, wang2025large, roohani2025biodiscoveryagent, ran2025exllm}. Despite these advances, constructing and continually refining principles for scientific discovery from progressively accumulated experimental observations remains underexplored.

In this work, we explore the test-time construction of empirical principles that explain how input characteristics contribute to observed outcomes using LLMs. However, operationalizing this process presents two challenges. First, an individual input and its observed outcome provide limited guidance for an LLM to distinguish among possible explanations. Multiple input characteristics can account for the same outcome, making it difficult to identify which factors to emphasize when generating a hypothesis. Hypothesis generation therefore requires evidence that helps narrow the range of possible explanations. Second, the observations accumulated during discovery cover only a small fraction of the vast space of scientific inputs. LLM-generated hypotheses based on this limited coverage can provide incomplete or incorrect explanations for other inputs, and these limitations can carry over into the principles constructed from them. As new observations accumulate, they can expose these limitations, making it necessary to reassess and revise earlier hypotheses.

To address these challenges, we propose \textbf{\textsc{Principia}}, an LLM-based hypotheses-to-principle construction framework that inductively derives principles from progressively accumulated experimental observations during test-time discovery. Drawing on how human researchers develop scientific knowledge, \textsc{Principia} consists of three key components: (i) comparative causal hypothesis generation, (ii) inductive principle construction, and (iii) hypothesis revision via new observations. Specifically, for hypothesis generation, \textsc{Principia} compares related inputs and their experimental outcomes rather than a single observation to formulate causal hypotheses, narrowing the factors considered to their shared and differing characteristics. For principle construction, it groups related inputs and inductively combines recurring relationships across their hypotheses into principles. These hypotheses and principles guide subsequent prediction or optimization through retrieval, getting new observations. Finally, new observations are used to revise the used hypotheses and reconstruct the resulting principles, allowing the knowledge to evolve during test-time discovery.

We evaluate \textsc{Principia} on extensive benchmarks of scientific discovery tasks, spanning molecular, protein, and gene optimization, as well as molecular property prediction. Our results demonstrate that \textsc{Principia} outperforms baselines that rely on internal knowledge, experience summaries, or external knowledge, demonstrating the effectiveness of test-time construction of hypotheses and principles. Further analyses show that \textsc{Principia} achieves comparable optimization performance with a 55\% reduction in experimental budget compared with an experience-summary baseline, demonstrating its cost-effectiveness in scientific optimization. Qualitative analyses show that the constructed hypotheses capture relationships between inputs and experimental outcomes, while the induced principles are consistent with scientific literature. These results show that constructing empirical principles at test time can improve both the effectiveness and efficiency of scientific discovery.

\vspace{-0.05in}
\section{Related Works}
\label{sec:related_works}
\paragraph{Automated Scientific Discovery and Optimization} Advances in machine learning have emerged as powerful tools for automating scientific discovery, ranging from systems that assist individual stages, such as hypothesis and idea generation~\citep{wang2024scimon, vasu2025hyper, li2025chain, baek2025researchagent}, to end-to-end AI scientists~\citep{schmidgall2025agent, schmidgall2026acceleratingscientificresearchgemini, lu2026towards, gottweis2026accelerating} that formulate research proposals, conduct experiments, and generate scientific papers with minimal human supervision. While these works seek to automate the broader research workflow, another line of research focuses on discovering novel scientific candidates with minimal prior context. For example, \citet{bagal2021molgpt, xu2022geodiff, lee2023exploring, watson2023novo} develop deep generative models that explore latent spaces to identify promising candidates by learning representations of the underlying scientific domains. Meanwhile, with the emergence of Large Language Models (LLMs), agentic approaches for candidate generation by LLMs have been developed~\citep{m2024augmenting, roohani2025biodiscoveryagent}. Building on this direction, recent approaches leverage the reasoning capabilities of LLMs, using them as surrogate operators in Evolutionary Algorithms~\citep{holland1962outline} to guide exploration of vast search spaces through iterative generation and evolutionary search~\citep{romera2024mathematical, wang2025efficient, wang2025large, novikov2025alphaevolve, yuksekgonul2026learning, lange2026shinkaevolve, zhang2026darwin} across diverse scientific research domains. Despite their progress, constructing reusable scientific knowledge from experimental observations during discovery remains underexplored. Our work addresses this gap by dynamically constructing reusable causal principles at test time, enabling more informed discovery. 

\paragraph{Knowledge-Augmented Scientific Discovery} 
Building on the success of Retrieval-Augmented Generation (RAG)~\citep{lewis2020retrieval}, a growing body of work has incorporated external knowledge into scientific discovery systems. These approaches include leveraging existing literature to generate open-ended hypotheses and research proposals through multi-agent systems~\citep{wang2024scimon, li2025chain, baek2025researchagent, huang2025idea2plan, liu2025scientific}, incorporating external knowledge bases into the discovery loop to inform experimental decisions~\citep{boiko2023autonomous, m2024augmenting, roohani2025biodiscoveryagent}, and employing deep-search agents to retrieve promising research directions~\citep{liu2025scientific}. Recent work has also explored using experience during discovery to guide subsequent exploration. For instance, \citet{pu2025piflow} balance exploration and exploitation over pre-configured high-quality scientific knowledge to guide subsequent optimization of both candidates and principles, while \citet{ran2025exllm} summarize characteristics of previously discovered candidates into a single document. Another line of works leverages predictive strategies for machine learning research~\citep{jiang2026deltaevolve, luo2026seaevo} or constructs offline memories from heterogeneous tasks for future reuse~\citep{zhang2026drugsage, yao2026towards}. Beyond retrieving literature or reusing summarized experience, our work focuses on causal principle induction from observations accumulated during test-time discovery. Rather than using these observations solely to guide subsequent actions, we develop explicit hypotheses about how input characteristics may contribute to observed outcomes and generalize recurring relationships into principles.

\section{Principia: Test-time Hypotheses to Principle Construction}
\label{sec:methodology}

\begin{figure}[t]
    \centering
    \includegraphics[width=\linewidth]{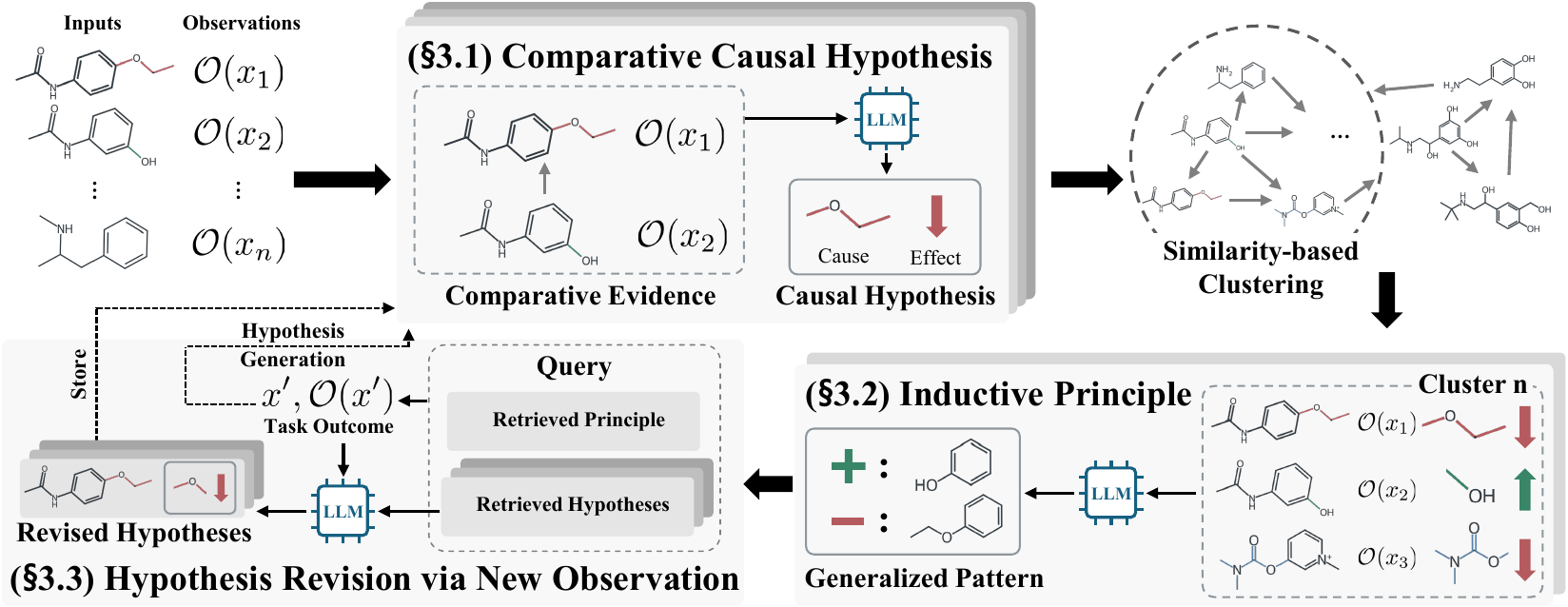}
    
    \caption{\textbf{Overview of \textsc{Principia}.} Given scientific inputs and their observed outcomes, \textsc{Principia} forms comparative causal hypotheses that explain how input characteristics may contribute to the outcomes. It groups related inputs through similarity-based clustering and inductively generalizes their associated hypotheses into scientific principles. Retrieved hypotheses and principles guide subsequent prediction or candidate generation. New observations are then used to revise the hypotheses involved, and the updated hypotheses support principle reconstruction, forming a continuous cycle of scientific knowledge construction and application.}
    \label{fig:method_overview}
\end{figure}

We present \textsc{Principia}, an LLM-based framework that constructs and refines empirical principles at test time by forming and generalizing hypotheses from progressively accumulated experimental observations (Fig.~\ref{fig:method_overview}).
We first introduce comparative causal hypotheses formed from related inputs and their observed outcomes, inductive principle construction that generalizes recurring patterns across these hypotheses, and hypotheses revision that applies the constructed hypotheses and principles to get new observations and updates the previously formed hypotheses via new observations.

\paragraph{Task Settings}
We consider two central tasks in scientific discovery: optimization and prediction. Let $x \in \mathcal{X}$ denote a scientific input, such as a molecule, and let $\mathcal{O}$ denote an experimental oracle that returns an observed property or characteristic $\mathcal{O}(x)$. We refer to this outcome as a \textit{scientific observation}, which could be a categorical label or a numerical measurement. In the optimization task, the goal is to find an optimal instance $x^*$ that best achieves a specified scientific objective such as maximizing a desired property, through successive modifications. Given an input $x$, the action is to propose a modified input $x'$, which is evaluated by the experimental oracle to obtain the observation $\mathcal{O}(x')$. In the prediction task, the goal is to predict a property or characteristic of a given input $x$. The action is to produce a prediction $\hat{\mathcal{O}}(x)$, which is evaluated against the observation $\mathcal{O}(x)$ without access to that observation during prediction. In both settings, the observations provide evidence for constructing and refining the knowledge that guides subsequent modifications or predictions.

\subsection{Comparative Causal Hypothesis}
\label{sec:comparative_hypotheses}

We now turn to the first step of our scientific knowledge construction, which is the formation of causal hypotheses. The goal of hypothesis formation is to propose explicit explanations of the factors that have contributed to an observed experimental outcome. However, an individual input and its observed outcome alone generally provide limited evidence for distinguishing the contributions of different input characteristics. To provide a more reliable basis for hypothesis formation, we construct comparative evidence by combining related inputs with their corresponding observed outcomes. From this evidence, we generate a \textit{comparative causal hypothesis}, defined as a tentative explanation of how similarities and differences between two inputs account for their observed outcomes.

\paragraph{Comparative Evidence}
To narrow the scope of comparison, we first form comparative evidence which consists of two similar inputs and their experimental outcomes. Specifically, we compose reference-to-target comparisons $x_i\rightarrow x_j$ from the available experimental data, $\mathcal{D}=\{(x_i,\mathcal{O}(x_i))\}_{i=1}^{N}$, by selecting related inputs in the sense that they share similar input characteristics. In the optimization task, we compare a generated candidate $x_j$ with the input $x_i$ from which it was derived, using their observed outcomes to examine the effects of the modification. In the prediction task, such modification histories are not generally available, so we compute the pairwise distances between each target input $x_j$ and nearby reference inputs $x_i$ from $\mathcal{D}$. To compute the pairwise distances, we leverage domain-appropriate distance measures, such as fingerprint-based distance for molecules, and select references with the smallest distances, regardless of whether their observed labels match the target's label. In both settings, each pair provides comparative evidence consisting of both inputs and their observations: $e_{i\rightarrow j}=\big(x_i,\mathcal{O}(x_i),x_j,\mathcal{O}(x_j)\big)$.

\paragraph{Hypothesis Generation}
Then, an LLM generates textual hypotheses $H_{i\rightarrow j}$ from the comparative evidence and a task description: $H_{i\rightarrow j}=\mathrm{LLM}_{\text{hypothesis}}\big(e_{i\rightarrow j};q\big)$, where $q$ is the task description. Based on this comparative evidence, an LLM identifies the similarities and differences between the inputs and their observations and links an input characteristic to their observations, proposing a possible explanation of the causal relationship grounded in the comparison. Consequently, when outcomes differ, the hypotheses focus on differences between the inputs that can explain the observed outcomes. Otherwise, they focus on describing shared characteristics that can maintain the observed outcomes. We then store each hypothesis together with its comparative evidence, $(e_{i\rightarrow j},H_{i\rightarrow j})$, in a graph-structured hypothesis memory $\mathcal{M}_h$, where nodes represent inputs and each directed edge stores the comparative evidence and its associated causal hypothesis for a reference-to-target comparison, building scientific knowledge specific to the inputs and observations involved.

\subsection{Inductive Principle}
\label{sec:principle_construction}

Even though each comparative causal hypothesis provides a tentative explanation grounded in a particular pair of inputs, a single hypothesis could depend on characteristics specific to those inputs, failing to model the general underlying scientific principles. Inspired by the human scientific process where the scientist generalizes from specific observations to broader principles, we construct \textit{inductive principles} that summarize recurring relationships between input characteristics and observed outcomes across multiple comparisons.

\paragraph{Similarity-based Clustering}
We first group the inputs represented by nodes in $\mathcal{M}_h$ to examine recurring patterns among inputs with similar characteristics by applying $k$-medoids clustering with the domain-specific pairwise distances. Let $G_k$ denote the set of input indices in the $k$-th cluster and $\mathcal{A}_h$ denote the directed edge set of the hypothesis graph $\mathcal{M}_h$. For each cluster, we collect the comparative evidence and causal hypotheses on all edges incident to its members: $\mathcal{E}_k = \left\{(e_{i\rightarrow j}, H_{i\rightarrow j}) \;|\; (i,j)\in\mathcal{A}_h, i\in G_k \lor j\in G_k\right\}$. This grouping brings together inputs with similar characteristics and the hypotheses associated with their comparisons, providing a basis for identifying recurring relationships across observations.

\paragraph{Principle Induction}
For each cluster, an LLM inductively infers a compact principle based on its associated evidence and hypotheses: $P_k=\mathrm{LLM}_{\text{principle}}\big(\mathcal{E}_k;q\big)$. The model identifies recurring relationships across the associated causal hypotheses and generalizes them into a causal principle. By integrating these across multiple comparisons, this process extends beyond individual observations to derive a causal principle that captures more general relationships. We store each principle with its associated cluster in a principle memory $\mathcal{M}_p=\{(G_k,P_k)\}_{k=1}^{K}$, organizing the induced principles into a more general level of scientific knowledge.

\subsection{Hypothesis Revision via New Observations}
\label{sec:hypothesis_revision}

Thus far, we have described how to generate comparative causal hypotheses and inductive principles from experimental observations: the inductive principles are constructed from comparative causal hypotheses, making the reliability of these principles dependent on the hypotheses underlying them. However, experimental observations cover only a small fraction of the space of possible inputs, and hypotheses formed from this limited evidence can overlook factors relevant to other inputs. These limitations can also affect the principles constructed from them, leading to potentially inaccurate or incomplete conclusions. Therefore, we revise hypotheses to enhance their reliability and improve the quality of the principles constructed from them. Specifically, inspired by how scientists validate existing hypotheses through further investigation, we use the generated hypotheses and principles to guide further task-specific predictions or candidate generation, obtaining additional experimental observations. We then revise the hypotheses using new observations, either by comparing predictions with observed labels or by examining the evaluated outcomes of generated candidates.

\paragraph{Hypothesis Application and Revision}
To obtain new observations for hypothesis revision, we first retrieve relevant hypotheses and principles and apply them to new inputs. Given a query input $x$, we first construct hypothesis and principle candidate pools drawn from $\mathcal{M}_h$ and $\mathcal{M}_p$, and then leverage an LLM-based retrieval, where an LLM that can handle long-context inputs independently selects $k_h$ hypotheses $\mathcal{R}_h(x)$ and $k_p$ principles $\mathcal{R}_p(x)$ from candidate pools, respectively, based on their relevance to the query and task description. The selected knowledge guides the task-specific action such as optimization or prediction, after which the corresponding observations are obtained by these actions. For more details on the retrieval and application of knowledge, please refer to Appendix~\ref{app:sec:implementation_details}.

We then revise the hypotheses retrieved and used during the current round. For each selected hypothesis $H_{i\rightarrow j}$, we construct new observation $\mathcal{B}_{i\rightarrow j}$ by sampling new observations from that round together with their associated inputs and outcomes. Specifically, in the optimization task, $\mathcal{B}_{i\rightarrow j}$ contains input pairs $(x,x')$ by the modification from $x$ to $x'$ and their observed outcomes $(\mathcal{O}(x),\mathcal{O}(x'))$. In the prediction task, it contains query inputs $x$, their observed labels $\mathcal{O}(x)$, and predictions $\hat{\mathcal{O}}(x)$ made without access to those labels. An LLM then revises each selected hypothesis using new observations, $\mathcal{B}_{i\rightarrow j}$, and the task description: $H'_{i\rightarrow j}=\mathrm{LLM}_{\text{refine}}\big(e_{i\rightarrow j},H_{i\rightarrow j},\mathcal{B}_{i\rightarrow j};q\big)$. By grounding revision in new observations obtained from applying the knowledge to new inputs, the framework provides an empirical basis for assessing and refining the hypotheses beyond the comparisons from which they were originally formed.

\paragraph{Knowledge Update}
After revision from additional observations, we replace $H_{i\rightarrow j}$ with its revised version $H'_{i\rightarrow j}$ in $\mathcal{M}_h$, while retaining the original comparative evidence $e_{i\rightarrow j}$. Then, the new observations and inputs are also used for generating their hypotheses following the procedure in Section~\ref{sec:comparative_hypotheses}. This extends the hypothesis memory with new observations while revision updates existing hypotheses. We then reconstruct principles from the updated hypotheses and associated observations following the procedure in Section~\ref{sec:principle_construction}. The resulting knowledge memories guide subsequent investigation, connecting observation, hypothesis generation, knowledge application, and revision via new observations in a continuous loop.

\section{Experimental Results}
\label{sec:experimental_results}

We evaluate \textsc{Principia} on scientific discovery tasks including scientific optimization tasks spanning molecules, proteins, and genes, as well as molecular property prediction.
We provide implementation details in Appendix~\ref{app:sec:implementation_details} and the algorithmic workflow for optimization and prediction in Appendix~\ref{app:sec:algorithms}.

\subsection{Scientific Optimization}
\label{sec:scientific_optimization}

\paragraph{Experimental Setting}
We evaluate \textsc{Principia} on scientific optimization tasks under a restricted oracle budget spanning molecules, proteins, and genes from the Scientific Discovery Evaluation benchmark~\citep{song2025evaluating}, involving designing molecules with desired properties~\citep{gao2022sample}, optimizing protein sequences for target functions~\citep{wang2025large}, and identifying gene perturbations that produce desired cellular responses~\citep{schmidt2022crispr,scharenberg2023spns1}. We report AUC@10 to measure optimization progress over oracle evaluations and the average oracle scores of the top 100 molecules or top 50 protein sequences.  For gene perturbation, we additionally report hit ratio following \citet{roohani2025biodiscoveryagent}. To ensure fair comparisons across variants, we normalize the fitness values of common variants for the Syn-3bfo, Scharenberg22, and IFNG tasks to the range $[0,1]$ when computing AUC. We use GPT-5-mini~\citep{singh2025openai} as the backbone. We provide task-specific objectives, the application of \textsc{Principia}, search budgets, and evaluation protocols in Appendix~\ref{app:sec:experimental_settings}.

\paragraph{Baselines}
We compare against MolLEO~\citep{wang2025efficient}, an optimizer without guidance, ExLLM~\citep{ran2025exllm} which guides optimization by an evolving textual summary from experience, WebSearch which provides guidance retrieved from the web, DeepEvolve~\citep{liu2025scientific} which uses web search guidance based on the optimization experience, and PiFlow~\citep{pu2025piflow} which guides optimization without observation-guided causal hypotheses. For molecular optimization, we additionally include REINVENT~\citep{olivecrona2017molecular}, a reinforcement learning method for molecular generation, Augmented Memory~\citep{guo2023augmented} which enhances reinforcement learning through experience replay, and Graph GA~\citep{jensen2019graph} which applies genetic operations to molecular graphs. We provide implementation details and configurations in Appendix~\ref{app:sec:optimization_baselines}.

\paragraph{Results}
As shown in Tables~\ref{tab:main_drug_discovery} and~\ref{tab:main_protein_gene}, \textsc{Principia} consistently outperforms baseline without guidance (MolLEO), demonstrating that constructing task-specific knowledge is beneficial for guiding optimization. \textsc{Principia} also outperforms baselines that use experience summaries (ExLLM), showing that constructing structured knowledge including causal hypotheses and inductive principles is more effective than a textual summary of past observations. Notably, \textsc{Principia} also outperforms baselines that use external knowledge (WebSearch, DeepEvolve), indicating that constructing task-specific knowledge at test-time provides core information to achieve the optimization objective, while external knowledge is less relevant to the specific task. On the other hand, the baseline without explicit modeling of causal relationships (PiFlow) often underperforms the baseline without guidance (MolLEO), suggesting that constructing causal hypotheses and generalizing them into principles is crucial for guiding optimization. These results demonstrate the effectiveness of constructing task-specific knowledge from accumulated observations under limited evaluation budgets.

\begin{table}[t!]
\caption{Comparison of \textsc{Principia} with ML-based and LLM-based optimization baselines on molecular optimization tasks. We report averaged values over three different runs for each metric. $^{*}$ denotes methods reimplemented for our evaluation setting. Avg. denotes the mean of each metric across all tasks. The best results are highlighted in \textbf{bold}.}
\vspace{-0.1in}
\centering
\setlength{\tabcolsep}{6pt}
\renewcommand{\arraystretch}{1.25}
\resizebox{\textwidth}{!}{%
\begin{tabular}{@{}l*{10}{c}aa}
\toprule
 & \multicolumn{2}{c}{ \textbf{JNK3}} & \multicolumn{2}{c}{ \textbf{GSK3$\beta$}} & \multicolumn{2}{c}{ \textbf{Median1}} & \multicolumn{2}{c}{ \textbf{Median2}} & \multicolumn{2}{c}{ \textbf{Valsartan Smarts}} & \multicolumn{2}{c}{\textbf{Avg.}} \\
\cmidrule(lr){2-3}
\cmidrule(lr){4-5}
\cmidrule(lr){6-7}
\cmidrule(lr){8-9}
\cmidrule(lr){10-11}
\cmidrule(lr){12-13}
\textbf{Methods} & Avg@100 & AUC@10 & Avg@100 & AUC@10 & Avg@100 & AUC@10 & Avg@100 & AUC@10 & Avg@100 & AUC@10 & Avg@100 & AUC@10 \\
\midrule
REINVENT & 0.164 & 0.160 & 0.438 & 0.368 & 0.202 & 0.187 & 0.164 & 0.167 & 0.000 & 0.000 & 0.194 & 0.176 \\
Memory-Aug & 0.282 & 0.211 & 0.535 & 0.414 & 0.213 & 0.189 & 0.178 & 0.167 & 0.000 & 0.000 & 0.242 & 0.196 \\
Graph GA & 0.316 & 0.290 & 0.352 & 0.315 & 0.175 & 0.177 & 0.162 & 0.167 & 0.000 & 0.000 & 0.201 & 0.190 \\
\cdashline{1-13}[0.5pt/1pt]\rule{-2pt}{2.5ex}
MolLEO$^{*}$ & 0.266 & 0.273 & 0.400 & 0.341 & 0.293 & 0.280 & 0.197 & 0.189 & 0.018 & 0.028 & 0.235 & 0.222 \\
ExLLM$^{*}$ & 0.245 & 0.227 & 0.438 & 0.363 & 0.293 & 0.281 & 0.198 & 0.190 & 0.096 & \textbf{0.188} & 0.254 & 0.250 \\
PiFlow$^{*}$ & 0.050 & 0.057 & 0.223 & 0.278 & 0.238 & 0.255 & 0.186 & 0.196 & 0.000 & 0.000 & 0.139 & 0.157 \\
WebSearch$^{*}$ & 0.192 & 0.199 & 0.505 & 0.421 & 0.290 & 0.277 & 0.197 & 0.189 & 0.040 & 0.066 & 0.245 & 0.230 \\
DeepEvolve$^{*}$ & 0.248 & 0.234 & 0.468 & 0.382 & 0.294 & 0.277 & 0.184 & 0.181 & 0.047 & 0.084 & 0.248 & 0.232 \\
\cdashline{1-13}[0.5pt/1pt]\rule{-2pt}{2.5ex}
\textbf{\textsc{Principia} (Ours)} & \textbf{0.448} & \textbf{0.350} & \textbf{0.617} & \textbf{0.452} & \textbf{0.298} & \textbf{0.284} & \textbf{0.220} & \textbf{0.203} & \textbf{0.117} & 0.106 & \textbf{0.340} & \textbf{0.279} \\
\bottomrule
\end{tabular}
}
\label{tab:main_drug_discovery}
\end{table}

\begin{table}[t!]
\caption{Comparison of \textsc{Principia} with LLM-based optimization baselines on protein sequence engineering and gene perturbation tasks. We report averaged values over three different runs for each metric. $^{*}$ denotes methods reimplemented for our evaluation setting. Avg. denotes the mean AUC@10 across all tasks. The best results are highlighted in \textbf{bold}.}
\vspace{-0.1in}
\centering
\setlength{\tabcolsep}{6pt}
\renewcommand{\arraystretch}{1.25}
\resizebox{\textwidth}{!}{%
\begin{tabular}{@{}l*{10}{c}a}
\toprule
 & \multicolumn{6}{c}{\textbf{Protein Sequence Engineering}} & \multicolumn{4}{c}{\textbf{Gene Perturbation}} & \multicolumn{1}{c}{} \\
\cmidrule(lr){2-7}
\cmidrule(lr){8-11}
 & \multicolumn{2}{c}{\textbf{Syn-3bfo}} & \multicolumn{2}{c}{\textbf{AAV}} & \multicolumn{2}{c}{\textbf{GB1}} & \multicolumn{2}{c}{\textbf{Scharenberg22}} & \multicolumn{2}{c}{\textbf{IFNG}} & \multicolumn{1}{c}{\textbf{Avg.}} \\
\cmidrule(lr){2-3}
\cmidrule(lr){4-5}
\cmidrule(lr){6-7}
\cmidrule(lr){8-9}
\cmidrule(lr){10-11}
\cmidrule(lr){12-12}
\textbf{Methods} & Avg@50 & AUC@10 & Avg@50 & AUC@10 & Avg@50 & AUC@10 & Hit Ratio & AUC@10 & Hit Ratio & AUC@10 & AUC@10 \\
\midrule
MolLEO$^{*}$ & -0.374 & 0.737 & 0.664 & 0.524 & 0.201 & 0.245 & 0.270 & 0.606 & 0.107 & 0.717 & 0.566 \\
ExLLM$^{*}$ & -0.286 & 0.738 & 0.660 & 0.522 & 0.236 & 0.277 & 0.277 & 0.605 & 0.109 & 0.735 & 0.575 \\
PiFlow$^{*}$ & -32.149 & 0.079 & 0.049 & 0.072 & 0.111 & 0.188 & 0.289 & 0.557 & 0.111 & 0.682 & 0.316 \\
WebSearch$^{*}$ & -0.367 & 0.738 & 0.659 & \textbf{0.527} & 0.197 & 0.233 & 0.226 & 0.533 & 0.110 & 0.702 & 0.547 \\
DeepEvolve$^{*}$ & -0.378 & 0.737 & 0.664 & 0.524 & 0.196 & 0.275 & 0.270 & 0.539 & 0.108 & 0.560 & 0.527 \\
\cdashline{1-12}[0.5pt/1pt]\rule{-2pt}{2.5ex}
\textbf{\textsc{Principia} (Ours)} & \textbf{0.084} & \textbf{0.741} & \textbf{0.681} & \textbf{0.527} & \textbf{0.398} & \textbf{0.383} & \textbf{0.308} & \textbf{0.630} & \textbf{0.114} & \textbf{0.742} & \textbf{0.605} \\
\bottomrule
\end{tabular}
}
\label{tab:main_protein_gene}
\end{table}

\subsection{Molecular Property Prediction}
\label{sec:molecular_property_prediction}
\paragraph{Experimental Setting}
We evaluate \textsc{Principia} on molecular property prediction with eight binary classification tasks from TDC~\citep{huang2021therapeutics}: four toxicity-related tasks (DILI, hERG, Ames, and Skin Reaction) and four CYP inhibition tasks (CYP1A2, CYP2C9, CYP2D6, and CYP3A4). We use a scaffold split to construct a held-out test set containing molecular scaffolds absent from the training set. To simulate the test-time scenario for prediction tasks, we equally split the training dataset and construct knowledge by progressively observing the labels of each subset. After constructing knowledge from the training dataset, we evaluate the resulting knowledge on the held-out test set, whose labels are excluded from knowledge construction and revision. We use GPT-5.4-mini as the backbone. We report accuracy and macro-F1, averaged over three random seeds. We provide details on data preparation, round construction, and evaluation protocols in Appendix~\ref{app:sec:prediction_settings}.

\paragraph{Baselines}
We compare against Direct prediction which predicts without additional context, and Few-shot prediction which conditions predictions on $k$ examples from the   whole training dataset composed of input molecules and their labels. We further consider Textual summary which summarizes sequentially observed training examples into texts, Internal principles which provides principles generated from the LLM's internal knowledge, and External principles which provides principles retrieved from web. We further provide implementation details and configurations in Appendix~\ref{app:sec:prediction_baselines}.

\begin{table*}[t]
\centering
\caption{Comparison of \textsc{Principia} with direct prediction, few-shot prediction, and knowledge-augmented baselines on molecular property prediction tasks. We report averaged values over three different runs for each metric. The best results are highlighted in \textbf{bold}.}
\label{tab:molecular_property_prediction}
\vspace{-0.1in}
\setlength{\tabcolsep}{6pt}
\renewcommand{\arraystretch}{1.3}
\resizebox{\textwidth}{!}{%
\begin{tabular}{@{}l*{16}{c}aa}
\toprule
 & \multicolumn{2}{c}{\textbf{DILI}} & \multicolumn{2}{c}{\textbf{hERG}} & \multicolumn{2}{c}{\textbf{Ames}} & \multicolumn{2}{c}{\textbf{Skin}} & \multicolumn{2}{c}{\textbf{CYP1A2}} & \multicolumn{2}{c}{\textbf{CYP2C9}} & \multicolumn{2}{c}{\textbf{CYP2D6}} & \multicolumn{2}{c}{\textbf{CYP3A4}} & \multicolumn{2}{c}{\textbf{Avg.}} \\
\cmidrule(lr){2-3}\cmidrule(lr){4-5}\cmidrule(lr){6-7}\cmidrule(lr){8-9}\cmidrule(lr){10-11}\cmidrule(lr){12-13}\cmidrule(lr){14-15}\cmidrule(lr){16-17}\cmidrule(lr){18-19}
\textbf{Methods} & Acc & F1 & Acc & F1 & Acc & F1 & Acc & F1 & Acc & F1 & Acc & F1 & Acc & F1 & Acc & F1 & Acc & F1 \\
\midrule
Direct & 52.43 & 44.11 & 51.33 & 37.27 & 58.33 & 57.51 & 45.12 & 43.54 & 63.33 & 60.18 & 67.33 & 64.73 & 68.33 & 66.64 & 60.00 & 54.26 & 58.28 & 53.53 \\
\cdashline{1-19}[0.5pt/1pt]\rule{-2pt}{2.5ex}
Few-shot ($k{=}1$) & 59.72 & 54.68 & 58.67 & 53.16 & 59.33 & 59.02 & 52.03 & 51.30 & 66.33 & 64.98 & 68.00 & 66.95 & 69.33 & 68.93 & 59.00 & 57.21 & 61.55 & 59.53 \\
Few-shot ($k{=}3$) & 62.15 & 57.95 & 62.00 & 59.76 & 62.33 & 61.03 & 52.85 & 51.49 & 64.67 & 63.06 & 70.33 & 69.75 & 71.33 & 71.31 & 61.00 & 59.63 & 63.33 & 61.75 \\
Few-shot ($k{=}5$) & 62.85 & 57.82 & 64.67 & 62.83 & 62.00 & 60.53 & 50.81 & 49.62 & 67.67 & 66.00 & 66.67 & 66.18 & 71.67 & 71.53 & 62.00 & 60.42 & 63.54 & 61.87 \\
Few-shot ($k{=}7$) & 62.15 & 57.74 & 64.00 & 62.60 & 62.33 & 60.94 & 50.81 & 48.40 & 66.33 & 64.65 & 69.67 & 69.42 & 71.67 & 71.65 & 64.67 & 62.97 & 63.95 & 62.30 \\
\cdashline{1-19}[0.5pt/1pt]\rule{-2pt}{2.5ex}
Internal principles & 55.21 & 49.74 & 53.67 & 44.26 & 56.67 & 53.52 & 51.63 & 49.45 & 68.33 & 67.34 & 66.33 & 62.67 & 69.67 & 68.90 & 64.33 & 61.91 & 60.73 & 57.22 \\
External principles & 54.51 & 48.13 & 61.00 & 58.52 & 55.67 & 52.18 & 52.44 & 50.77 & 67.67 & 66.17 & 62.00 & 60.71 & 67.33 & 66.91 & 60.67 & 54.99 & 60.16 & 57.30 \\
Textual summary & 61.46 & 56.95 & 64.33 & 62.70 & 57.00 & 55.26 & 51.63 & 50.81 & 67.67 & 64.86 & 70.00 & 67.10 & 70.67 & 70.12 & 63.33 & 60.40 & 63.26 & 61.03 \\
\cdashline{1-19}[0.5pt/1pt]\rule{-2pt}{2.5ex}
\textbf{\textsc{Principia} (Ours)} & \textbf{70.49} & \textbf{67.48} & \textbf{70.33} & \textbf{69.20} & \textbf{67.67} & \textbf{67.30} & \textbf{58.54} & \textbf{58.44} & \textbf{71.33} & \textbf{70.81} & \textbf{76.67} & \textbf{76.72} & \textbf{73.00} & \textbf{72.76} & \textbf{70.00} & \textbf{69.56} & \textbf{69.75} & \textbf{69.03} \\
\bottomrule
\end{tabular}%
}
\end{table*}

\paragraph{Results}
As shown in Table~\ref{tab:molecular_property_prediction}, \textsc{Principia} consistently outperforms all baselines across eight molecular property prediction tasks, demonstrating the effectiveness of constructing hypotheses and principles for molecular property prediction. In particular, its improvements over few-shot prediction suggest that organizing observations into explicit causal hypotheses and principles provides more effective guidance than supplying labeled examples alone. Since the knowledge is constructed from training observations and evaluated on held-out molecules with different structures, these results further suggest that the constructed knowledge captures relationships that extend beyond the observed examples and enhances the prediction accuracies for unseen molecules.

\subsection{Analyses}
\label{sec:analysis}

\begin{figure*}[t!]
\centering
\begin{minipage}[t]{0.31\textwidth}
    \vspace{0pt}
\captionof{table}{Analysis of the effect of causal hypotheses on protein sequence engineering tasks, including GB1 and Syn-3bfo.
}
\vspace{-0.08in}
\label{tab:effect-of-causal-hypotheses}
\centering
\resizebox{\linewidth}{!}{%
\setlength{\tabcolsep}{1.6pt}
\renewcommand{\arraystretch}{1.1}
\begin{tabular}{@{}p{0.4cm}lcc@{}}
\toprule
& \textbf{Variant}
& \textbf{AUC@10} $\uparrow$
& \textbf{$\Delta$} $\uparrow$ \\
\midrule

\multicolumn{4}{l}{\textbf{GB1}} \\

&
None
& $0.245$
& --- \\

&
\textbf{w/ Hypotheses}
& $\mathbf{0.337}$
& \textcolor{teal}{$+\mathbf{0.092}$} \\

\noalign{\vskip 0.4ex}
\midrule

\multicolumn{4}{l}{\textbf{Syn-3bfo}} \\[-0.2ex]

&
None
& $-0.019$
& --- \\

&
\textbf{w/ Hypotheses}
& $\mathbf{0.116}$
& \textcolor{teal}{$+\mathbf{0.135}$} \\
\bottomrule
\end{tabular}
}

\end{minipage}
\hfill
\begin{minipage}[t]{0.64\textwidth}
    \input{Figures/comparative_hypothesis}
\end{minipage}
\vspace{-0.05in}
\end{figure*}

\paragraph{Effect of Comparative Causal Hypotheses}
We analyze the effect of comparative causal hypotheses on scientific optimization. Table~\ref{tab:effect-of-causal-hypotheses} shows that providing these hypotheses improves performance over the optimizer without hypothesis guidance, indicating that explicit hypotheses about the relationships between input characteristics and observed outcomes provide useful guidance for subsequent optimization. We also compare comparative hypothesis generation with a non-comparative variant that generates each hypothesis from a single input and its observed outcome (w/o comparative). As shown in Fig.~\ref{fig:comparative_hypothesis}, comparative hypothesis generation improves optimization performance, supporting our rationale that comparative evidence helps narrow the factors considered during hypothesis generation and yields more effective guidance.

\begin{figure*}[t!]
\centering
\begin{minipage}[t]{0.31\textwidth}
    \input{Figures/principle_refinement_ablation}
\end{minipage}
\hfill
\begin{minipage}[t]{0.31\textwidth}
    \input{Figures/sample_efficiency}
\end{minipage}
\hfill
\begin{minipage}[t]{0.31\textwidth}
    \input{Figures/lclm_retrieval}
\end{minipage}
\vspace{-0.1in}
\end{figure*}

\begin{table*}[t]
    \centering
    \caption{An example of comparative hypotheses and an induced principle. Matching colors link epoxide rings (blue), phenoxy groups (green), and aminoalcohol side chains (red) to their descriptions. Arrows indicate reference-to-target comparisons, and labels are dataset observations.}
    \vspace{-0.1in}
    \label{tab:hypothesis_principle_qualitative}
    \begingroup
\color{black}
\definecolor{amesepoxide}{RGB}{0,90,200}
\definecolor{amesphenoxy}{RGB}{0,125,65}
\definecolor{amesaminoalcohol}{RGB}{200,40,40}
\newcommand{\amep}[1]{\textcolor{amesepoxide}{\textbf{#1}}}
\newcommand{\amcore}[1]{\textcolor{amesphenoxy}{\textbf{#1}}}
\fontsize{8.5}{10.5}\selectfont
\setlength{\tabcolsep}{0pt}
\renewcommand{\arraystretch}{1.05}
\newsavebox{\amesprinciplebox}
\sbox{\amesprinciplebox}{%
\begin{minipage}[t]{0.35\linewidth}
    \vspace{0pt}\vspace{2pt}
    \textbf{Positive pattern}\par
    \centering
    \includegraphics[width=0.5\linewidth]{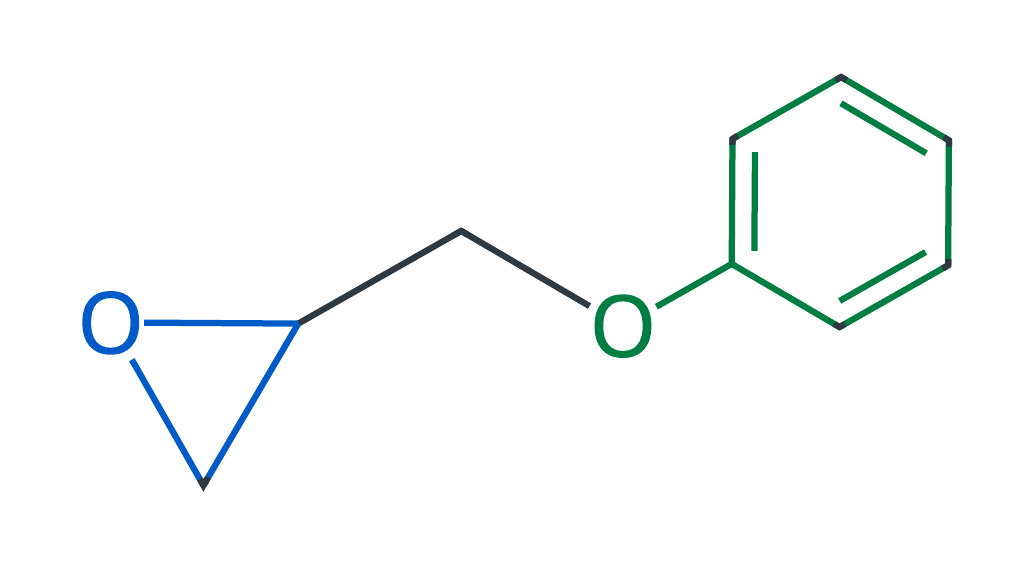}\par
    \raggedright
    \amep{Epoxide/oxirane-bearing} \amcore{aryl ether} motifs recur as a likely mutagenicity driver, especially Ar-O-CH2-CH \ldots{}
    \par\smallskip
    \textbf{Negative pattern}\par\smallskip
    More polar \textcolor{amesaminoalcohol}{\textbf{aminoalcohol}} side chains \ldots{} are often tolerated when they do not contain a strained electrophile.
\end{minipage}}
\newsavebox{\amesfirstobservations}
\sbox{\amesfirstobservations}{%
\begin{minipage}[t]{0.38\linewidth}
    \vspace{0pt}\vspace{2pt}
    \begin{minipage}[c]{0.455\linewidth}
        \centering
        \includegraphics[width=\linewidth]{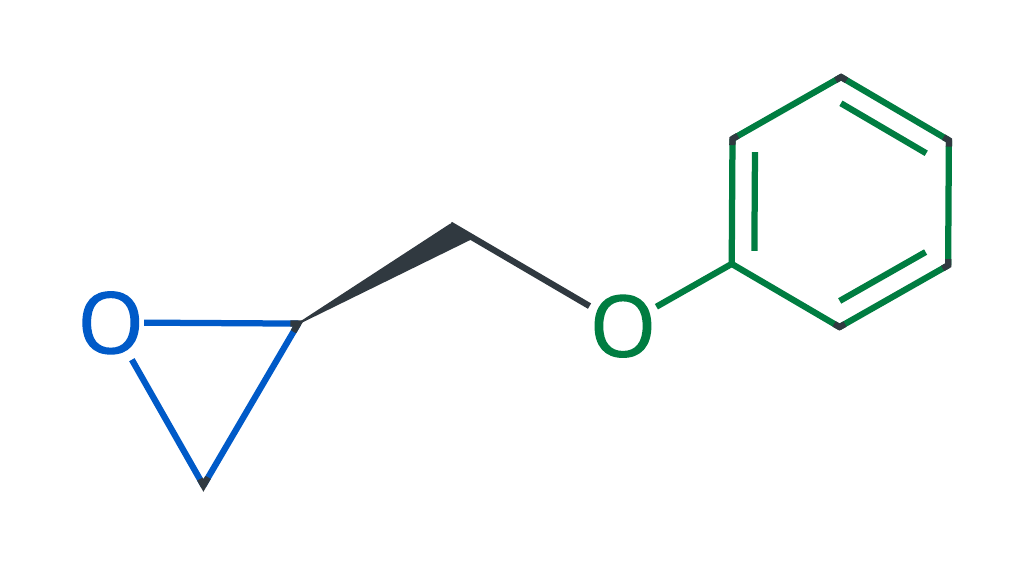}\par
        \textbf{Mutagenic (+)}
    \end{minipage}\hfill
    \begin{minipage}[c]{0.06\linewidth}
        \centering $\rightarrow$\par\vspace{10.5pt}
    \end{minipage}\hfill
    \begin{minipage}[c]{0.455\linewidth}
        \centering
        \includegraphics[width=\linewidth]{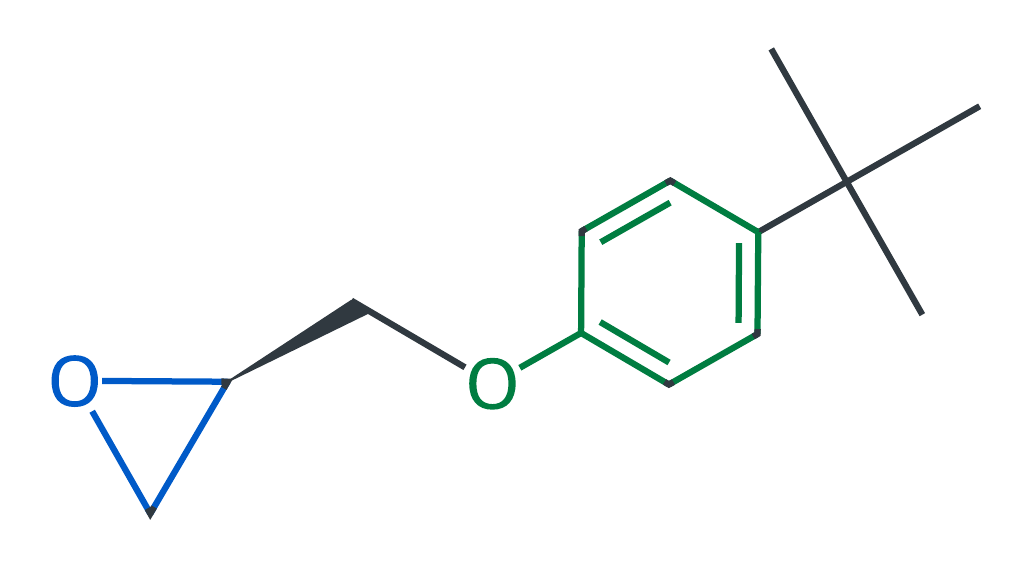}\par
        \textbf{Mutagenic (+)}
    \end{minipage}
\end{minipage}}
\newsavebox{\amesfirsthypothesis}
\sbox{\amesfirsthypothesis}{%
\begin{minipage}[t]{0.23\linewidth}
    \vspace{0pt}\vspace{2pt}\raggedright
    The shared \amcore{aryl ether} \amep{epoxide} motif, \ldots{} is the most plausible mutagenicity driver because it is retained in both compounds.  %
\end{minipage}}
\newsavebox{\amessecondobservations}
\sbox{\amessecondobservations}{%
\begin{minipage}[t]{0.38\linewidth}
    \vspace{0pt}\vspace{2pt}
    \begin{minipage}[c]{0.455\linewidth}
        \centering
        \includegraphics[width=\linewidth]{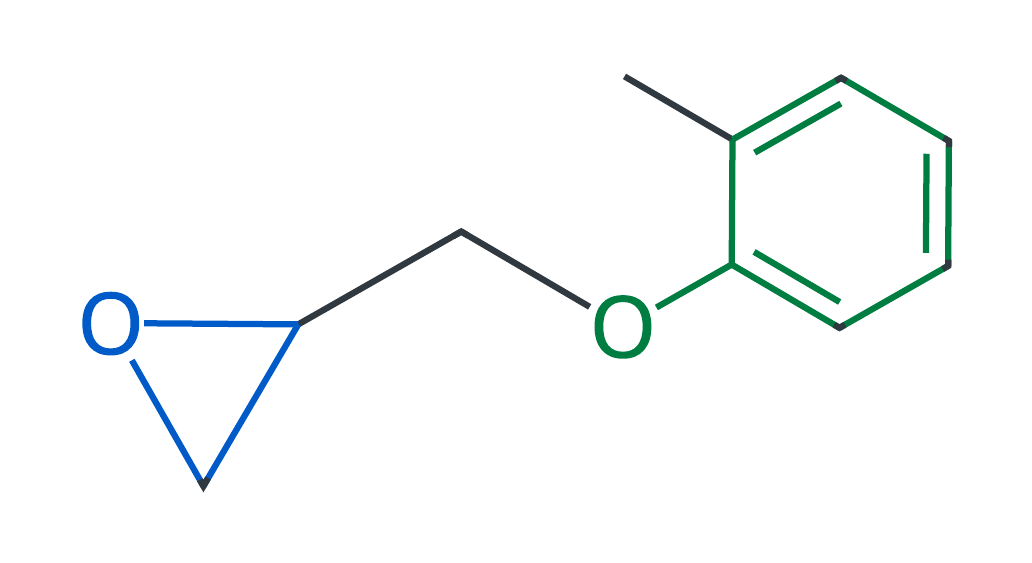}\par
        \textbf{Mutagenic (+)}
    \end{minipage}\hfill
    \begin{minipage}[c]{0.06\linewidth}
        \centering $\rightarrow$\par\vspace{10.5pt}
    \end{minipage}\hfill
    \begin{minipage}[c]{0.455\linewidth}
        \centering
        \includegraphics[width=\linewidth]{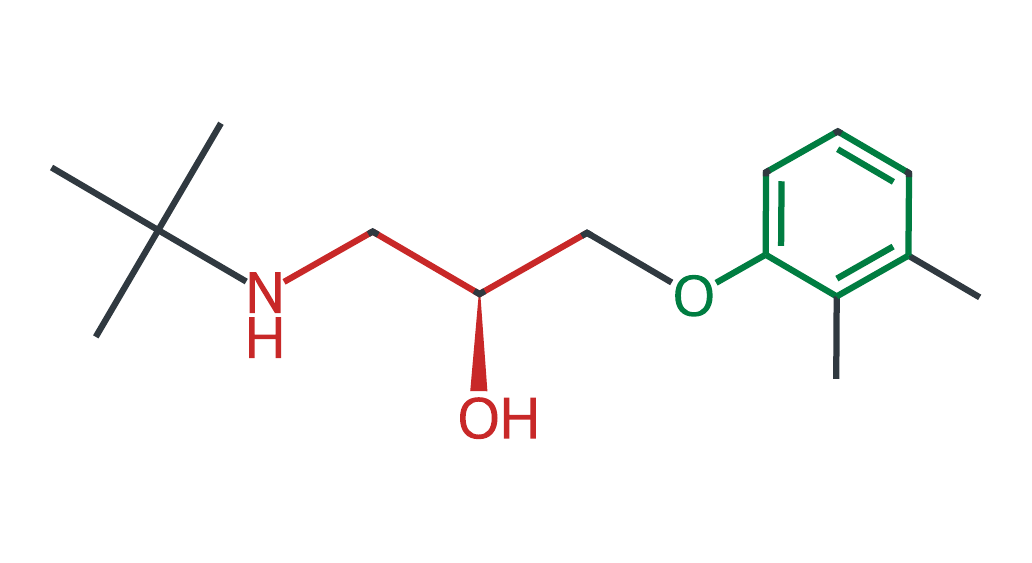}\par
        \makebox[\linewidth][c]{\textbf{Non-mutagenic ($-$)}}
    \end{minipage}
\end{minipage}}
\newsavebox{\amessecondhypothesis}
\sbox{\amessecondhypothesis}{%
\begin{minipage}[t]{0.23\linewidth}
    \vspace{0pt}\vspace{2pt}\raggedright
    mutagenicity is driven by its \amep{epoxide/oxirane-like ring} \ldots{} has \textcolor{amesaminoalcohol}{\textbf{open-chain alcohol/amine side chain}}, which is far less prone to covalent DNA damage \ldots{}
\end{minipage}}
\newlength{\amesfirstrowheight}
\newlength{\amessecondrowheight}
\setlength{\amesfirstrowheight}{\dimexpr\ht\amesfirstobservations+\dp\amesfirstobservations\relax}
\ifdim\dimexpr\ht\amesfirsthypothesis+\dp\amesfirsthypothesis\relax>\amesfirstrowheight
    \setlength{\amesfirstrowheight}{\dimexpr\ht\amesfirsthypothesis+\dp\amesfirsthypothesis\relax}
\fi
\addtolength{\amesfirstrowheight}{2.5pt}
\setlength{\amessecondrowheight}{\dimexpr\ht\amessecondobservations+\dp\amessecondobservations\relax}
\ifdim\dimexpr\ht\amessecondhypothesis+\dp\amessecondhypothesis\relax>\amessecondrowheight
    \setlength{\amessecondrowheight}{\dimexpr\ht\amessecondhypothesis+\dp\amessecondhypothesis\relax}
\fi
\ifdim\dimexpr\ht\amesprinciplebox+\dp\amesprinciplebox-\amesfirstrowheight\relax>\amessecondrowheight
    \setlength{\amessecondrowheight}{\dimexpr\ht\amesprinciplebox+\dp\amesprinciplebox-\amesfirstrowheight\relax}
\fi
\newlength{\amesrowtoptrim}
\setlength{\amesrowtoptrim}{\belowrulesep}
\addtolength{\amesrowtoptrim}{\arraystretch\ht\strutbox}
\addtolength{\amesrowtoptrim}{-2pt}
\begin{tabular}{@{}p{0.38\linewidth}@{\hspace{0.02\linewidth}}p{0.23\linewidth}@{\hspace{0.02\linewidth}}p{0.35\linewidth}@{}}
\toprule
\textbf{(A) Comparative evidence} &
\textbf{(B) Causal hypotheses} &
\textbf{(C) Inductive principle} \\
\midrule
\noalign{\vskip-\amesrowtoptrim}
\begin{minipage}[t][\amesfirstrowheight][t]{\linewidth}
    \vspace{0pt}\usebox{\amesfirstobservations}
\end{minipage}
&
\usebox{\amesfirsthypothesis}
&
\multirow[t]{2}{\linewidth}{%
\raisebox{0pt}[\ht\strutbox][0pt]{%
\usebox{\amesprinciplebox}}}
\\
\cmidrule(lr){1-2}
\noalign{\vskip-\amesrowtoptrim}
\begin{minipage}[t][\amessecondrowheight][t]{\linewidth}
    \vspace{0pt}\usebox{\amessecondobservations}
\end{minipage}
&
\usebox{\amessecondhypothesis}
& \\
\noalign{\vskip-2pt}
\bottomrule
\end{tabular}
\endgroup
\vspace{-0.1in}

\end{table*}

\paragraph{Effect of Principle Construction and Hypothesis Revision}
To see the effect of principle construction and hypothesis revision, we ablate each component separately from \textsc{Principia}. Fig.~\ref{fig:principle_refinement_ablation} shows that removing principle construction reduces performance, suggesting that generalizing individual hypotheses into principles provides critical information for subsequent discovery. Removing hypothesis revision also lowers performance, highlighting the effectiveness of reassessing existing hypotheses and reconstructing principles from the updated explanations. These results demonstrate the importance of both constructing principles from individual hypotheses and revising hypotheses based on new observations to provide up-to-date guidance for test-time scientific discovery.
\vspace{-0.06in}
\paragraph{Evaluation Budget Efficiency}
Scientific experiments can be costly, making it important to reach desired outcomes with fewer evaluations. We therefore examine whether the constructed causal principles help reduce this evaluation burden under an extended oracle budget. As shown in Fig.~\ref{fig:sample_efficiency}, \textsc{Principia} achieves the best final performance among the compared methods. Surprisingly, \textsc{Principia} matches the final performance of ExLLM, WebSearch, MolLEO, and DeepEvolve with approximately 55\%, 84\%, 86\%, and 86\% fewer oracle evaluations, respectively. These results suggest that constructing task-specific knowledge can reduce the number of evaluations required for discovery, potentially lowering the cost of experimental investigation.
\vspace{-0.06in}
\paragraph{Effect of Knowledge Retrieval}
In our framework, knowledge retrieval plays a central role: knowledge retrieval determines which hypotheses guide subsequent actions and are later revisited in light of new observations. To examine its contribution, we ablate the LLM-based selection. As shown in Fig.~\ref{fig:lclm_retrieval}, removing this lowers performance, suggesting that query-dependent selection helps identify useful knowledge for subsequent investigation, through which the selected hypotheses can be reassessed against new observations. Notably, removing LLM-based retrieval yields an AUC@10 of 0.364, comparable to 0.374 obtained without hypothesis revision in Fig.~\ref{fig:principle_refinement_ablation}, suggesting that selecting relevant knowledge is important for realizing the benefits of hypothesis revision during optimization.
\vspace{-0.06in}
\paragraph{Qualitative Analysis}
We further qualitatively analyze the constructed hypotheses and principles to examine their reasonableness and scientific plausibility. As shown in Table~\ref{tab:hypothesis_principle_qualitative}, for pairs with the same labels, the hypotheses propose a possible explanation based on the shared structures. If the labels differ, they propose structural differences that may account for the contrasting outcomes. The induced principle provides a broader view by generalizing this explanation across related comparisons, identifying recurring motifs (i.e., epoxide) while distinguishing a specific structure (i.e., aminoalcohol) that lacks a relevant property (i.e., strained electrophile). Interestingly, this principle is consistent with prior studies linking epoxide-containing compounds to mutagenicity~\citep{seiler1984mutagenicity,rosman1988mutagenicity}. This result shows how comparative hypotheses provide reasonable explanations for observed outcomes, while the induced principle captures scientifically plausible relationships across related molecules. An additional qualitative example of hypothesis revision is provided in Appendix~\ref{app:sec:supplementary_experiments}.
\section{Conclusion}
\label{sec:conclusion}

We studied the test-time construction of empirical principles for scientific discovery, focusing on how progressively accumulated observations can be transformed into knowledge that guides further investigation. To address this problem, we proposed \textsc{Principia}, an LLM-based framework that forms comparative causal hypotheses, generalizes recurring relationships into inductive principles, and uses new observations to revise the hypotheses and reconstruct the principles. Experiments spanning molecular optimization, protein engineering, gene perturbation, and molecular property prediction demonstrate that \textsc{Principia} outperforms baselines relying on internal knowledge, experience summaries, or external knowledge, with further analyses supporting its efficiency in reducing oracle evaluations and plausibility of the constructed principles. We believe our work highlights test-time hypotheses-to-principle construction as an important direction for advancing LLM-based scientific discovery, enabling accumulated observations to continually inform subsequent investigation.

\bibliography{iclr2027_conference}
\bibliographystyle{iclr2027_conference}

\appendix
\clearpage

\appendix

\begin{center}
    {\Large\bfseries Appendix}
\end{center}

\section{Implementation Details of \textsc{Principia}}
\label{app:sec:implementation_details}

\paragraph{Comparative Evidence and Hypothesis Generation}
For scientific optimization, we form reference-to-target comparisons from the inputs used in candidate generation to the resulting evaluated candidates. Each comparison provides comparative evidence containing both inputs and their scientific observations, expressed as oracle scores, from which the LLM generates comparative causal hypotheses. For molecular property prediction, we pair each newly observed target molecule with up to three reference molecules ranked by Tanimoto similarity~\citep{willett1998chemical}, excluding the target itself. The reference pool contains the molecules observed in previous rounds and those whose labels have become available in the current round. Reference selection does not filter by label, so comparisons can involve either matching or differing observations. For each reference-to-target comparison, we store the comparative evidence and its associated hypotheses as a hypothesis record $(e_{i\rightarrow j},H_{i\rightarrow j})$ in $\mathcal{M}_h$. A record may contain multiple hypotheses explaining the same comparison. Molecular inputs are represented as SMILES~\citep{weininger1988smiles}, protein inputs as amino-acid sequences, and gene inputs as HGNC\footnote{\url{https://www.genenames.org/}} gene symbols. The task-specific hypothesis-generation prompts are provided in Appendix~\ref{app:sec:prompts}.

\paragraph{Similarity-based Clustering and Principle Induction}
We group the inputs represented by nodes in $\mathcal{M}_h$ using soft-balanced $k$-medoids with domain-specific pairwise distances. For molecules, we compute Tanimoto distance from 2,048-bit Morgan fingerprints with radius 2. For protein sequences, we use normalized Hamming distance for equal-length sequences and Jaccard distance between sets of overlapping amino-acid 3-mers for unequal-length sequences. For genes, we embed the gene symbols using \texttt{text-embedding-3-large}~\citep{openai_textemb3}, normalize the embeddings to unit length, and use cosine distance. The hypothesis memory retains inputs and their associated observations, including initial inputs without hypothesis edges. In optimization, clustering uses the $N$ highest-scoring input nodes in $\mathcal{M}_h$, corresponding to the current population. In prediction, it uses all observed molecules represented in $\mathcal{M}_h$. For $n$ inputs, we set the number of clusters to $\max(1,\lfloor n/10\rfloor)$, targeting ten inputs per cluster. We use a size-balancing penalty of $0.01$, minimum and maximum cluster sizes of 5 and 15. For each cluster, we collect the comparative evidence and associated hypotheses linked to its members. An LLM then inductively infers a principle by generalizing recurring relationships across these hypotheses while accounting for conflicting observations. We store each inductive principle with its associated cluster in $\mathcal{M}_p$. The task-specific principle-induction prompts are provided in Appendix~\ref{app:sec:prompts}.

\paragraph{Knowledge Retrieval}
We independently retrieve hypothesis records from $\mathcal{M}_h$ and inductive principles from $\mathcal{M}_p$ by first constructing candidate pools and then using an LLM to select knowledge relevant to the query input and task description. For hypothesis retrieval in optimization, we rank hypothesis records by the fitness difference between the reference and target inputs. We then alternate between the largest improvements and declines, selecting up to 100 unique records. For prediction, we rank hypothesis records by the Tanimoto similarity between each record's target molecule and the query. We then select up to 50 records with matching reference and target labels and 50 with different labels. The LLM receives this candidate pool and selects relevant hypothesis records, each containing comparative evidence and its associated hypotheses. To construct the principle candidate pool, we rank clusters by their maximum member similarity to the query using the domain-specific similarity corresponding to the pairwise distance above, retain at most 100 clusters, and ask the LLM to select relevant inductive principles. The optimization configuration retrieves up to three hypothesis records and one principle per query. In prediction, the number of retrieved hypothesis records is configurable, while the LLM determines how many principles to select. The retrieved knowledge is provided as textual context for the task-specific action: candidate generation or property prediction. The task-specific retrieval prompts are provided in Appendix~\ref{app:sec:prompts}.

\paragraph{Hypothesis Revision via New Observations}
We revise the hypotheses retrieved and used during the current generation or prediction round. For each selected hypothesis record, we uniformly sample up to five new observations from that generation or round. In optimization, the sampled observations include input pairs $(x,x')$ and their observed outcomes $(\mathcal{O}(x),\mathcal{O}(x'))$. In prediction, they include query molecules $x$, their predictions $\hat{\mathcal{O}}(x)$, and their observed labels $\mathcal{O}(x)$. The LLM receives the original comparative evidence, the current hypotheses, the sampled new observations, and the task description, and can retain, revise, remove, or add hypotheses. We update the hypotheses in the corresponding records in $\mathcal{M}_h$ while retaining their original comparative evidence. The task-specific hypothesis-revision prompts are provided in Appendix~\ref{app:sec:prompts}.

\section{Task-Specific Experimental Settings}
\label{app:sec:experimental_settings}

This section describes the datasets, the application of \textsc{Principia} to each task, and the experimental configurations.

\subsection{Scientific Optimization}
\label{app:sec:optimization_tasks}

We use GPT-5-mini-2025-08-07~\citep{singh2025openai} as the backbone model for scientific optimization, with temperature set to 1.0. We report mean results over three random seeds.

\paragraph{Shared Procedure for Scientific Optimization}
\textsc{Principia}, MolLEO, ExLLM, WebSearch, and DeepEvolve share a population-based optimization architecture based on MolLEO~\citep{wang2025efficient} and the implementation of \citet{song2025evaluating}. For optimization, we maintain a population $\mathcal{P}$ consisting of the $N$ highest-scoring distinct inputs in the accumulated experimental records $\mathcal{D}$. We sample a mating pool from $\mathcal{P}$ according to normalized fitness, handling negative scores before converting them into selection probabilities. Following \citet{song2025evaluating}, the mating-pool size equals the offspring budget per generation. The task-specific operations are described below. When the LLM proposes multiple candidates, we select the first valid, previously unevaluated candidate, falling back to the first valid candidate if all proposals are duplicates. Previously evaluated inputs reuse their recorded scores. Newly evaluated inputs are added to $\mathcal{D}$, from which the population for the next generation is selected.

\paragraph{Knowledge Update Schedule of \textsc{Principia}}
\textsc{Principia} starts optimization with empty hypothesis and principle memories, so the first generation proceeds without retrieved guidance. After candidate evaluation, we generate comparative causal hypotheses from the reference-to-target comparisons and their observed outcomes. From the second generation onward, we also revise the hypotheses retrieved and used during that generation using its new observations. We then reconstruct inductive principles from the updated hypotheses and population, making the resulting memories available for candidate generation in the next generation.

\subsubsection{Molecular Optimization}
\label{app:sec:molecular_optimization_settings}

\paragraph{Datasets and Objectives}
We evaluate five tasks from the Practical Molecule Optimization (PMO) benchmark~\citep{gao2022sample}: JNK3, GSK3$\beta$, Median1, Median2, and Valsartan SMARTS. JNK3 and GSK3$\beta$ reward predicted inhibitory activity against their respective targets. Median1 and Median2 measure simultaneous similarity to two reference molecules: camphor and menthol for Median1, and tadalafil and sildenafil for Median2. Valsartan SMARTS combines a target-substructure requirement with physicochemical property objectives. We use the benchmark scoring functions as experimental oracles, with higher scores indicating better satisfaction of the objective. We adopt the task descriptions and objectives available in \citet{wang2025efficient}; for the remaining task, we follow \citet{ran2025exllm} and match the instruction style.

\paragraph{Task Execution and Application of \textsc{Principia}}
Each generation applies rule-based crossover to selected parent pairs, followed by LLM-based mutation. Before each mutation call, we use the crossover product as the query input and provide the retrieved hypotheses and principles in the mutation prompt. Both crossover products and mutated offspring are evaluated by the experimental oracle. We form reference-to-target comparisons from each parent to its crossover product and from the crossover product to its mutated offspring. Each comparison includes both inputs and their observed outcomes as comparative evidence for the generation-end knowledge update. The five tasks share this execution process, with their respective task descriptions and oracles determining the optimization objective. The prompts are provided in Appendix~\ref{app:sec:prompts_molecule_optimization}.

\paragraph{Experimental Configuration and Evaluation}
Each run starts from a pre-sampled population of 200 molecules drawn randomly from the ZINC~\citep{irwin2005zinc} pool, shared across corresponding methods. We generate up to 100 offspring per generation and use a budget of 1,000 distinct oracle evaluations, including initialization. We report AUC@10, the area under the trajectory of the mean score of the ten best molecules observed so far, following \citet{gao2022sample,wang2025efficient}. We evaluate this trajectory every 100 oracle calls, apply trapezoidal integration, and normalize by the evaluation budget. We also report Avg@100, the mean oracle score of the 100 highest-scoring evaluated molecules.

\subsubsection{Protein Sequence Engineering}
\label{app:sec:protein_engineering_settings}

\paragraph{Datasets and Objectives}
We evaluate Syn-3bfo, AAV, and GB1 using the task-specific fitness oracles provided by the protein optimization benchmark~\citep{wang2025large}. Syn-3bfo uses a synthetic fitness landscape derived from a Potts model, AAV uses a pretrained convolutional predictor of capsid fitness, and GB1 uses experimentally measured fitness values through a sequence lookup table. We follow the corresponding oracle implementations for score normalization. Each task maximizes its fitness objective over amino-acid sequences of the task-specific length using the 20 standard amino acids. We construct task descriptions from these objectives, as the original benchmark does not provide task-specific optimization prompts.

\paragraph{Task Execution and Application of \textsc{Principia}}
We apply length-preserving uniform crossover to selected parent sequences, followed by LLM-based mutation under the task-specific sequence constraints. The crossover sequence serves as the query input, and the retrieved hypotheses and principles are supplied to the mutation prompt. As in molecular optimization, we retain both the intermediate crossover sequences and the mutated offspring with their observed outcomes. We form reference-to-target comparisons from each parent to its crossover sequence and from the crossover sequence to its mutated offspring, providing comparative evidence for the generation-end knowledge update. Syn-3bfo, AAV, and GB1 share this process while using their respective task descriptions, sequence constraints, and fitness oracles. The prompts are provided in Appendix~\ref{app:sec:prompts_protein_engineering}.

\paragraph{Experimental Configuration and Evaluation}
Each run starts from 48 pre-sampled sequences, including the wild-type sequence, and generates up to 48 offspring per generation. Corresponding methods share the initial population, and the budget is 200 distinct oracle evaluations, including initialization. We compute AUC@10 from the mean fitness of the ten best sequences observed so far, using an evaluation frequency of 100 oracle calls and trapezoidal integration normalized by the evaluation budget. For Syn-3bfo, fitness values are normalized to $[0,1]$ before computing AUC, using a common normalization across methods. We also report Avg@50, the mean fitness of the 50 highest-scoring evaluated sequences.

\vspace{-0.02in}
\subsubsection{Gene Perturbation}
\label{app:sec:gene_perturbation_settings}

\vspace{-0.02in}
\paragraph{Datasets and Objectives}
We evaluate Scharenberg22 and IFNG using the CRISPR-screen datasets and task definitions adopted by BioDiscoveryAgent~\citep{roohani2025biodiscoveryagent}. Inputs are gene symbols, and the oracle returns a score derived from the recorded perturbation effect. Scharenberg22 concerns lysosomal choline recycling~\citep{scharenberg2023spns1}, and we maximize its recorded screen score. IFNG concerns regulation of interferon-$\gamma$ production in human T cells, and we maximize the absolute recorded effect, rewarding both strong increases and strong decreases. IFNG has a larger gene search space than Scharenberg22. We reuse the benchmark task descriptions.

\vspace{-0.02in}
\paragraph{Task Execution and Application of \textsc{Principia}}
Given a pair of selected genes and their scores, the LLM directly proposes another gene to evaluate, without a separate crossover operation. We use the selected genes as query inputs to retrieve hypotheses and principles for the proposal prompt. Previously proposed genes are also included to reduce duplicate proposals. The proposed gene is evaluated using its recorded perturbation effect. We form a reference-to-target comparison from each selected gene to the proposed gene, retaining both inputs and their observed outcomes as comparative evidence for the generation-end knowledge update. Scharenberg22 and IFNG share this process, with their respective objectives and gene search spaces. The prompts are provided in Appendix~\ref{app:sec:prompts_gene_perturbation}.

\vspace{-0.02in}
\paragraph{Experimental Configuration and Evaluation}
We use pre-sampled initial populations of 32 genes for Scharenberg22 and 128 for IFNG, with matching offspring budgets. The oracle budgets are 192 and 768 distinct genes, respectively, corresponding to 160 and 640 newly proposed genes beyond initialization, following \citet{roohani2025biodiscoveryagent}. We compute AUC@10 from the mean score of the ten best genes observed so far, after normalizing scores to $[0,1]$ using a common normalization across methods for each dataset. The trajectory is evaluated every 100 oracle calls and integrated with the trapezoidal rule, then normalized by the evaluation budget. We also report Hit Ratio, the fraction of benchmark hit genes recovered among the distinct newly proposed genes. Hit Ratio excludes the initial population and measures recall rather than the fraction of proposed genes that are hits. We report the unfiltered hit sets in Table~\ref{tab:main_protein_gene} and an additional evaluation excluding essential genes in Table~\ref{tab:gene_perturbation_additional_metric}. Previously proposed genes are supplied to all LLM-based baselines to reduce duplicate proposals; the additional PiFlow-specific handling is described in Appendix~\ref{app:sec:optimization_baselines}.

\vspace{-0.02in}
\subsection{Molecular Property Prediction}
\label{app:sec:prediction_settings}

\vspace{-0.02in}
We use gpt-5.4-mini-2026-03-17 as the backbone model for molecular property prediction, with temperature set to 1.0. We report mean results over three random seeds.

\vspace{-0.02in}
\paragraph{Datasets and Objectives}
We evaluate eight binary classification tasks from Therapeutics Data Commons~\citep{huang2021therapeutics}. The toxicity-related tasks are DILI, hERG, Ames, and Skin Reaction, which concern liver injury, hERG channel blockade, mutagenicity, and skin reaction, respectively. The remaining tasks predict inhibition of CYP1A2, CYP2C9, CYP2D6, and CYP3A4. Each task predicts a molecule's label from its SMILES and task description.

\vspace{-0.02in}
\paragraph{Task Execution and Application of \textsc{Principia}}
We construct knowledge over successive rounds in which new molecular observations become available. The first round reveals labels to initialize the hypothesis and principle memories, without prediction or revision. In each subsequent round, we retrieve hypotheses and principles for each newly introduced query molecule and provide them as context for label prediction, keeping that molecule's observed label hidden. After the round's predictions are complete, we reveal the labels and revise the hypotheses used during prediction using the resulting new observations. We then generate comparative causal hypotheses for the newly labeled molecules and reconstruct inductive principles before proceeding to the next round. After the final round, the memories are fixed and used for held-out test prediction without further updates. All eight datasets follow this schedule with their respective task descriptions and labels. The prompts are provided in Appendix~\ref{app:sec:prompts_molecular_property_prediction}.

\vspace{-0.02in}
\paragraph{Experimental Configuration and Evaluation}
We prepare training, validation, and test partitions using scaffold splitting and randomly sample up to 200 training molecules and 50 test molecules per class for each run. The training molecules are divided into six rounds, with each molecule appearing once. We retrieve up to three hypothesis records per prediction. The validation partition is not used in sequential knowledge construction, and held-out test labels are excluded from hypothesis generation, revision, and principle construction. We evaluate the fixed memories using accuracy and macro-F1, the unweighted mean of the class-specific F1 scores. Responses without a valid task label are treated as incorrect predictions and remain in the evaluation set.

\vspace{-0.02in}
\section{Baselines}
\label{app:sec:baselines}

\vspace{-0.02in}
\subsection{Scientific Optimization}
\label{app:sec:optimization_baselines}
\vspace{-0.02in}
\paragraph{REINVENT} We restrict our evaluation of the reinforcement-learning-based REINVENT algorithm to the pre-implemented code from the open-source official PMO~\citep{gao2022sample} repository\footnote{\url{https://github.com/wenhao-gao/mol\_opt.git}}. We adopt the default hyperparameters listed in the official code repository. Specifically, the agent policy is optimized with a learning rate of 0.0005 and a batch size of 64 molecules sampled per update. The reward-scaling coefficient sigma is set to 500, and the experience-replay parameter is set to 24, specifying the number of previously seen high-scoring molecules that are re-sampled from a prioritized memory and added to each update to reinforce promising candidates. We set the maximum number of oracle calls to 1000 and the patience to 10, with patience threshold of 1e-3, and AUC computation frequency value to 100, matching the experimental setting of \textsc{Principia} and its LLM-based discovery baselines.

\vspace{-0.04in}
\paragraph{Augmented Memory} We additionally evaluate optimization algorithms with Augmented Memory, a reinforced extension of REINVENT that introduces memory augmentation. Our implementation follows the official PMO repository. Consistent with the REINVENT protocol, we cap the maximum number of oracle calls at 1,000 and use a patience of 10 with a threshold of 1e-3, evaluating AUC every 100 oracle calls. As with REINVENT, we retain the default hyperparameters hard-coded in the official repository: a batch size of 64, sigma of 500, replay-buffer size of 1,000, learning rate of 5e-4, and 2 augmentation rounds per augmented-memory update.

\vspace{-0.04in}
\paragraph{Graph GA} Likewise, as with other machine-learning-based optimization algorithms, we follow the implementation details provided in the official PMO repository. We set the maximum number of oracle calls to 1,000, the patience to 10, with patience threshold of 1e-3, and the AUC evaluation frequency to 100, consistent with the settings used by other machine-learning-based molecular optimization methods. Since Graph GA is a population-based evolutionary search method, we use the same pre-sampled initial population as \textsc{Principia} and the other baselines, consisting of 200 samples. We additionally set the offspring population size to 100 to align its hyperparameter settings with those used for \textsc{Principia}. For the mutation rate, we follow the default configuration in the official repository and set it to 0.067.

\vspace{-0.04in}
\paragraph{MolLEO}
We reimplement MolLEO based on its original implementation~\citep{wang2025efficient}\footnote{\url{https://github.com/zoom-wang112358/MOLLEO}} and the implementation of \citet{song2025evaluating}\footnote{\url{https://github.com/HowieHwong/sde-harness}}. It provides the shared optimization architecture described in Appendix~\ref{app:sec:optimization_tasks}, using crossover followed by LLM mutation for molecules and protein sequences and direct LLM-based proposals for gene perturbation. MolLEO uses this architecture without the hypothesis and principle guidance supplied by \textsc{Principia}. Its oracle budget, population settings, patience, and convergence threshold are matched to \textsc{Principia}.

\vspace{-0.04in}
\paragraph{ExLLM} Following the experience-guided memory mechanism of ExLLM~\citep{ran2025exllm} implemented in the official repository\footnote{\url{https://github.com/HuskyNian/ExLLM}}, our textual-memory baseline augments the MolLEO baseline with an experience summarization mechanism. Once per generation, after more than $100$ candidates have been evaluated, an LLM analyzes $10$ sampled high-performing and $10$ low-performing candidates from top-100 candidates based on scalar fitness to distill a rolling, $200$-word natural-language experience summary. The newly generated summary is integrated with the previous summary and injected into each generation-level LLM call with probability $0.5$, using ExLLM's experience-injection wrapper and analysis instructions. Since ExLLM's base implementation targets a multi-objective optimization setting, whereas our evaluation optimizes a single scalar objective, we adapt its candidate-summarization procedure to account for this mismatch in the optimization setting. Specifically, we describe each candidate using its scalar fitness value together with a task-specific description prompt, rather than an explicit listing of multiple objective values. All remaining optimization hyperparameters are kept identical to those used by \textsc{Principia}.

\vspace{-0.04in}
\paragraph{PiFlow}
Since no official implementation is publicly available, we reimplement PiFlow~\citep{pu2025piflow} following the original paper. We retain its Min-Max principle scoring, score-based Refine/Validate/Explore procedure, and round-robin group-chat protocol, using the default hyperparameters $\lambda=0.5$, $K=3$, and $M=10$ without tuning. We minimally adapt the Planner and Hypothesis prompts to our experimental settings, replace the Experiment agent with a deterministic oracle, and use the same backbone LLM as \textsc{Principia}. We use \texttt{text-embedding-3-large}~\citep{openai_textemb3} for semantic novelty and initialize three principles using the paper's cold-start procedure, due to the absence of pre-configured principles in our experimental setting. At each LLM call, we generate three candidates and evaluate the first valid candidate without explicit duplicate checking, matching original paper's intent for generating single candidate based on single hypothesis. For molecular optimization and protein sequence engineering, we follow the paper's 24-iteration short-horizon protocol, including the cold-start iteration, and evaluate top-$k$ fitness and AUC over de-duplicated candidates. For gene perturbation, we instead match the oracle-call budgets of \textsc{Principia} and other baselines ($768$ and $192$) and compute the metric (hit-ratio) over the resulting distinct candidates.

\vspace{-0.04in}
\paragraph{DeepEvolve} We adapt DeepEvolve's deep-research agent~\citep{liu2025scientific} and official implementation\footnote{\url{https://github.com/liugangcode/deepevolve.git}} as a retrieval-augmented baseline. Since candidate generation is already handled by our MolLEO architecture, we retain only DeepEvolve's query-planning and web-search summarization components, injecting the resulting summaries into the design prompts. For each selected parent candidate (or pair), the planner generates $K=4$ queries based on the task objective, candidate(s) and oracle scores, evolutionary history, and top-$N=5$ scoring candidates; the summarizer executes the queries and concatenates the results, with exclusive access to the search tool. We preserve the original prompt structure while replacing its generation-based progress signal with the fraction of the oracle-call budget consumed and retaining the $0.5$ early/mature threshold. We reconstruct candidate ancestry chronologically and use the top-$N$ candidates as $(\mathrm{candidate},\mathrm{score})$ inspiration pairs. To limit information leakage, queries are restricted to general scientific principles rather than specific candidate solutions. The planner and summarizer use fixed models independent of the optimization backbone, while GPT-5-mini~\citep{singh2025openai} is used as the optimization backbone, with all other optimizer hyperparameters matched to \textsc{Principia}.

\vspace{-0.04in}
\paragraph{WebSearch} As a web-based retrieval baseline, we use Google Search via the Serper API\footnote{\url{https://serper.dev/}}. For each parent candidate (or parent pair) selected for an LLM operation, a planner generates a search query conditioned on the target property, task objective, parent candidate(s) and oracle scores, and current optimization progress. The query is issued to the search API, and the top-$k$ retrieved passages are directly injected into the design prompt without LLM-based summarization. We adapt the query-planning prompt from our DeepEvolve implementation by removing the evolutionary-history and inspiration blocks and instructing the planner to generate concise, keyword-style queries of 5--8 words for general scientific search. To align with \textsc{Principia}, we fix the number of injected passages to $4$, measure progress as the fraction of the oracle-call budget consumed, and use GPT-5-mini~\citep{singh2025openai} as the fixed planner. The planner is constrained to retrieve general scientific principles rather than information about specific candidates to reduce potential information leakage and align with our research objectives (scientific principles), while all other optimization hyperparameters, including the initial population and population size, are matched to \textsc{Principia}.

\vspace{-0.04in}
\subsection{Molecular Property Prediction}
\label{app:sec:prediction_baselines}
\vspace{-0.04in}
\paragraph{Direct Prediction}
The LLM predicts the molecular property from the task description and the query molecule represented as SMILES. The prompt specifies the required answer format without providing labeled examples or additional scientific knowledge.

\vspace{-0.04in}
\paragraph{Few-Shot Prediction}
We provide $k\in\{1,3,5,7\}$ labeled training molecules as examples for each query. Examples are selected by their Tanimoto similarity to the query using 2,048-bit Morgan fingerprints with radius 2. Each example contains the molecular SMILES and its observed label, and the selected examples are included before the query in the prediction prompt.

\vspace{-0.04in}
\paragraph{Internal Principles}
We augment the prediction prompt with a fixed set of task-specific scientific principles generated by the internal knowledge of LLMs. These principles describe molecular characteristics and physicochemical factors relevant to the target property. The same principles are provided for every query within a task and remain fixed throughout evaluation, without being updated from the observed training labels.

\vspace{-0.04in}
\paragraph{External Principles}
We obtain task-specific scientific principles through web search. Given the task description, an LLM formulates search queries targeting general mechanisms, structure-activity relationships, and physicochemical determinants. We issue these queries through Google Search via the Serper API and collect result snippets together with page content from the top-ranked sources. An LLM then summarizes the retrieved material into predictive principles, with instructions to include only claims supported by the sources. Search queries target general domain knowledge and exclude specific query molecules, benchmark names, and lists of labeled compounds. The resulting principles are stored and provided as the same textual context for all query molecules within the task.

\vspace{-0.04in}
\paragraph{Textual Summary}
We construct a single textual summary by sequentially processing the batched training molecules. At each step, the LLM receives the task description, the previous summary, and the new molecular SMILES-label pairs. It updates the summary to capture label-discriminative patterns, structural alerts, recurring molecular features, and exceptions supported by the observations. After all training batches have been processed, the final summary is fixed and included in the prediction prompt for every test molecule. This baseline accumulates guidance directly in a single text without maintaining separate comparative causal hypotheses and inductive principles.

\vspace{-0.05in}
\section{Algorithms}
\label{app:sec:algorithms}
\vspace{-0.05in}
In this section, we present the algorithmic details of \textsc{Principia}. Algorithm~\ref{alg:main} summarizes the overall workflow, integrating comparative causal hypothesis generation (Section~\ref{sec:comparative_hypotheses}), inductive principle construction (Section~\ref{sec:principle_construction}), and hypothesis revision via new observations (Section~\ref{sec:hypothesis_revision}). We describe the task-specific initialization and application procedures for molecular property prediction and scientific optimization in Algorithms~\ref{alg:tcpc_pred} and~\ref{alg:tcpc_opt}, respectively, followed by the shared knowledge construction and hypothesis revision procedures in Algorithm~\ref{alg:tcpc_common}.

\vspace{-0.05in}
\begin{algorithm}[h]
\caption{\textsc{Principia}- Main Workflow}
\label{alg:main}
\begin{algorithmic}[1]
\Require Task data $\mathcal{D}_{\text{task}}$; task description $q$; experimental oracle $\mathcal{O}$; pairwise distance $d(\cdot,\cdot)$; reference count $k_{\text{ref}}$; cluster size $k_c$; retrieval counts $k_h$ (hypothesis records), $k_p$ (principles); new observations per revision $k_r$; rounds $T$; population size $N$, offspring size $M$, and oracle-call budget $B$ (optimization)
\Ensure Hypothesis memory $\mathcal{M}_h$, principle memory $\mathcal{M}_p$
\State $\mathcal{M}_h \gets \emptyset$;\; $\mathcal{M}_p \gets \emptyset$
\State $\mathcal{D}\gets\textsc{InitializeTask}(\mathcal{D}_{\text{task}},T)$
\For{$t = 1$ \textbf{to} $T$}
    \State $(\mathcal{D}, \mathcal{C}, \mathcal{B}, \mathcal{U}) \gets \textsc{TaskApplication}(\mathcal{D}, \mathcal{M}_h, \mathcal{M}_p; t)$
    \If{$t > 1$}
        \State $\mathcal{M}_h \gets \textsc{ReviseHypotheses}(\mathcal{M}_h, \mathcal{U}, \mathcal{B})$ \Comment{Section~\ref{sec:hypothesis_revision}}
    \EndIf
    \State $\mathcal{M}_h \gets \textsc{GenerateHypotheses} (\mathcal{M}_h, \mathcal{C})$ \Comment{Section \ref{sec:comparative_hypotheses}}
    \State $\mathcal{M}_p \gets \textsc{InducePrinciples}(\mathcal{M}_h)$\Comment{Section \ref{sec:principle_construction}}
    \If{task is optimization \textbf{and} $|\mathcal{D}| \geq B$}
        \State \textbf{break} \Comment{Oracle-call budget exhausted}
    \EndIf
\EndFor
\State \Return $\mathcal{M}_h, \mathcal{M}_p$
\end{algorithmic}
\end{algorithm}

\vspace{-0.05in}
\begin{algorithm*}[h]
\caption{\textsc{Principia} - Prediction-specific procedures}
\label{alg:tcpc_pred}
\begin{algorithmic}[1]
\Statex \textbf{Shared state:} round-specific data batches $\{\mathcal{Z}_t\}_{t=1}^{T}$
\Procedure{InitializeTask}{$\mathcal{D}_{\text{task}},T$}
    \State Partition $\mathcal{D}_{\text{task}}$ into $T$ batches $\{\mathcal{Z}_t\}_{t=1}^{T}$
    \State \Return $\emptyset$
\EndProcedure
\Statex
\Procedure{TaskApplication}{$\mathcal{D}, \mathcal{M}_h, \mathcal{M}_p; t$}
    \State $\mathcal{C},\mathcal{B},\mathcal{U}\gets\emptyset$
    \If{$t > 1$}
        \For{$(x,\mathcal{O}(x))\in\mathcal{Z}_t$}
            \State $\mathcal{R}_h\gets\mathrm{LLM}_{\text{select}}(x,\mathcal{M}_h,k_h;q)$ \Comment{Hypotheses retrieval}
            \State $\mathcal{R}_p\gets\mathrm{LLM}_{\text{select}}(x,\mathcal{M}_p,k_p;q)$ \Comment{Principle retrieval}
            \State $\hat{\mathcal{O}}(x)\gets\mathrm{LLM}_{\text{predict}}(x,\mathcal{R}_h,\mathcal{R}_p;q)$ \Comment{Property prediction}
            \State $\mathcal{U}\gets\mathcal{U}\cup\mathcal{R}_h$ \Comment{Used hypotheses}
            \State $\mathcal{B}\gets\mathcal{B}\cup\{(x,\mathcal{O}(x),\hat{\mathcal{O}}(x))\}$ \Comment{New observations}
        \EndFor
    \Else
        \State $\mathcal{B}\gets\{(x,\mathcal{O}(x),\bot)\mid(x,\mathcal{O}(x))\in\mathcal{Z}_1\}$ \Comment{Initial labels; no predictions}
    \EndIf
    \State $\mathcal{D}\gets\mathcal{D}\cup\{(x,\mathcal{O}(x))\mid(x,\mathcal{O}(x),\cdot)\in\mathcal{B}\}$
    \For{$(x_j,\mathcal{O}(x_j),\cdot)\in\mathcal{B}$}
        \State $\mathcal{N}_j\gets\operatorname{kNN}_d\big(x_j,\{x_i\mid(x_i,\mathcal{O}(x_i))\in\mathcal{D},\ i\neq j\},k_{\text{ref}}\big)$
        \State $\mathcal{C}\gets\mathcal{C}\cup\{e_{i\rightarrow j}\mid x_i\in\mathcal{N}_j\}$ \Comment{Comparative evidence}
    \EndFor
    \State \Return $(\mathcal{D}, \mathcal{C}, \mathcal{B}, \mathcal{U})$
\EndProcedure
\end{algorithmic}
\end{algorithm*}

\vspace{-0.02in}
\begin{algorithm*}[h]
\caption{\textsc{Principia} - Optimization-specific procedures}
\label{alg:tcpc_opt}
\begin{algorithmic}[1]
\Procedure{InitializeTask}{$\mathcal{D}_{\text{task}},T$}
    \State \Return $\mathcal{D}_{\text{task}}$ \Comment{Evaluated initial inputs}
\EndProcedure
\Statex
\Procedure{TaskApplication}{$\mathcal{D}, \mathcal{M}_h, \mathcal{M}_p; t$}
    \State $\mathcal{C},\mathcal{B},\mathcal{U}\gets\emptyset$
    \State $\mathcal{P}\gets\mathrm{Top}_N\big(\{x\mid(x,\mathcal{O}(x))\in\mathcal{D}\}\big)$ \Comment{Population selection}
    \State $\mathcal{Q}\gets\textsc{Select}(\mathcal{P}, M)$ \Comment{Parent selection}
    \For{$(x_1,x_2)\in\mathcal{Q}$}
        \If{task is gene perturbation}
            \State $\mathcal{X}\gets\{x_1,x_2\}$
        \Else
            \State $x_c\gets\textsc{Crossover}(x_1,x_2)$
            \State $\mathcal{B}\gets\mathcal{B}\cup\{(x_i,\mathcal{O}(x_i),x_c,\mathcal{O}(x_c))\mid i\in\{1,2\}\}$
            \State $\mathcal{X}\gets\{x_c\}$
        \EndIf
        \For{$x\in\mathcal{X}$}
            \State $\mathcal{R}_h\gets\mathrm{LLM}_{\text{select}}(x,\mathcal{M}_h,k_h;q)$ \Comment{Hypotheses retrieval}
            \State $\mathcal{R}_p\gets\mathrm{LLM}_{\text{select}}(x,\mathcal{M}_p,k_p;q)$ \Comment{Principle retrieval}
            \State $x'\gets\mathrm{LLM}_{\text{generate}}\big(x,\mathcal{O}(x),\mathcal{R}_h,\mathcal{R}_p;q\big)$ \Comment{Candidate generation}
            \State $\mathcal{U}\gets\mathcal{U}\cup\mathcal{R}_h$ \Comment{Used hypotheses}
            \State $\mathcal{B}\gets\mathcal{B}\cup\{(x,\mathcal{O}(x),x',\mathcal{O}(x'))\}$ \Comment{New observations}
        \EndFor
    \EndFor
    \State $\mathcal{D}\gets\mathcal{D}\cup\{(x',\mathcal{O}(x'))\mid(\cdot,\cdot,x',\mathcal{O}(x'))\in\mathcal{B}\}$
    \State $\mathcal{C}\gets\mathcal{B}$ \Comment{Comparative evidence}
    \State \Return $(\mathcal{D}, \mathcal{C}, \mathcal{B}, \mathcal{U})$
\EndProcedure
\end{algorithmic}
\end{algorithm*}

\vspace{-0.02in}
\begin{algorithm}[h]
\caption{\textsc{Principia} - Shared knowledge construction and hypothesis revision procedures}
\label{alg:tcpc_common}
\begin{algorithmic}[1]
\Procedure{ReviseHypotheses}{$\mathcal{M}_h, \mathcal{U}, \mathcal{B}$}
    \For{$(e_{i\rightarrow j}, H_{i\rightarrow j}) \in \mathcal{U}$}
        \State $H_{i\rightarrow j} \gets \mathrm{LLM}_{\text{refine}}\big(e_{i\rightarrow j}, H_{i\rightarrow j}, \textsc{Sample}(\mathcal{B}, k_r); q\big)$
    \EndFor
    \State \Return $\mathcal{M}_h$
\EndProcedure
\Statex
\Procedure{GenerateHypotheses}{$\mathcal{M}_h, \mathcal{C}$}
    \For{$e_{i\rightarrow j} \in \mathcal{C}$}
        \State Store $\big(e_{i\rightarrow j}, \mathrm{LLM}_{\text{hypothesis}}(e_{i\rightarrow j}; q)\big)$ in $\mathcal{M}_h$ as directed edge $x_i \rightarrow x_j$
    \EndFor
    \State \Return $\mathcal{M}_h$
\EndProcedure
\Statex
\Procedure{InducePrinciples}{$\mathcal{M}_h$}
    \State $\{G_k\}_{k=1}^{K} \gets \textsc{ClusterInputs}(\mathcal{M}_h;d,k_c)$
    \For{$k = 1$ \textbf{to} $K$}
        \State $\mathcal{E}_k \gets \{(e_{i\rightarrow j}, H_{i\rightarrow j}) \mid (i,j)\in\mathcal{A}_h,\ i\in G_k \lor j\in G_k\}$
        \State $P_k \gets \mathrm{LLM}_{\text{principle}}(\mathcal{E}_k;q)$
    \EndFor
    \State \Return $\mathcal{M}_p = \{(G_k, P_k)\}_{k=1}^{K}$
\EndProcedure
\end{algorithmic}
\end{algorithm}

\clearpage
\section{Supplementary Experiments}
\label{app:sec:supplementary_experiments}
\paragraph{Importance of Diversifying Learning Signals for Optimization} In Table~\ref{tab:hypothesis_retrieval_strategy}, we analyze different strategies for retrieving comparative causal hypotheses during scientific optimization. The results show that combining hypotheses associated with both increases and decreases in property fitness from reference to target leads to more effective discovery than exclusively retrieving hypotheses associated with either direction. This suggests that diversifying learning signals enables language models to learn from a broader range of comparative evidence and associated hypotheses, thereby supporting the generation of more accurate comparative causal hypotheses and inductive principles aligned with the discovery objective, validating our design choice explained in Appendix Section \ref{app:sec:implementation_details}.

Furthermore, retrieving hypotheses with consistently increasing or decreasing gains provides stronger learning signals than random retrieval, where informative signals may be diluted. Notably, \textsc{Principia} consistently outperforms the plain optimization algorithm even when using diverse hypothesis retrieval strategies. This suggests that the comparative causal hypotheses and inductive principles constructed by \textsc{Principia} remain useful across different retrieval strategies, supporting the overall quality of the constructed principles and the effectiveness of \textsc{Principia}.

\paragraph{Efficacy in Multi-Objective Discovery} While we have demonstrated the effectiveness of \textsc{Principia} across various discovery tasks, most involve single-objective optimization. To evaluate its robustness under diverse objectives, we consider a multi-objective protein sequence engineering task that maximizes the Syn-3bfo fitness score, equivalent to the normalized negative Potts-model energy on the PDB 3bfo-derived synthetic landscape, while minimizing the Hamming distance from the wild-type sequence. Following~\citet{wang2025large}, we scalarize the two objectives with equal weights. As shown in Figure \ref{fig:multi_objective}, \textsc{Principia} improves over the plain optimization algorithm, while other knowledge-augmented methods degrade over plain optimization algorithm, demonstrating the robustness of guidance from inductive principles constructed at test time across diverse discovery objectives.
\begin{figure*}[t]
\centering
\begin{minipage}[t]{0.30\textwidth} \vspace{0pt}
    \centering
    \captionof{table}{Importance of diversifying learning signals for hypothesis retrieval in Scientific Discovery on GSK3$\beta$ optimization, using \texttt{gpt-5-mini} as backbone models.}
    \vspace{-0.05in}
    \resizebox{\linewidth}{!}{
    \setlength{\tabcolsep}{1.0pt}
    \renewcommand{\arraystretch}{1.6}
    \begin{tabular}{lcc}
        \toprule
        \noalign{\vskip 0.5ex}
        \textbf{Method} & \textbf{AUC@10} $\uparrow$ & \textbf{$\Delta$} $\uparrow$ \\
        \midrule
        None & 0.341 & --- \\
        \cdashline{1-3}[0.5pt/1pt]\rule{-2pt}{2.5ex}
        $\Delta$ Random & 0.366 & \textcolor{teal}{$+\mathbf{0.025}$} \\
        $\Delta$ Worst & 0.379 & \textcolor{teal}{$+\mathbf{0.038}$} \\
        $\Delta$ Best & 0.372 & \textcolor{teal}{$+\mathbf{0.031}$} \\
        \midrule
        \textbf{$\Delta$ Mixture (Ours)} & \textbf{0.452} &  \textcolor{teal}{$+\mathbf{0.111}$} \\
        \bottomrule
    \end{tabular}
    }
    \label{tab:hypothesis_retrieval_strategy}

\end{minipage}
\hfill
\begin{minipage}[t]{0.36\textwidth} \input{Figures/Appendix/multi_objective_figure}
\end{minipage}
\hfill
\begin{minipage}[t]{0.30\textwidth}
    \vspace{0pt}
\captionof{table}{Analysis of the performance on the Scharenberg22 and IFNG tasks using an additional metric, Hit Ratio w/o Essential, using \texttt{gpt-5-mini} as backbone models.}
\vspace{-0.05in}
\label{tab:gene_perturbation_additional_metric}
\centering
\resizebox{\linewidth}{!}{%
\setlength{\tabcolsep}{2pt}
\renewcommand{\arraystretch}{1.3}
\begin{tabular}{lcc}
\toprule
\textbf{Variant}
& \textbf{Scharenberg22}
& \textbf{IFNG} \\
\midrule

MolLEO$^{*}$
& 0.265
& 0.114 \\

ExLLM$^{*}$
& 0.286
& 0.117 \\

PiFlow$^{*}$
& 0.272
& 0.118 \\

WebSearch$^{*}$
& 0.218
& 0.117 \\

DeepEvolve$^{*}$
& 0.265
& 0.113 \\

\midrule
\textbf{Ours} & \textbf{0.320} & \textbf{0.123} \\

\bottomrule
\end{tabular}
}

\end{minipage}
\end{figure*}

\paragraph{Analysis with Additional Metrics} To further demonstrate the effectiveness of \textsc{Principia} in accelerating scientific discovery through inductive principles constructed at test time, we evaluate its performance using additional metrics. In Table~\ref{tab:gene_perturbation_additional_metric}, we report the Hit Ratio after excluding essential genes from the ground-truth set, following the additional evaluation protocol adopted in \citet{roohani2025biodiscoveryagent}. \textsc{Principia} achieves the highest Hit Ratio on both the Scharenberg22 and IFNG gene perturbation tasks, outperforming the baseline optimization algorithms across both tasks.

\begin{figure}[t]
    \centering

    \begin{subfigure}[b]{0.47\columnwidth}
        \centering
        \includegraphics[width=\linewidth]{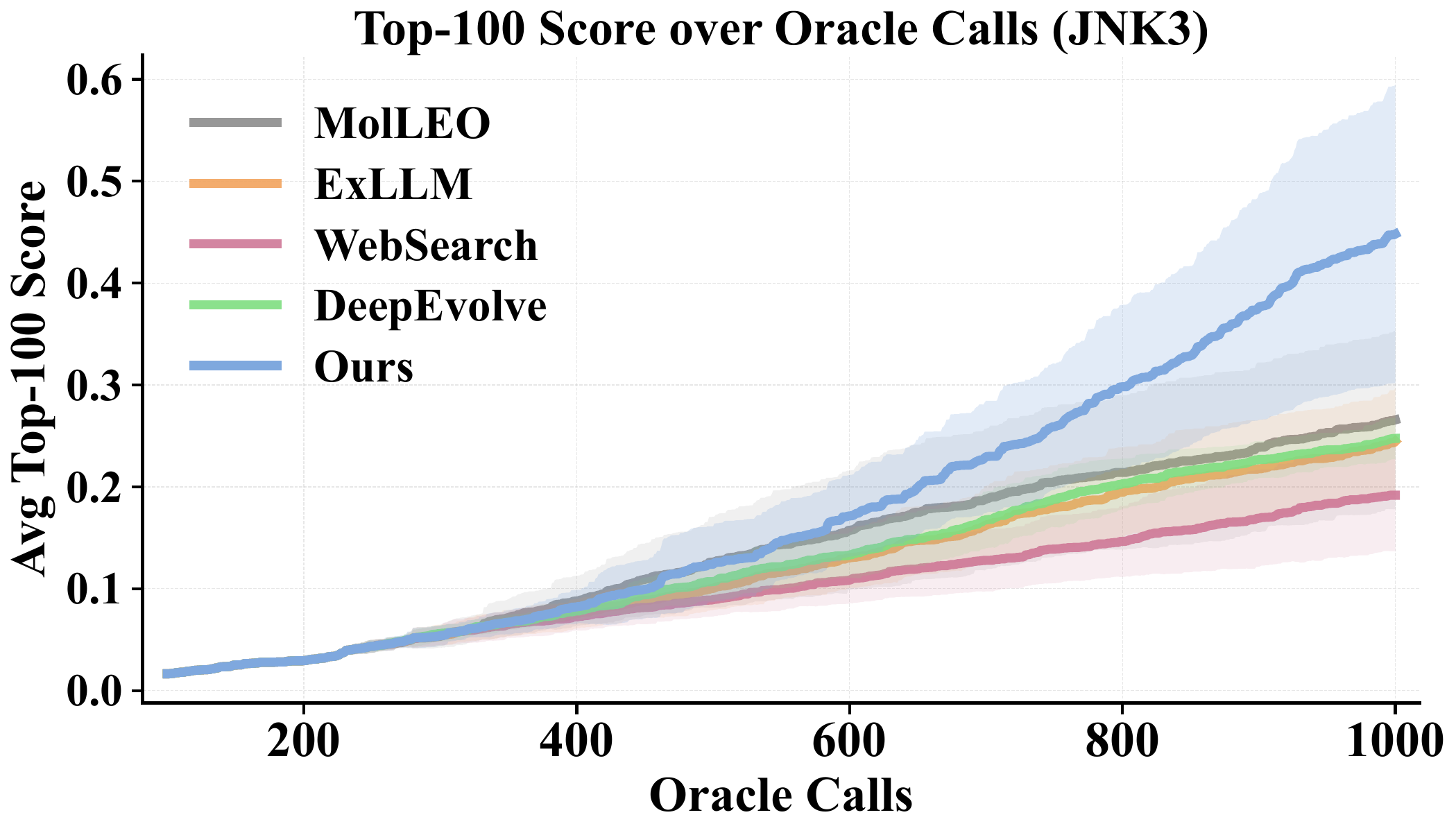}
        \label{fig:JNK3}
    \end{subfigure}
    \hfill
    \begin{subfigure}[b]{0.47\columnwidth}
        \centering
        \includegraphics[width=\linewidth]{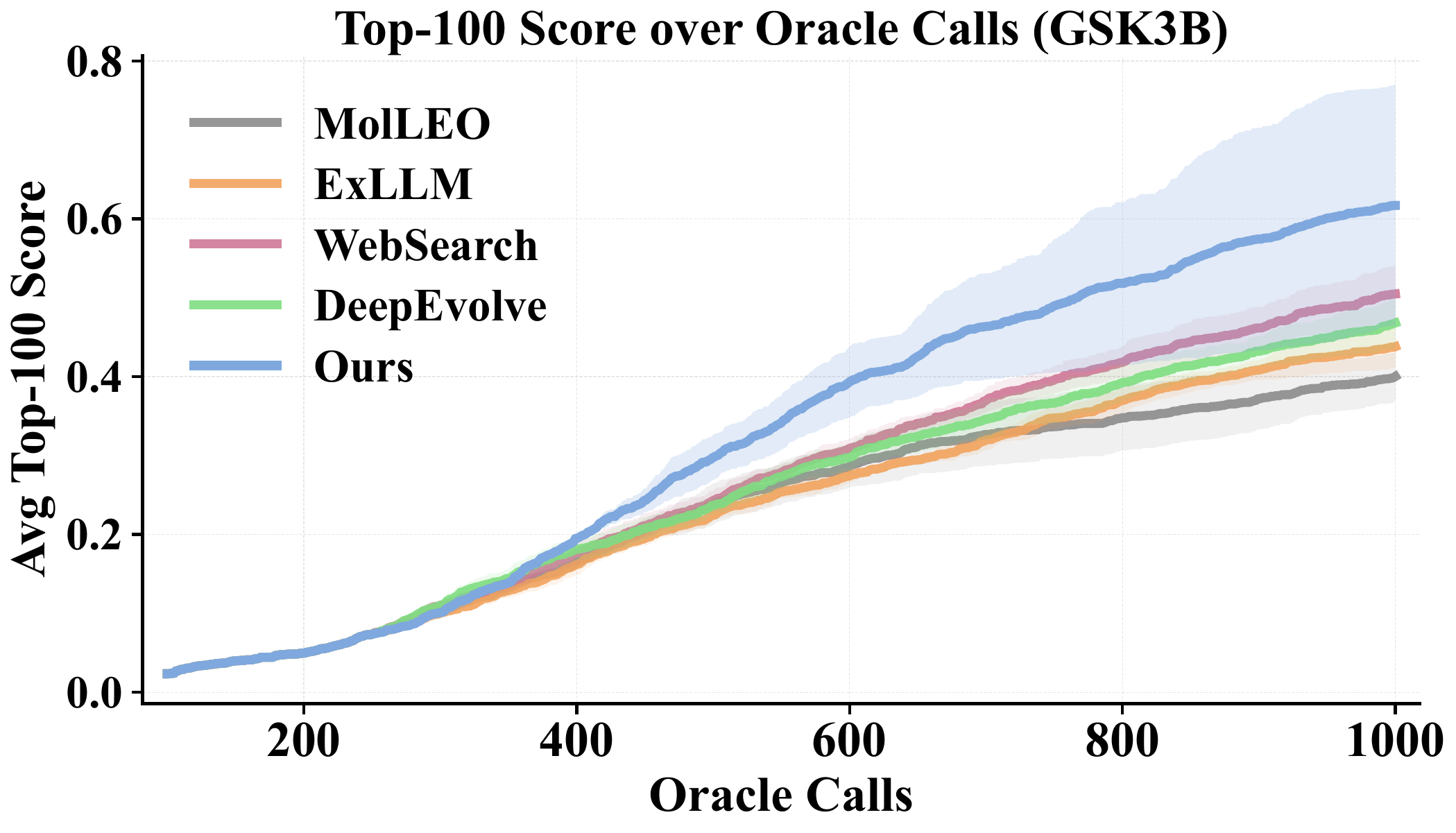}
        \label{fig:gsk3b}
    \end{subfigure}

    \vspace{-0.1in} %

    \begin{subfigure}[b]{0.47\columnwidth}
        \centering
        \includegraphics[width=\linewidth]{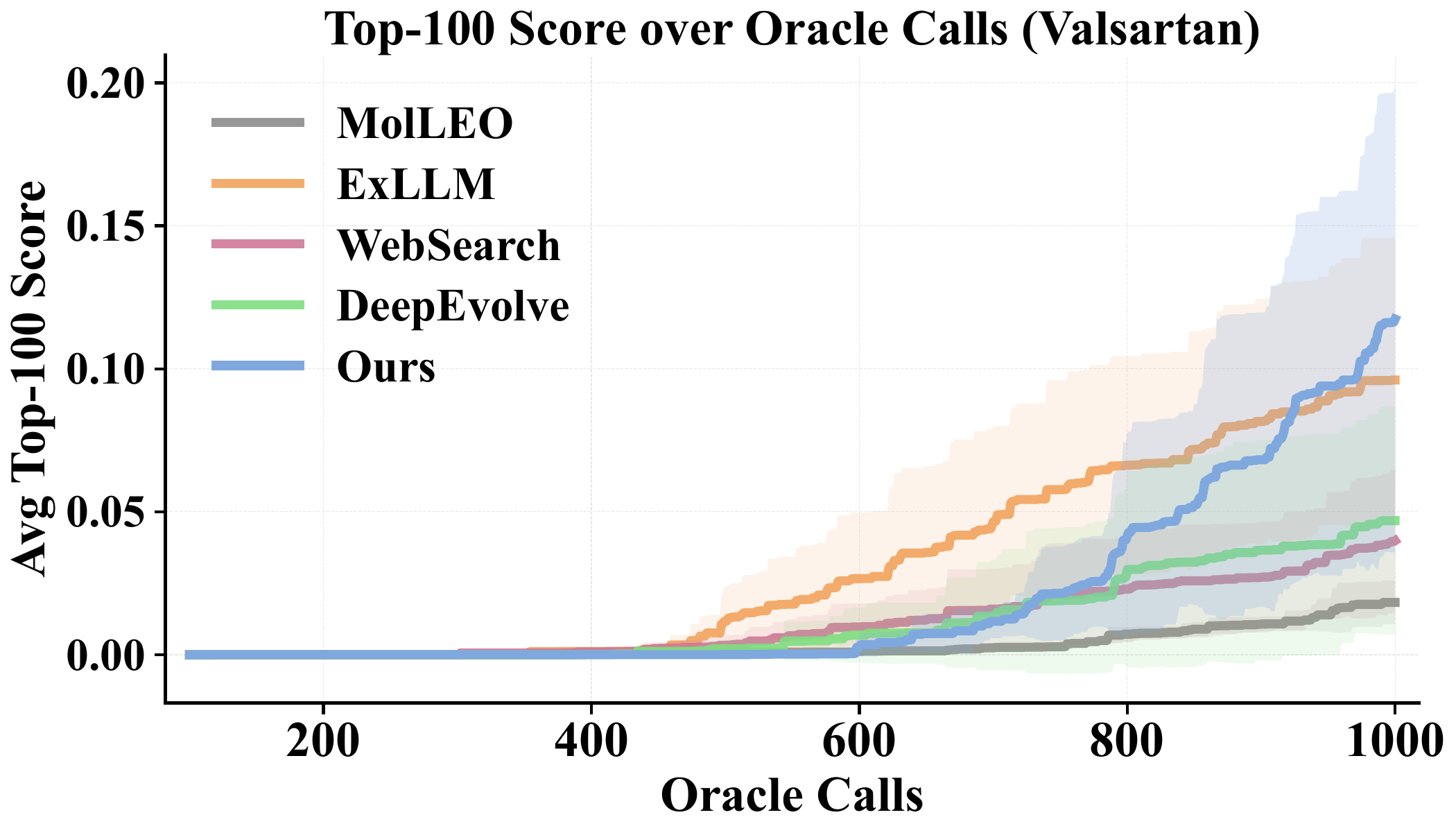}
        \label{fig:valsartan}
    \end{subfigure}
    \hfill
    \begin{subfigure}[b]{0.47\columnwidth}
        \centering
        \includegraphics[width=\linewidth]{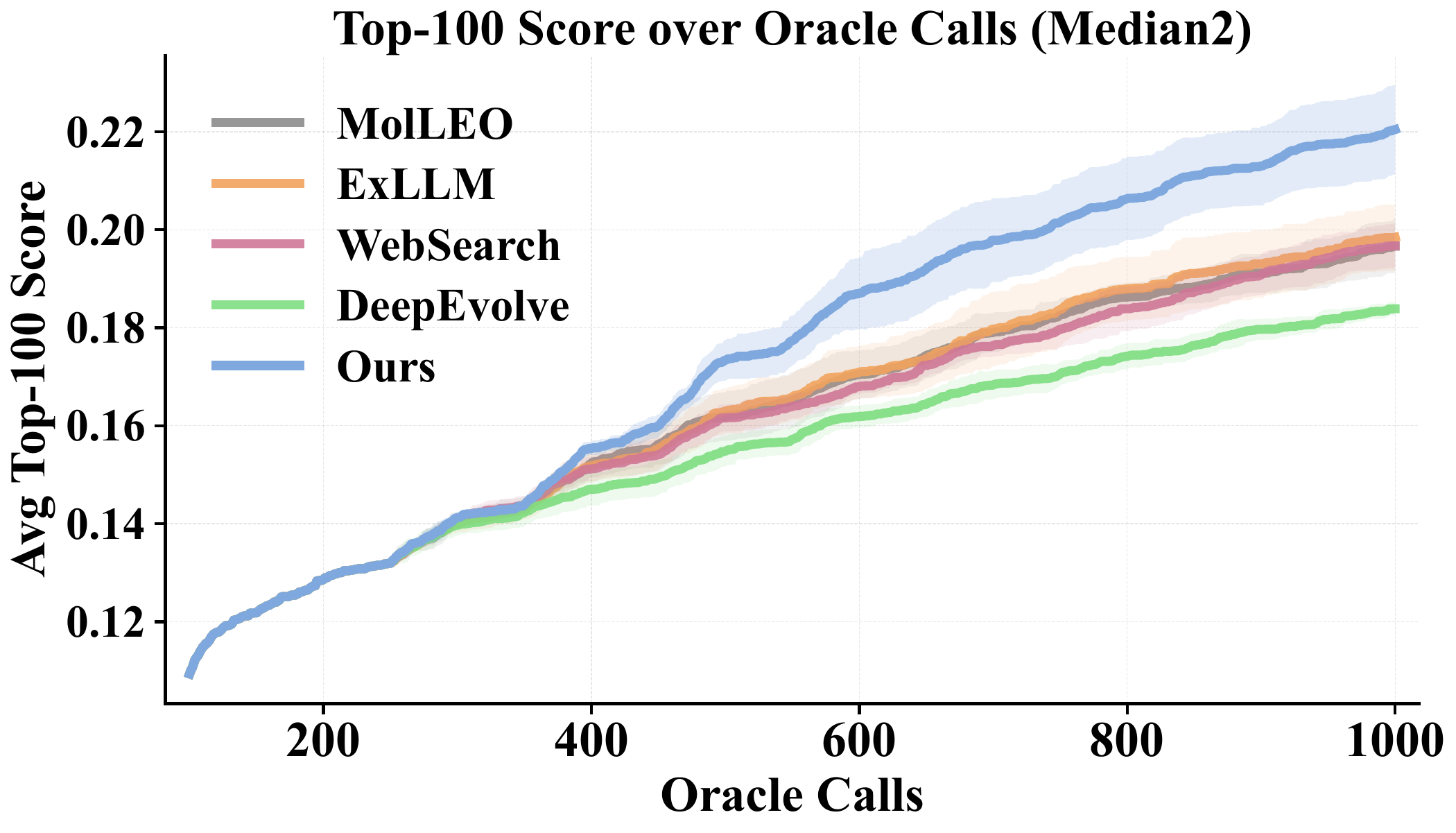}
        \label{fig:median2}
    \end{subfigure}

    \vspace{-0.1in}

    \begin{subfigure}[b]{0.47\columnwidth}
        \centering
        \includegraphics[width=\linewidth]{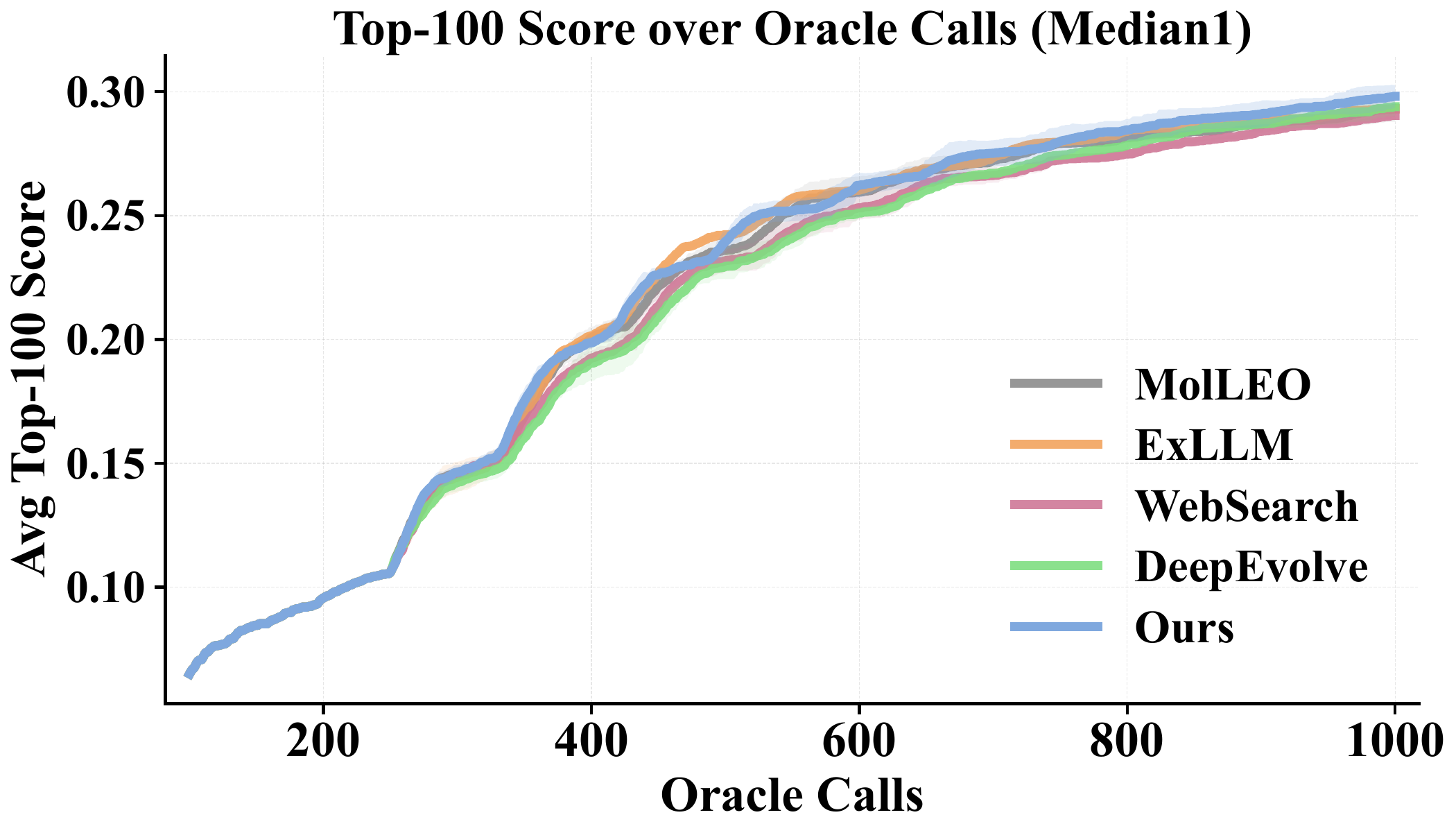}
        \label{fig:median1}
    \end{subfigure}
    \hfill
    \begin{subfigure}[b]{0.47\columnwidth}
        \centering
        \includegraphics[width=\linewidth]{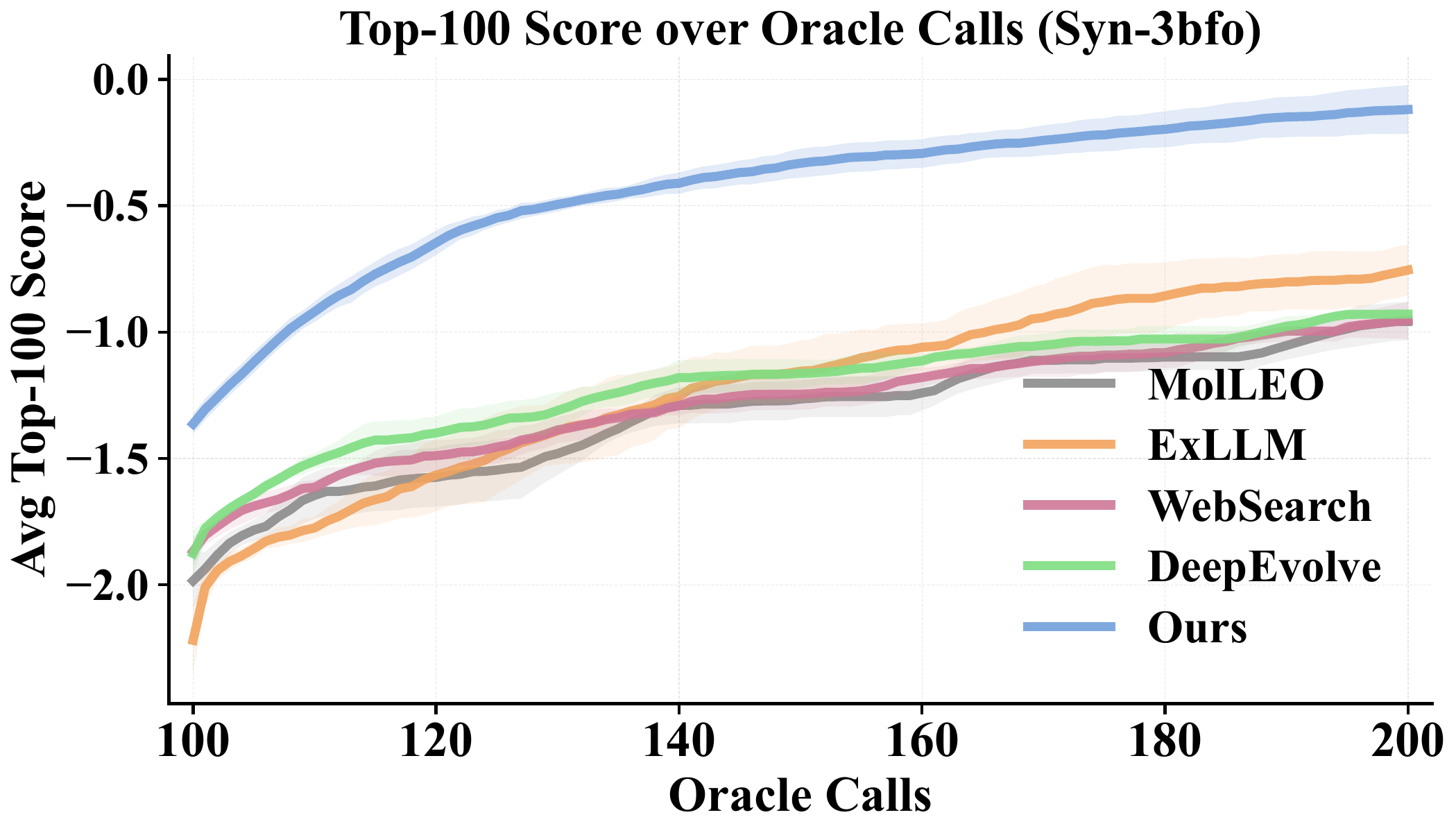}
        \label{fig:syn-3bfo}
    \end{subfigure}

    \vspace{-0.1in}

    \begin{subfigure}[b]{0.47\columnwidth}
        \centering
        \includegraphics[width=\linewidth]{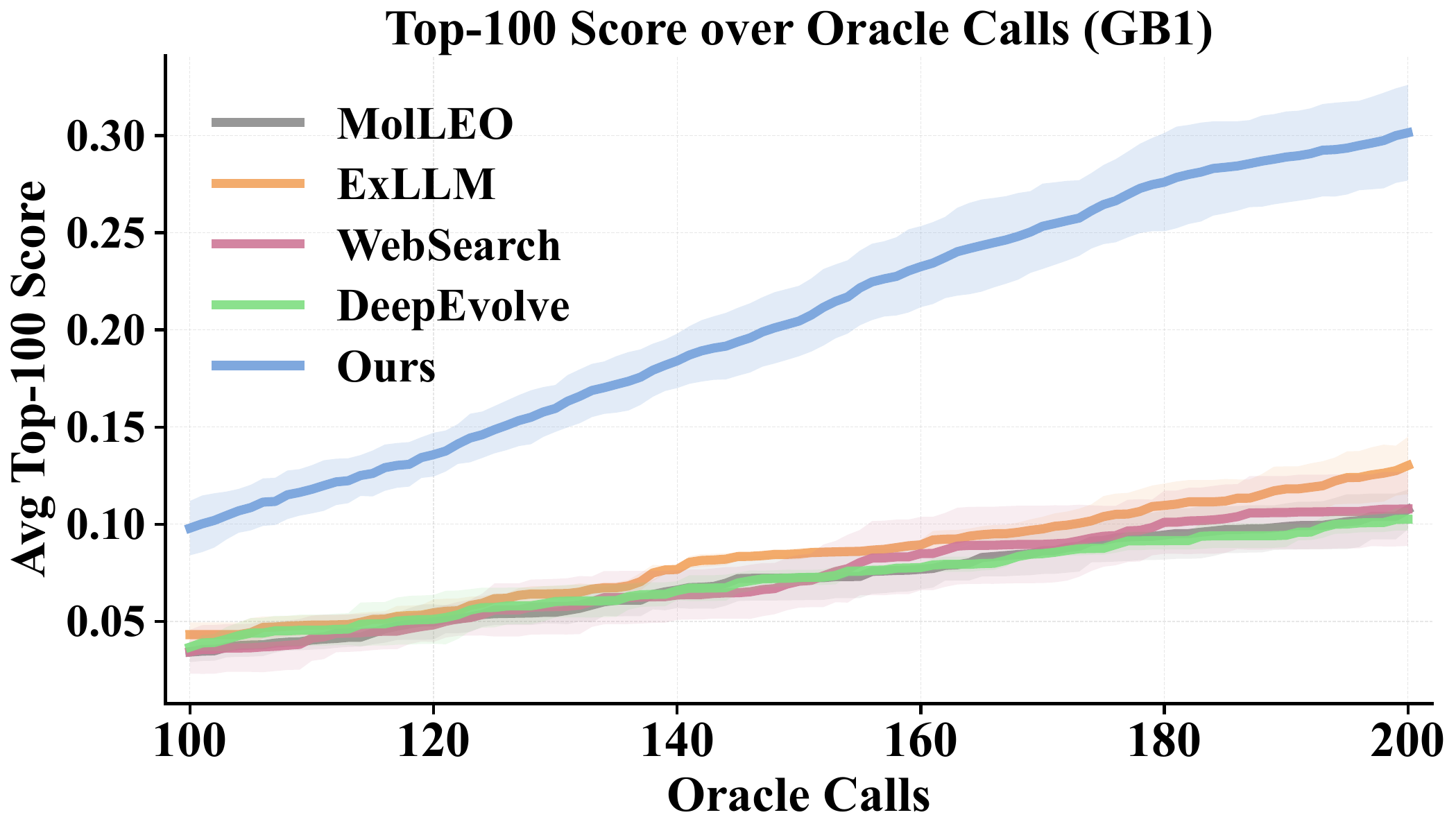}
        \label{fig:gb1}
    \end{subfigure}
    \hfill
    \begin{subfigure}[b]{0.47\columnwidth}
        \centering
        \includegraphics[width=\linewidth]{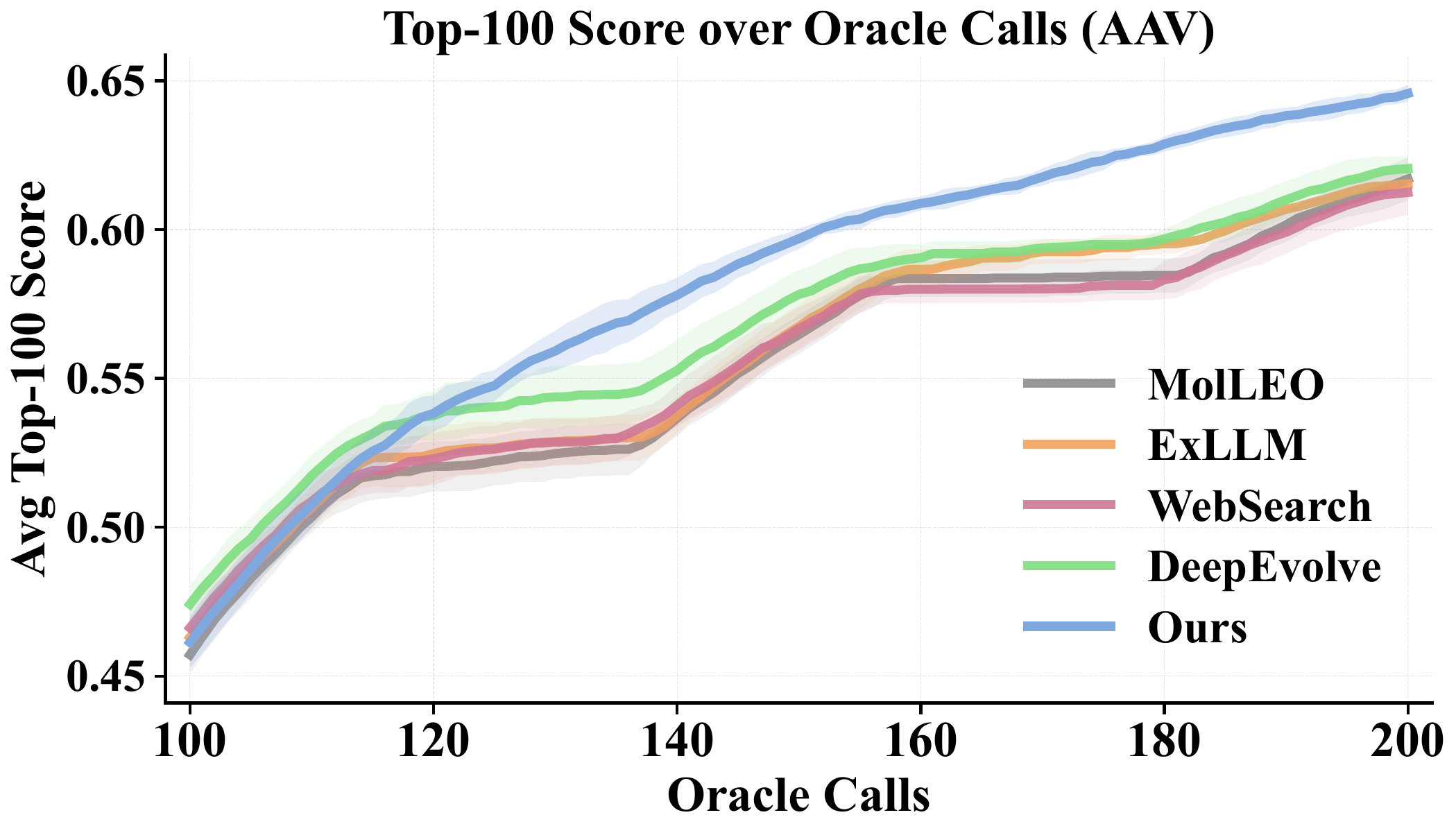}
        \label{fig:plot8}
    \end{subfigure}

    \vspace{-0.1in}

    \begin{subfigure}[b]{0.47\columnwidth}
        \centering
        \includegraphics[width=\linewidth]{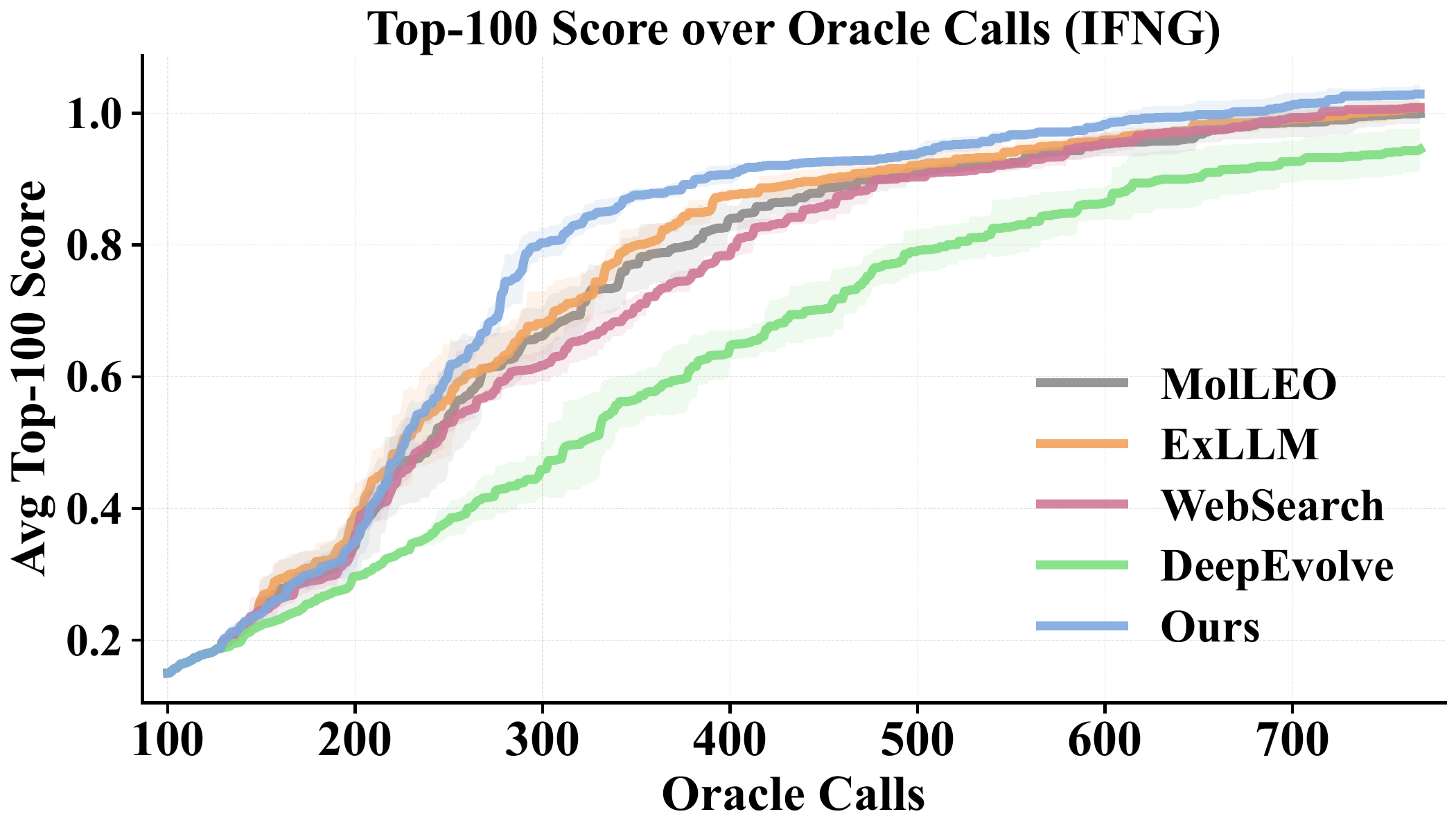}
        \label{fig:plot9}
    \end{subfigure}
    \hfill
    \hspace{0.263in}
    \begin{subfigure}[b]{0.47\columnwidth}
        \centering
        \includegraphics[width=\linewidth]{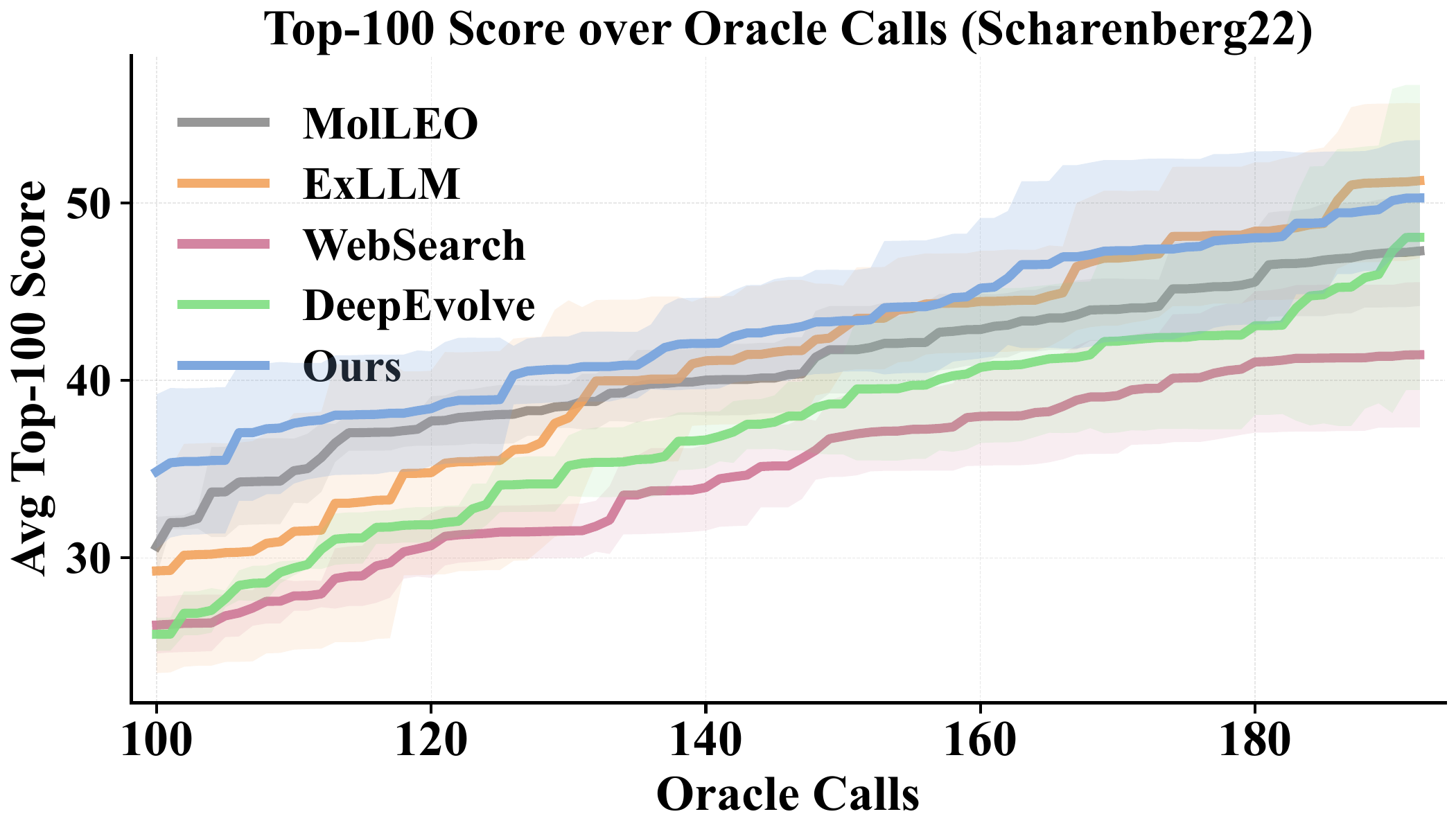}
        \label{fig:plot10}
    \end{subfigure}
    \vspace{-0.3in}
    \caption{Top-100 average score trends with respect to the number of oracle calls for molecular optimization, protein sequence engineering, and gene perturbation tasks reported in Tables~\ref{tab:main_drug_discovery} and~\ref{tab:main_protein_gene}. For each task, lines represent the mean fitness values over three random seeds, and shaded regions indicate the standard deviation.}
    \label{fig:overall_results}
\end{figure}

Furthermore, Figure~\ref{fig:overall_results} presents the Top-100 Average Score trends with respect to the number of oracle calls, complementary to the Top-10 AUC and Top-100 Average scores after a fixed amount of oracle calls. The results show that \textsc{Principia} consistently achieves higher Top-100 Average Scores than the existing optimization algorithms throughout the optimization process. These results provide further evidence that \textsc{Principia} can effectively improve the performance of scientific discovery by constructing and applying task-specific inductive principles at test time.
\begin{table*}[t]
    \centering
    \begingroup
\color{black}
\definecolor{cypamide}{RGB}{0,90,200}
\definecolor{cypthiourea}{RGB}{200,40,40}
\definecolor{cyparomatic}{RGB}{0,125,65}
\newcommand{\cypam}[1]{\textcolor{cypamide}{\textbf{#1}}}
\newcommand{\cypth}[1]{\textcolor{cypthiourea}{\textbf{#1}}}
\newcommand{\cypar}[1]{\textcolor{cyparomatic}{\textbf{#1}}}
\fontsize{8.5}{10.5}\selectfont
\setlength{\tabcolsep}{0pt}
\renewcommand{\arraystretch}{1.05}
\noindent\begin{tabular}{@{}p{0.54\linewidth}@{\hspace{0.03\linewidth}}p{0.43\linewidth}@{}}
\toprule
\makebox[\linewidth][c]{\textbf{Reference \& Target}} & \makebox[\linewidth][c]{\textbf{Original Hypothesis}} \\
\specialrule{\lightrulewidth}{\aboverulesep}{0pt}
\begin{minipage}[t]{\linewidth}
    \vspace{0pt}\vspace{-\ht\strutbox}\vspace{3pt}\centering
    \begin{minipage}[c]{0.44\linewidth}
        \centering
        \includegraphics[width=\linewidth]{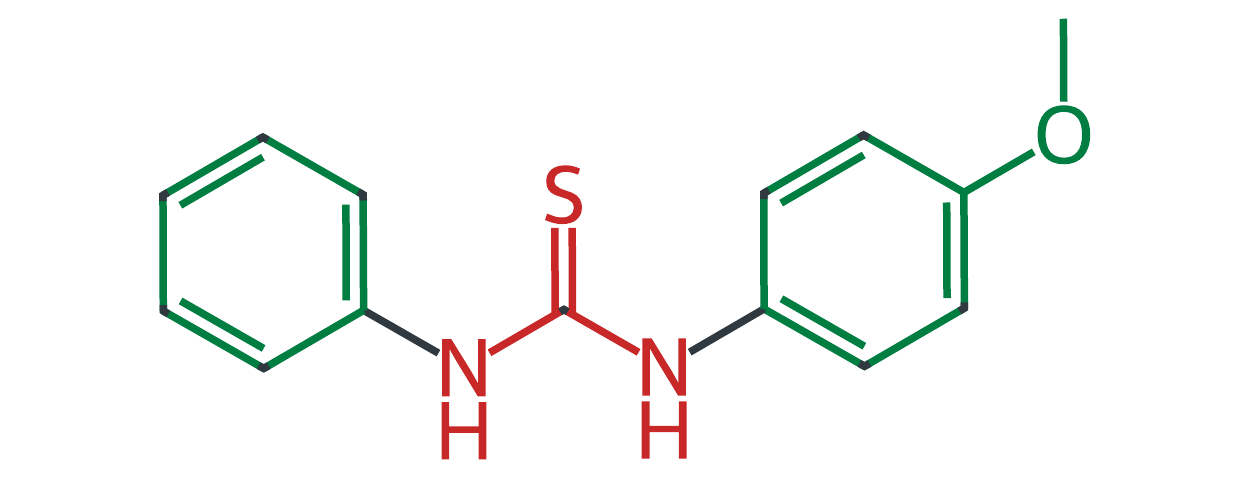}
    \end{minipage}\hfill
    \begin{minipage}[c]{0.08\linewidth}
        \centering $\rightarrow$
    \end{minipage}\hfill
    \begin{minipage}[c]{0.44\linewidth}
        \centering
        \includegraphics[width=\linewidth]{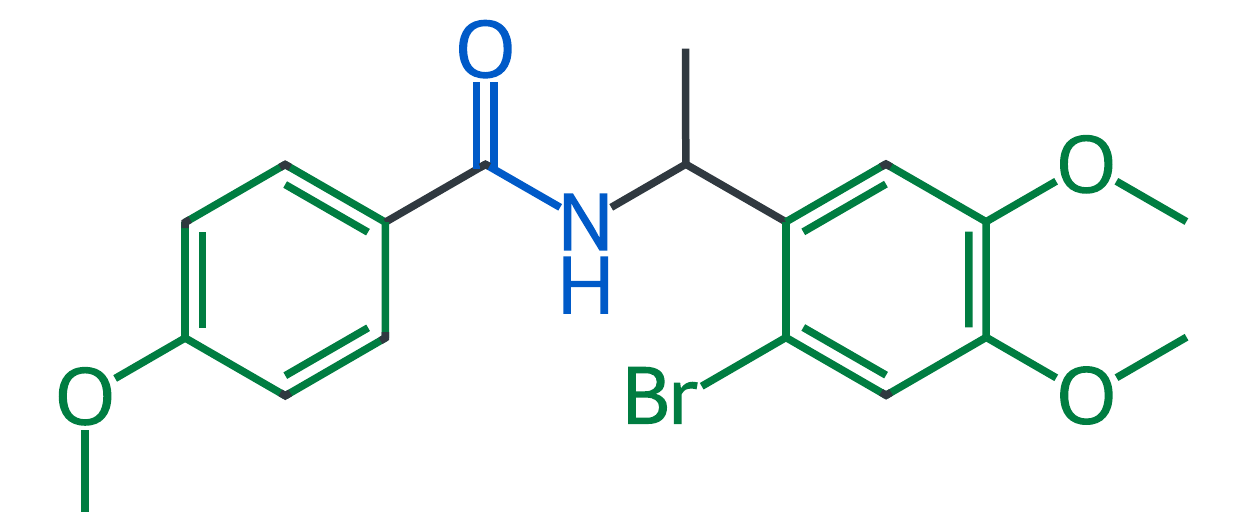}
    \end{minipage}\par\smallskip
    \makebox[0.44\linewidth][c]{Non-inhibitor}\hfill
    \makebox[0.44\linewidth][c]{Inhibitor}
\end{minipage}
&
\begin{minipage}[t]{\linewidth}
    \vspace{0pt}\vspace{3pt}\raggedright
    Compared with the reference's \cypth{thiourea linker}, the target's
    \cypam{benzamide linker} and \cypar{substituted aromatic scaffold}
    may contribute to CYP2C9 inhibition.
\end{minipage}
\\[7pt]
\midrule
\makebox[\linewidth][c]{\textbf{New Observation}} & \makebox[\linewidth][c]{\textbf{Revised Hypothesis}} \\
\specialrule{\lightrulewidth}{\aboverulesep}{0pt}
\begin{minipage}[t]{\linewidth}
    \vspace{0pt}\vspace{-\ht\strutbox}\vspace{3pt}\centering
    \makebox[\linewidth][c]{%
        \begin{minipage}[c]{0.414\linewidth}
            \centering
            \includegraphics[width=\linewidth]{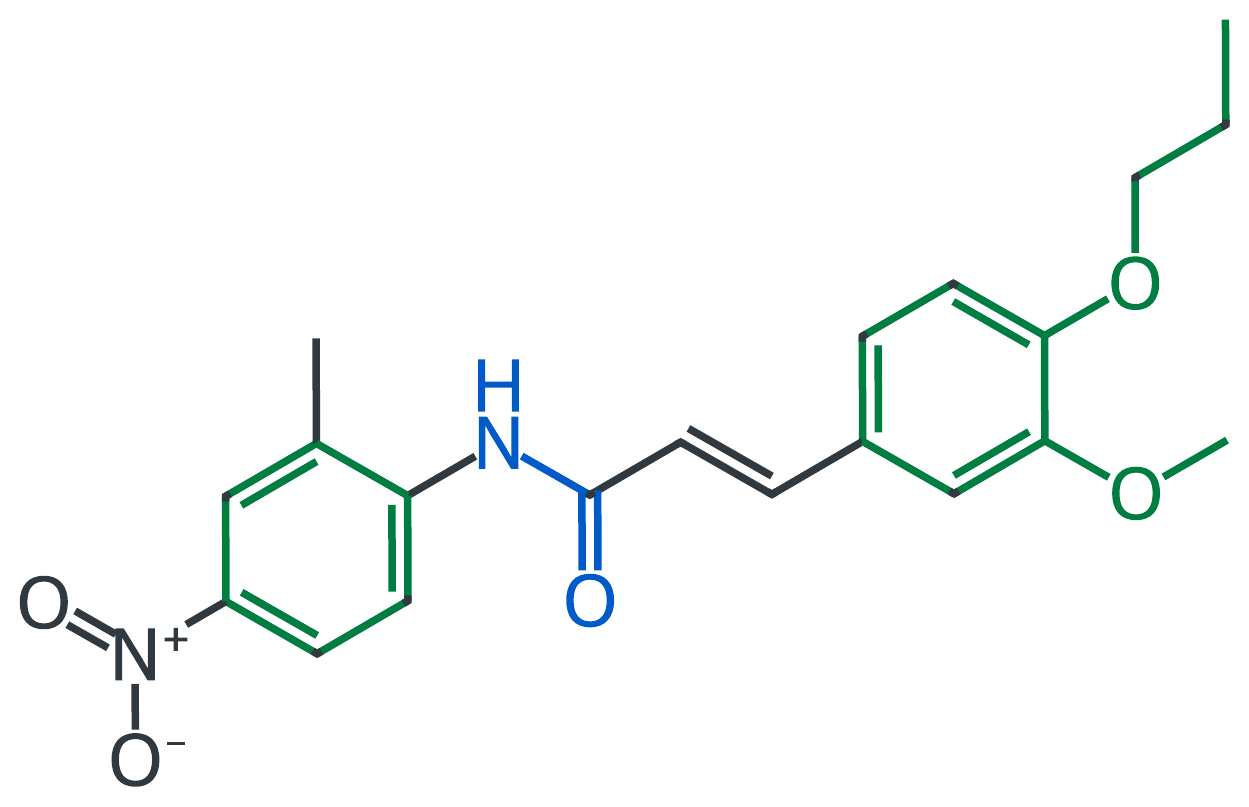}
        \end{minipage}%
        \hspace{8pt}%
        \begin{tabular}[c]{@{}l@{}}
            Predicted: Inhibitor \\[2pt]
            \textbf{Observed: Non-inhibitor}
        \end{tabular}%
    }
\end{minipage}
&
\begin{minipage}[t]{\linewidth}
    \vspace{0pt}\vspace{3pt}\raggedright
    An \cypam{amide motif} alone does not reliably explain inhibition;
    its role must be considered in the context of the surrounding
    \cypar{aromatic arrangement}.
\end{minipage}
\\[7pt]
\bottomrule
\end{tabular}\par
\endgroup

    \caption{\textbf{Hypothesis revision via new observations.}
A comparison between a non-inhibitor and an inhibitor yields comparative causal hypotheses about CYP2C9 inhibition.
These hypotheses were retrieved when predicting a new molecule, whose observed label differed from the prediction.
After incorporating new observations, including this case, the revised hypotheses explicitly limit the explanatory role of the amide motif alone.
Matching colors link molecular features to their mentions in the text: thiourea (red), amide (blue), and aromatic rings with selected substituents (green).
Hypothesis text is summarized; the revision qualifies the existing explanation rather than replacing it entirely.}
    \label{app:tab:revision_qualitative}
\end{table*}

\paragraph{Qualitative Analysis of Hypothesis Revision}
Table~\ref{app:tab:revision_qualitative} illustrates how new observations qualify an existing hypothesis in CYP2C9 prediction. The original comparative causal hypotheses associate the target's inhibitory activity with its benzamide linker and substituted aromatic scaffold. These hypotheses were retrieved when predicting a new amide-containing molecule that was predicted to inhibit CYP2C9 but was observed to be a non-inhibitor. After incorporating new observations, including this case, the revised hypotheses explicitly state that the presence of an amide alone does not reliably explain inhibition, clarifying the applicability of the original explanation.

\begin{figure}[t]
    \centering

    \begin{subfigure}[b]{0.47\columnwidth}
        \centering
        \includegraphics[width=\linewidth]{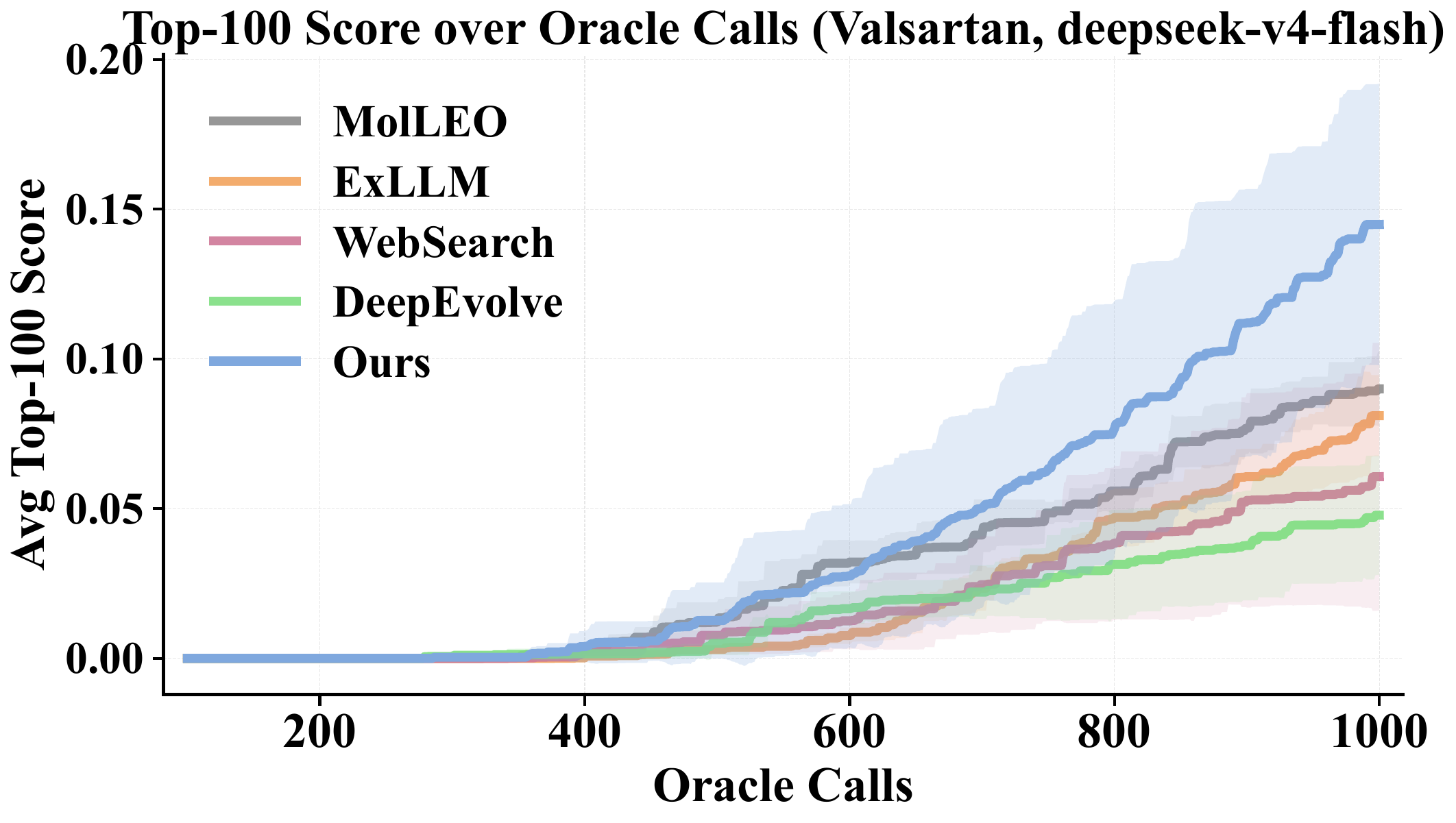}
        \label{fig:valsartan_deepseek_v4_flash}
    \end{subfigure}
    \hfill
    \begin{subfigure}[b]{0.47\columnwidth}
        \centering
        \includegraphics[width=\linewidth]{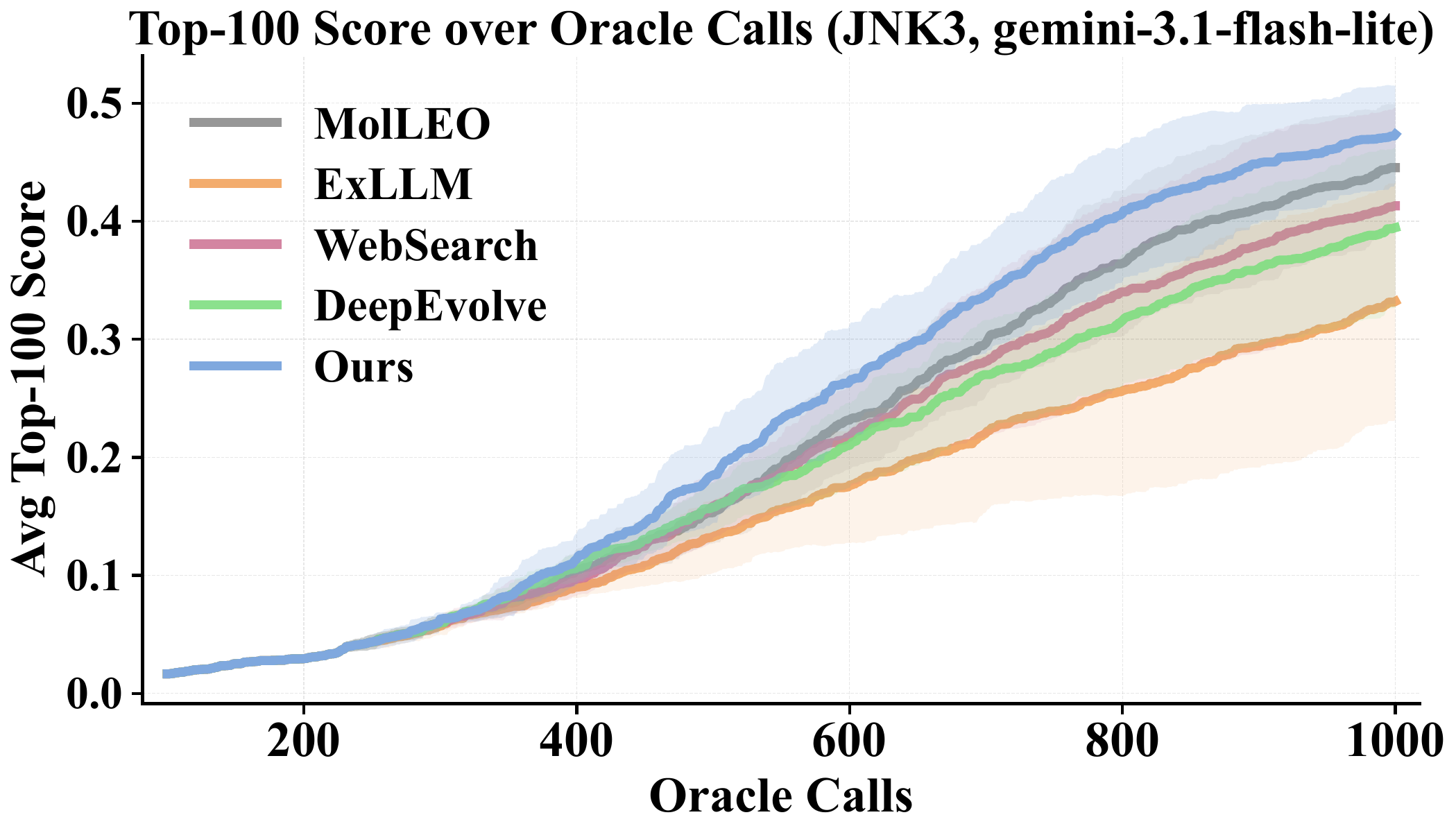}
        \label{fig:aav_deepseek_v4_flash}
    \end{subfigure}
    
    \begin{subfigure}[b]{0.47\columnwidth}
        \centering
        \includegraphics[width=\linewidth]{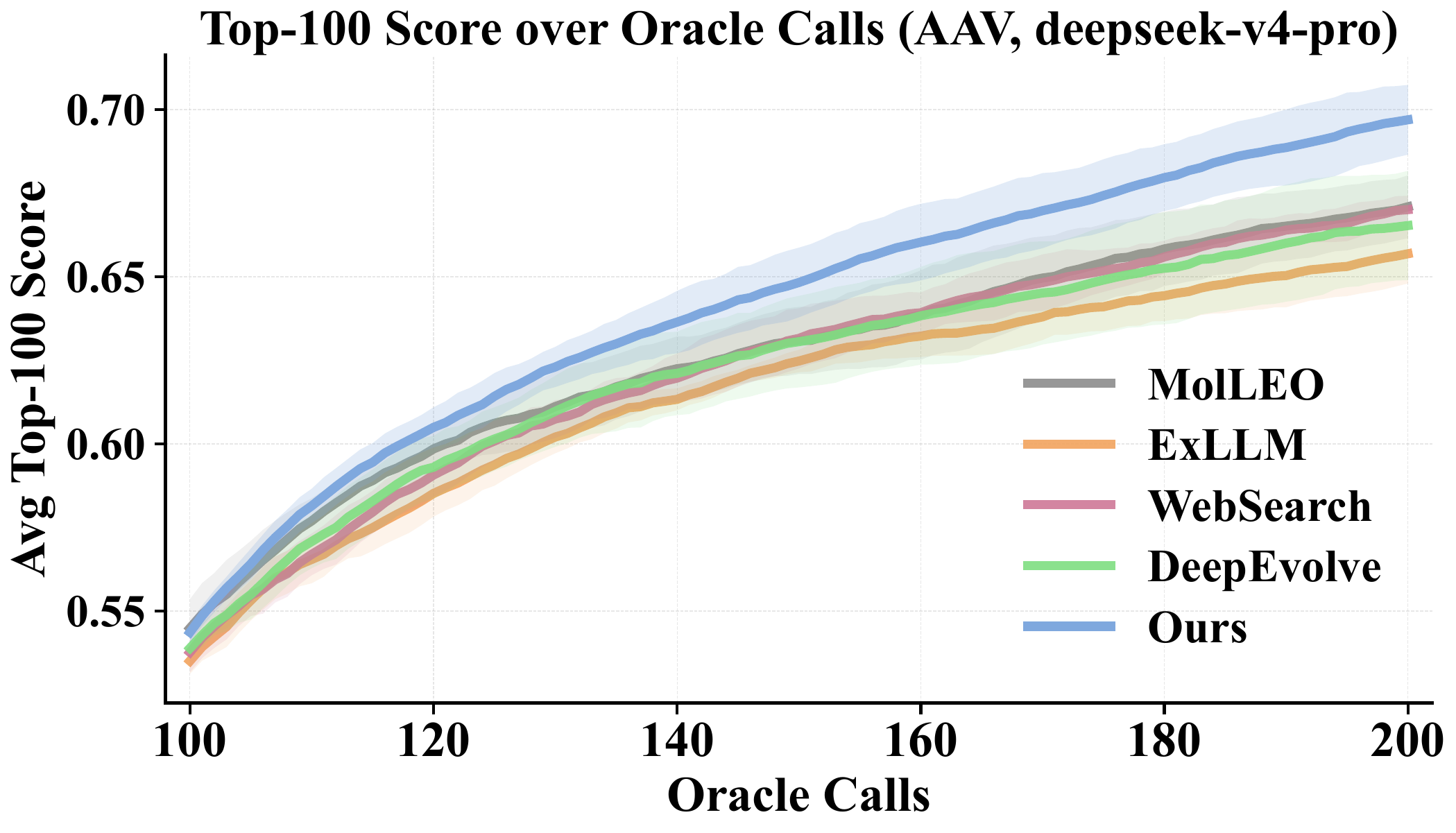}
        \label{fig:aav_deepseek_v4_pro}
    \end{subfigure}
    \hfill
    \begin{subfigure}[b]{0.47\columnwidth}
        \centering
        \includegraphics[width=\linewidth]{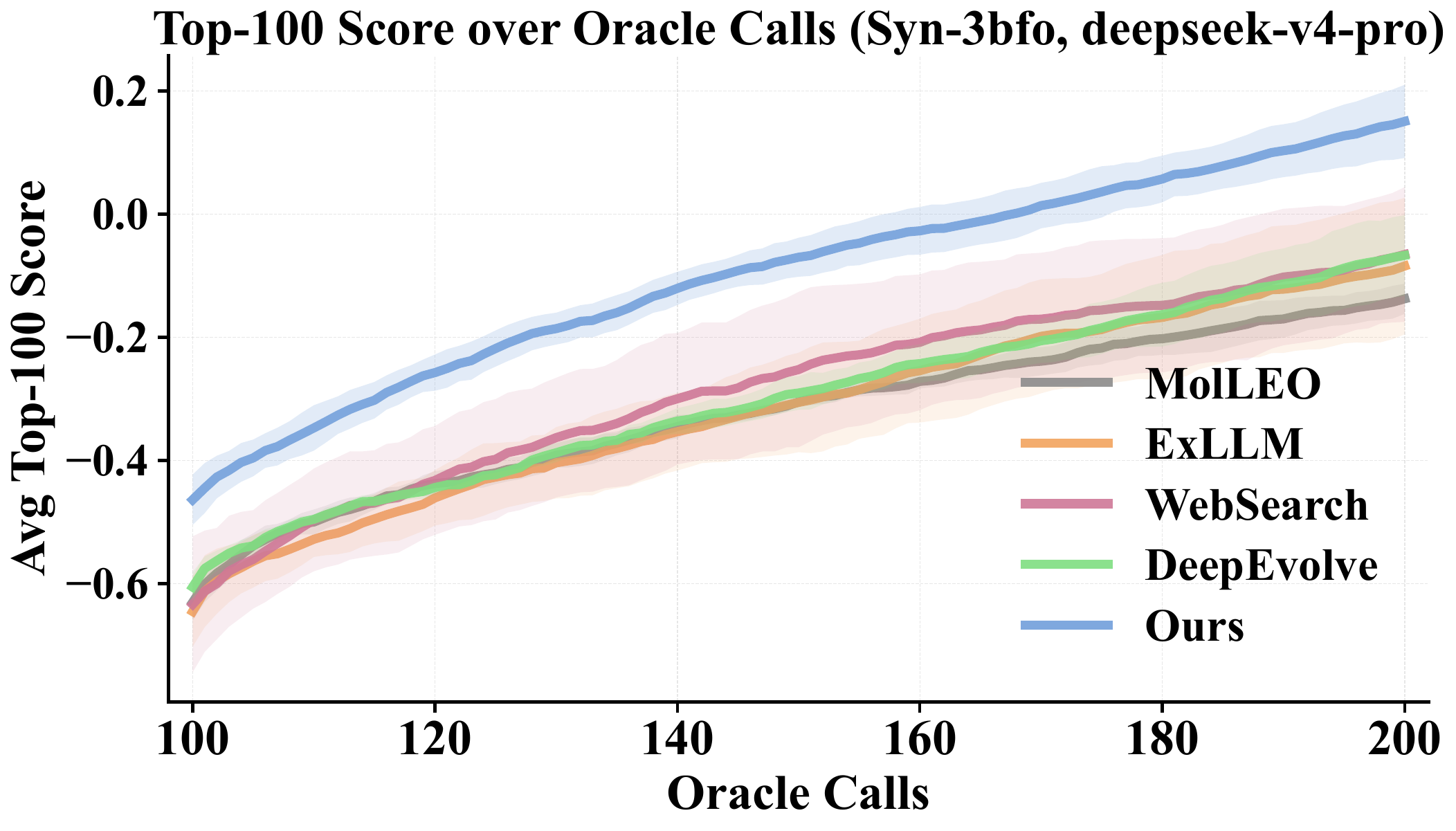}
        \label{fig:syn-3bfo_deepseek_v4_pro}
    \end{subfigure}

    \begin{subfigure}[b]{0.47\columnwidth}
        \centering
        \includegraphics[width=\linewidth]{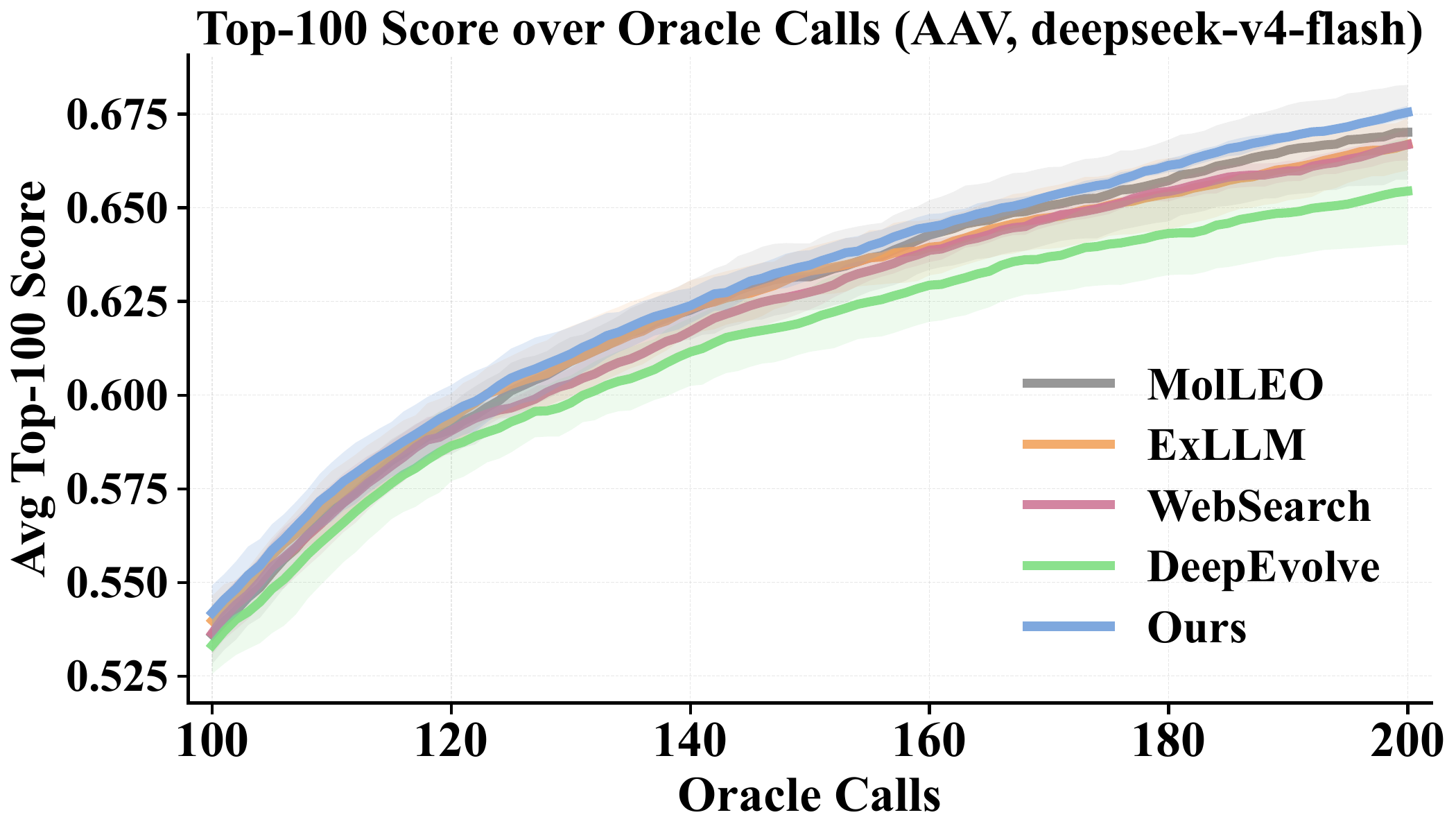}
        \label{fig:plot5}
    \end{subfigure}
    \hfill
    \begin{subfigure}[b]{0.47\columnwidth}
        \centering
        \includegraphics[width=\linewidth]{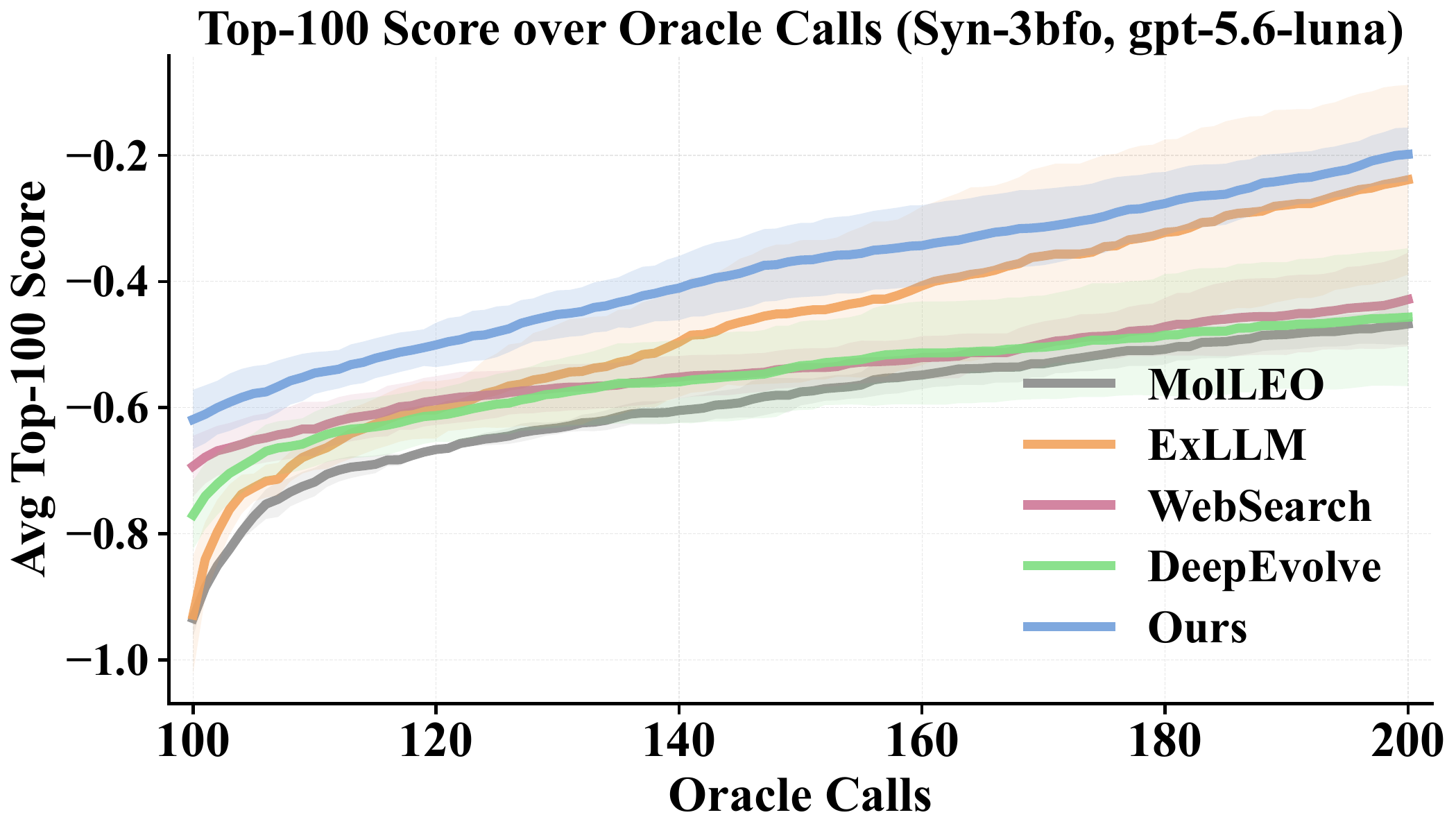}
        \label{fig:plot6}
    \end{subfigure}
    \vspace{-0.1in}
    \caption{Additional Top-100 average score trends for \textsc{Principia} across different backbone models and optimization tasks. For each task and backbone model, lines represent the mean fitness values over three random seeds, and shaded regions indicate the standard deviation.
}
    \label{fig:additional_results}
\end{figure}

\paragraph{Robustness of \textsc{Principia} to other model variants} The results in Figure~\ref{fig:additional_results} demonstrate the applicability of \textsc{Principia} across different backbone large language models, showing that its effectiveness is not limited to the models used in our main experiments. Specifically, across four backbone models including DeepSeek-V4-Flash~\citep{xu2026deepseek}, DeepSeek-V4-Pro~\citep{xu2026deepseek}, GPT-5.6-Luna~\citep{openai_gpt56}, and Gemini-3.1-Flash-Lite~\citep{google_deepmind_2026_gemini31flashlite}, \textsc{Principia} achieves higher Top-100 average fitness values than the corresponding baseline methods throughout the optimization process. These results demonstrate that the effectiveness of \textsc{Principia} generalizes across different backbone models and scientific discovery tasks.

\paragraph{Optimization Process over Oracle Calls} In Figures~\ref{fig:additional_results} and~\ref{fig:overall_results}, we present the trends of the Average Top-100 scores as the number of oracle calls increases. Across 16 combinations of optimization tasks and backbone models, \textsc{Principia} outperforms baselines. Notably, the performance advantage becomes more pronounced as additional observations accumulate in tasks such as JNK3, GSK3B, and GB1. These trends suggest that the knowledge constructed during discovery helps guide subsequent evaluations toward more promising candidates, improving the use of the available oracle budget.

\clearpage

\clearpage
\section{Prompts}
\label{app:sec:prompts}
In this section, we provide the prompts used for \textsc{Principia} in our experiments, along with those used for the baseline methods. The prompts retain the terminology used in the implementation: \emph{episodic memories} refer to hypothesis records containing comparative evidence and associated causal hypotheses, while \emph{semantic memories} refer to inductive principles. The source input in a source-to-target comparison corresponds to the reference input in the main text, and hypothesis refinement refers to hypothesis revision via new observations.

\subsection{Prompt for Molecular Optimization}
\label{app:sec:prompts_molecule_optimization}
Figure~\ref{app:fig:prompt_molecule_optimization_hypothesis_generation} shows the prompt for comparative causal hypothesis generation, and Figure~\ref{app:fig:prompt_molecule_optimization_principle_induction} shows the prompt for principle induction. Figure~\ref{app:fig:prompt_molecule_optimization_hypothesis_retrieval} and Figure~\ref{app:fig:prompt_molecule_optimization_principle_retrieval} provide the hypothesis and principle retrieval prompts, respectively. Figure~\ref{app:fig:prompt_molecule_optimization_hypothesis_revision} presents the prompt for hypothesis revision via new observations. Figure~\ref{app:fig:prompt_molecule_optimization_candidate_generation} shows the candidate generation prompt. Figure~\ref{app:fig:prompt_molecule_optimization_task_description} lists the task properties, descriptions, and objectives.

\subsection{Prompt for Protein Sequence Engineering}
\label{app:sec:prompts_protein_engineering}
Figure~\ref{app:fig:prompt_protein_engineering_hypothesis_generation} shows the prompt for comparative causal hypothesis generation, and Figure~\ref{app:fig:prompt_protein_engineering_principle_induction} shows the prompt for principle induction. Figure~\ref{app:fig:prompt_protein_engineering_hypothesis_retrieval} and Figure~\ref{app:fig:prompt_protein_engineering_principle_retrieval} provide the hypothesis and principle retrieval prompts, respectively. Figure~\ref{app:fig:prompt_protein_engineering_hypothesis_revision} presents the prompt for hypothesis revision via new observations. Figure~\ref{app:fig:prompt_protein_engineering_candidate_generation} shows the candidate generation prompt. Figure~\ref{app:fig:prompt_protein_engineering_task_description} lists the task properties, descriptions, and objectives.

\subsection{Prompt for Gene Perturbation}
\label{app:sec:prompts_gene_perturbation}
Figure~\ref{app:fig:prompt_gene_perturbation_hypothesis_generation} shows the prompt for comparative causal hypothesis generation, and Figure~\ref{app:fig:prompt_gene_perturbation_principle_induction} shows the prompt for principle induction. Figure~\ref{app:fig:prompt_gene_perturbation_hypothesis_retrieval} and Figure~\ref{app:fig:prompt_gene_perturbation_principle_retrieval} provide the hypothesis and principle retrieval prompts, respectively. Figure~\ref{app:fig:prompt_gene_perturbation_hypothesis_revision} presents the prompt for hypothesis revision via new observations. Figure~\ref{app:fig:prompt_gene_perturbation_candidate_generation} shows the candidate generation prompt, including instructions to exclude previously proposed genes. Figure~\ref{app:fig:prompt_gene_perturbation_task_description} lists the task properties, descriptions, and objectives.

\subsection{Prompt for Molecular Property Prediction}
\label{app:sec:prompts_molecular_property_prediction}
Figure~\ref{app:fig:prompt_molecular_property_prediction_hypothesis_generation} shows the prompt for comparative causal hypothesis generation, and Figure~\ref{app:fig:prompt_molecular_property_prediction_principle_induction} shows the prompt for principle induction. Figure~\ref{app:fig:prompt_molecular_property_prediction_hypothesis_retrieval} and Figure~\ref{app:fig:prompt_molecular_property_prediction_principle_retrieval} provide the hypothesis and principle retrieval prompts, respectively. Figure~\ref{app:fig:prompt_molecular_property_prediction_hypothesis_revision} presents the prompt for hypothesis revision via new observations. Figure~\ref{app:fig:prompt_molecular_property_prediction_prediction} shows the prompt for held-out prediction using retrieved hypotheses and principles. Figure~\ref{app:fig:prompt_molecular_property_prediction_task_description} lists the eight task descriptions, questions, observed labels, and answer formats. These templates correspond to the six-round sequential setting. Braced placeholders denote task-specific text or runtime data; block placeholders contain the corresponding hypothesis records, cluster members, hypotheses, or observations.

\newtcolorbox[use counter=figure]{promptbox21}{
  colback=blue!4,
  colframe=black!70,
  boxrule=1.2pt,
  arc=3pt,
  left=10pt,
  right=10pt,
  top=10pt,
  bottom=10pt,
  width=\textwidth,
  fontupper=\ttfamily\fontsize{8}{9}\selectfont,
  fonttitle=\small,
  title={Figure~\thefigure: Prompt for Comparative Causal Hypothesis Generation on Molecular Optimization},
  label={app:fig:prompt_molecule_optimization_hypothesis_generation},
  enhanced,
  breakable,
}

\begin{promptbox21}
You are a computational chemist analyzing the results of molecular optimization.\\
\\
The following molecular transformation was performed during an evolutionary optimization process:\\
\\
Parent molecule:\\
\{parent\_info\}\\
\\
Child molecule:\\
SMILES: \{child\_smiles\}\\
Fitness score: \{child\_fitness:.4f\}\\
\\
Target property: \{target\_property\}\\
Task Description: \{task\_description\}\\
\\
Fitness change: \{fitness\_change\_description\}\\
\\
Based on the structural differences between the parent and the child molecule, and the resulting fitness change, provide a concise causal hypothesis explaining why this transformation led to the observed fitness change.\\
\\
Return the output as a valid JSON object with the following structure only:\\
\\
\{\\
\{\\
        "hypotheses": [\\
        \{\\
        \{\\
            "causality": "A free-text mechanism description",\\
            "evidence\_features": ["feature1", ...]\\
        \}\\
        \},\\
        ...]\\
\}\\
\}\\
\\
Requirements:\\
- "causality" must be a concise explanation of why the structural transformation likely caused the observed fitness change.\\
- Focus on specific structural modifications such as functional group additions/removals, ring changes, scaffold changes, or atom substitutions.\\
- "evidence\_features" must contain short phrases naming the key structural changes that support the hypothesis.
- Output only the JSON object and nothing else.
\end{promptbox21}
\newtcolorbox[use counter=figure]{promptbox4}{
  colback=blue!4,
  colframe=black!70,
  boxrule=1.2pt,
  arc=3pt,
  left=10pt,
  right=10pt,
  top=10pt,
  bottom=10pt,
  width=\textwidth,
  fontupper=\ttfamily\fontsize{8}{9}\selectfont,
  fonttitle=\small,
  title={Figure~\thefigure: Prompt for Principle Induction on Molecular Optimization},
  label={app:fig:prompt_molecule_optimization_principle_induction},
  enhanced,
  breakable,
}

\begin{promptbox4}
You are summarizing a molecular similarity cluster from an evolutionary optimization run.\\
\\
You will be given multiple molecules from the same cluster. For each molecule, you will see:\\
- SMILES\\
- Fitness score (\{target\_property\})\\
- Hypotheses from parent→child transformations (if any)\\
\\
Your task is to extract a short, reusable, cluster-specific semantic memory that captures \
structural patterns that are associated with fitness changes.\\
\\
Internally do this before producing the final JSON:\\
1. Identify the structural features that recur across the SMILES in this cluster.\\
2. Separate core scaffold features from peripheral substituent variation.\\
3. Check whether the hypotheses support those recurring structural features.\\
4. Remove any statement that is too generic (e.g., "adding functional groups improves fitness").\\
5. Keep only features that are specific to this cluster.\\
\\
Priority order:\\
1. Shared scaffold / chemotype\\
2. Recurring substituent or functional group patterns linked to fitness improvement\\
3. Structural modifications that consistently increased or decreased fitness\\
4. Contradictions or inconsistencies across members\\
Definition of a good cluster memory:\\
- Names the recurring scaffold shape or motif\\
- Explains what structural changes improve vs. harm fitness\\
- Is specific enough to guide future mutations for similar molecules\\
\\
Definition of a bad cluster memory:\\
- Mostly generic chemistry language without cluster-specific detail\\
- Ignores the actual recurring scaffold in the SMILES\\
- Restates hypotheses verbatim without synthesis\\
- Explicitly lists the index of each molecule as evidence\\
Return JSON:\\
\{\{\\
  "cluster\_summary": "...",\\
  "patterns": \{\{\\
    "positive": ["structural patterns linked to higher fitness..."],\\
    "negative": ["structural patterns linked to lower fitness..."]\\
  \}\},\\
  "contradictions": ["...", "..."]\\
\}\}\\
Output constraints:\\
- Mention the actual recurring scaffold in cluster\_summary.\\
- Use cautious frequency qualifiers (e.g., "tends to", "often").\\
- If all transformations improved fitness, set "contradictions" to [].\\
- Do not include claims unsupported by the cluster examples.\\
\\
Target property: \{target\_property\}\\
\\
Cluster data:\\
\{cluster\_examples\}
\end{promptbox4}
\newtcolorbox[use counter=figure]{promptbox12}{
  colback=blue!4,
  colframe=black!70,
  boxrule=1.2pt,
  arc=3pt,
  left=10pt,
  right=10pt,
  top=10pt,
  bottom=10pt,
  width=\textwidth,
  fontupper=\ttfamily\fontsize{8}{9}\selectfont,
  fonttitle=\small,
  title={Figure~\thefigure: Prompt for Hypothesis Retrieval on Molecular Optimization},
  label={app:fig:prompt_molecule_optimization_hypothesis_retrieval},
  enhanced,
  breakable,
}

\begin{promptbox12}
You are selecting the most relevant episodic memories for a molecular optimization step.\\
\\
Query molecules:\\
\{query\_molecules\_text\}\\
\\
Target property: \{target\_property\}\\
Task Description: \{task\_description\}\\
\\
You will receive a list of candidate episodic memories. Each candidate includes:\\
- an index\\
- a parent molecule and child molecule\\
- the observed fitness change\\
- the inferred hypotheses for that transformation\\
\\
Select the indices of the memories that are most relevant for guiding the next mutation or crossover of the query molecules.\\
\\
Selection priorities:\\
1. Hypotheses that are structurally relevant to the query molecules\\
2. Transformations with actionable, reusable guidance\\
3. Transformations that improved fitness, unless a decreasing-fitness example is clearly useful as a warning\\
4. Avoid redundant memories that say essentially the same thing\\
\\
Return only valid JSON with this schema:\\
\{\{\\
  "indices": [0, 3, 5]\\
\}\}\\
\\
Requirements:\\
- Select at most \{top\_k\} indices.\\
- Indices must come only from the provided candidate list.\\
- If fewer than \{top\_k\} are useful, return fewer.\\
- Output JSON only.\\
\\
Candidate episodic memories:\\
\{candidate\_memories\_text\}
\end{promptbox12}
\newtcolorbox[use counter=figure]{promptbox15}{
  colback=blue!4,
  colframe=black!70,
  boxrule=1.2pt,
  arc=3pt,
  left=10pt,
  right=10pt,
  top=10pt,
  bottom=10pt,
  width=\textwidth,
  fontupper=\ttfamily\fontsize{8}{9}\selectfont,
  fonttitle=\small,
  title={Figure~\thefigure: Prompt for Principle Retrieval on Molecular Optimization},
  label={app:fig:prompt_molecule_optimization_principle_retrieval},
  enhanced,
  breakable,
}

\begin{promptbox15}
You are selecting the most relevant semantic memories for a molecular optimization step.\\
\\
Query molecules:\\
\{query\_molecules\_text\}\\
\\
Target property: \{target\_property\}\\
Task Description: \{task\_description\}\\
\\
You will receive a list of candidate semantic memories. Each candidate is a cluster-level summary of recurring structural patterns observed during optimization.\\
\\
Select the indices of the semantic memories that are most relevant for guiding the next mutation or crossover of the query molecules.\\
\\
Selection priorities:\\
1. Cluster summaries and patterns that best match the likely scaffold or substituent space of the query molecules\\
2. Memories with concrete positive/negative structural guidance\\
3. Diverse but complementary guidance when selecting multiple memories\\
4. Avoid redundant clusters with overlapping advice\\
\\
Return only valid JSON with this schema:\\
\{\{\\
  "indices": [1, 2]\\
\}\}\\
\\
Requirements:\\
- Select at most \{top\_k\} indices.\\
- Indices must come only from the provided candidate list.\\
- If fewer than \{top\_k\} are useful, return fewer.\\
- Output JSON only.\\
\\
Candidate semantic memories:\\
\{candidate\_memories\_text\}
\end{promptbox15}

\newtcolorbox[use counter=figure]{promptbox17}{
  colback=blue!4,
  colframe=black!70,
  boxrule=1.2pt,
  arc=3pt,
  left=10pt,
  right=10pt,
  top=10pt,
  bottom=10pt,
  width=\textwidth,
  fontupper=\ttfamily\fontsize{8}{9}\selectfont,
  fonttitle=\small,
  title={Figure~\thefigure: Prompt for Hypothesis Revision on Molecular Optimization},
  label={app:fig:prompt_molecule_optimization_hypothesis_revision},
  enhanced,
  breakable,
}

\begin{promptbox17}
You are a computational chemist refining causal hypotheses about molecular optimization.\\
\\
During an evolutionary optimization process, a source molecule was mutated into a target molecule.\\
The hypotheses below were originally inferred to explain this transformation.\\
Now, additional mutation outcomes (observations) have been collected from other transformations.\\
Use these observations to refine the hypotheses.\\
\\
=== Edge Being Refined ===\\
Source molecule:\\
  SMILES: \{source\_smiles\}\\
  Fitness score: \{source\_fitness:.4f\}\\
\\
Target molecule:\\
  SMILES: \{target\_smiles\}\\
  Fitness score: \{target\_fitness:.4f\}\\
\\
Target property: \{target\_property\}\\
Task Description: \{task\_description\}\\
\\
=== Current Hypotheses for This Edge ===\\
\{current\_hypotheses\_text\}\\
\\
=== Observations from Other Mutations ===\\
\{observations\_text\}\\
\\
Based on the observations above, please refine the current hypotheses for the source→target edge.
Consider the following:\\
- Which hypotheses are supported by the observed outcomes? Keep and strengthen them.\\
- Which hypotheses are contradicted by the outcomes? Revise or remove them.\\
- Are there any new patterns in the observations that suggest new hypotheses? Add them.\\
\\
Return the output as a valid JSON object with the following structure only:\\
\\
\{\{\\
    "hypotheses": [\\
        \{\{\\
            "causality": "A free-text mechanism description",\\
            "evidence\_features": ["feature1", ...]\}\},\\
        ...]\}\}\\
\\
Requirements:
- "causality" must be a concise explanation of the refined causal mechanism.\\
- "evidence\_features" must contain short phrases naming the key structural changes that support the hypothesis.\\
- Output only the JSON object and nothing else.
\end{promptbox17}
\newtcolorbox[use counter=figure]{promptbox7}{
  colback=blue!4,
  colframe=black!70,
  boxrule=1.2pt,
  arc=3pt,
  left=10pt,
  right=10pt,
  top=10pt,
  bottom=10pt,
  width=\textwidth,
  fontupper=\ttfamily\fontsize{8}{9}\selectfont,
  fonttitle=\small,
  title={Figure~\thefigure: Prompt for Candidate Generation on Molecular Optimization},
  label={app:fig:prompt_molecule_optimization_candidate_generation},
  enhanced,
  breakable,
}

\begin{promptbox7}
You are a molecular designer tasked with creating chemical analogs by mutating parent smiles strings.\\
\\
Given the following molecule:\\
Parent SMILES: \{parent\_smiles\}\\
Parent \{target\_property\} score: \{parent\_score\}
\\
Target property: \{target\_property\}\\
Task Description: \{task\_description\}\\
Task Objective: \{task\_objective\}\\
\\
\{in\_task\_experience\_memory\}
\\
Requirements:\\
1. Each analog should be chemically valid\\
2. Propose exactly \{num\_mutations\} mutated molecules\\
\\
Output format:\\
Return only the SMILES strings, one per line in /box\{\{SMILES\_HERE\}\}, without any additional text or explanations. YOU MUST FOLLOW THIS FORMAT.
\end{promptbox7}

\newtcolorbox[use counter=figure]{promptbox2}{
  colback=blue!4,
  colframe=black!70,
  boxrule=1.2pt,
  arc=3pt,
  left=10pt,
  right=10pt,
  top=10pt,
  bottom=10pt,
  width=\textwidth,
  fontupper=\ttfamily\fontsize{8}{9}\selectfont,
  fonttitle=\small,
  title={Figure~\thefigure: Task Property, Description, and Objective on Molecular Optimization},
  label={app:fig:prompt_molecule_optimization_task_description},
  enhanced,
  breakable,
}
\begin{promptbox2}
Task Property: JNK3\\
Task Description: I have a molecule and their JNK3 scores. The JNK3 score measures a molecular\'s biological activity against JNK3.\\
Task Objective: Please propose a new molecule that has a higher JNK3 score. You can either make mutations based on the given molecules or just propose a new molecule based on your knowledge.\\
\\
Task Property: GSK3$\beta$ \\
Task Description: I have a molecule and their GSK3$\beta$ scores. The GSK3$\beta$ score measures a molecular\'s biological activity against Glycogen Synthase Kinase 3 Beta.\\
Task Objective: Please propose a new molecule that has a higher GSK3$\beta$ score. You can either make mutations based on the given molecules or just propose a new molecule based on your knowledge.\\
\\
Task Property: Median1\\
Task Description: I have a molecule and their median1 scores. The median1 score measures the geometric mean of a molecule\'s ECFC4-fingerprint Tanimoto similarities to both camphor and menthol.\\
Task Objective: Please propose a new molecule that has a higher median1 score. You can either make mutations based on the given molecules or just propose a new molecule based on your knowledge.\\
\\
Task Property: Median2\\
Task Description: I have a molecule and their median2 scores. The median2 score measures the geometric mean of a molecule\'s ECFC6-fingerprint Tanimoto similarities to both tadalafil and sildenafil.\\
Task Objective: Please propose a new molecule that has a higher median2 score. You can either make mutations based on the given molecules or just propose a new molecule based on your knowledge.\\
\\
Task Property: Valsartan Smarts\\
Task Description: I have a molecule and their valsartan SMARTS scores. The valsartan SMARTS score measures whether a molecule contains a valsartan-related pharmacophore SMARTS pattern while matching target physicochemical properties (LogP, TPSA and Bertz index).\\
Task Objective: Please propose a new molecule that has a higher valsartan SMARTS score. You can either make mutations based on the given molecules or just propose a new molecule based on your knowledge.
\end{promptbox2}

\newtcolorbox[use counter=figure]{promptbox23}{
  colback=red!4,
  colframe=black!70,
  boxrule=1.2pt,
  arc=3pt,
  left=10pt,
  right=10pt,
  top=10pt,
  bottom=10pt,
  width=\textwidth,
  fontupper=\ttfamily\fontsize{8}{9}\selectfont,
  fonttitle=\small,
  title={Figure~\thefigure: Prompt for Comparative Causal Hypothesis Generation on Protein Sequence Engineering},
  label={app:fig:prompt_protein_engineering_hypothesis_generation},
  enhanced,
  breakable,
}

\begin{promptbox23}
You are a computational biologist analyzing the results of protein sequence optimization.\\
\\
The following sequence mutation was performed during an evolutionary optimization process:\\
\\
Parent sequence:\\
\{parent\_info\}\\
\\
Child sequence:\\
Sequence: \{child\_sequence\}\\
Fitness score: \{child\_fitness:.4f\}\\
\\
Target property: \{target\_property\}\\
Task Description: \{task\_description\}\\
\\
Fitness change: \{fitness\_change\_description\}\\
\\
Based on the differences between the parent and the child sequence, and the resulting fitness change, provide a concise causal hypothesis explaining why this mutation led to the observed fitness change.\\
\\
Return the output as a valid JSON object with the following structure only:\\
\\
\{\{\\
    "hypotheses": [\{\{\\
            "causality": "A free-text mechanism description",\\
            "evidence\_features": ["feature1", ...]\}\},\\
        ...]\}\}\\
\\
Requirements:\\
- "causality" must be a concise explanation of why the sequence change likely caused the observed fitness change.\\
- Focus on specific residue substitutions, changes at conserved positions, or motif changes.\\
- "evidence\_features" must contain short phrases naming the key sequence changes that support the hypothesis.\\
- Output only the JSON object and nothing else.
\end{promptbox23}
\newtcolorbox[use counter=figure]{promptbox6}{
  colback=red!4,
  colframe=black!70,
  boxrule=1.2pt,
  arc=3pt,
  left=10pt,
  right=10pt,
  top=10pt,
  bottom=10pt,
  width=\textwidth,
  fontupper=\ttfamily\fontsize{8}{9}\selectfont,
  fonttitle=\small,
  title={Figure~\thefigure: Prompt for Principle Induction on Protein Sequence Engineering},
  label={app:fig:prompt_protein_engineering_principle_induction},
  enhanced,
  breakable,
}

\begin{promptbox6}
You are summarizing a protein sequence similarity cluster from an evolutionary optimization run.\\
\\
You will be given multiple protein sequences from the same cluster. For each sequence, you will see:\\
- Sequence\\
- Fitness score (\{target\_property\})\\
- Hypotheses from parent→child mutations (if any)\\
\\
Your task is to extract a short, reusable, cluster-specific semantic memory that captures \
sequence patterns that are associated with fitness changes.\\
\\
Internally do this before producing the final JSON:\\
1. Identify the residues/motifs that recur across the sequences in this cluster.\\
2. Separate conserved core positions from variable positions.\\
3. Check whether the hypotheses support those recurring sequence features.\\
4. Remove any statement that is too generic (e.g., "mutations improve fitness").\\
5. Keep only features that are specific to this cluster.\\
Priority order:\\
1. Shared conserved motif / consensus positions\\
2. Recurring residue patterns at specific positions linked to fitness improvement\\
3. Residue substitutions that consistently increased or decreased fitness\\
4. Contradictions or inconsistencies across members\\
Definition of a good cluster memory:\\
- Names the recurring sequence motif or conserved positions\\
- Explains what residue changes improve vs. harm fitness\\
- Is specific enough to guide future mutations for similar sequences\\
\\
Definition of a bad cluster memory:\\
- Mostly generic language without cluster-specific detail\\
- Ignores the actual recurring motif in the sequences\\
- Restates hypotheses verbatim without synthesis\\
- Explicitly lists the index of each sequence as evidence\\
Return JSON:\\
\{\{\\
  "cluster\_summary": "...",\\
  "patterns": \{\{\\
    "positive": ["sequence patterns linked to higher fitness..."],\\
    "negative": ["sequence patterns linked to lower fitness..."]\\
  \}\},\\
  "contradictions": ["...", "..."]\\
\}\}\\
\\
Output constraints:\\
- Mention the actual recurring motif or conserved positions in cluster\_summary.\\
- Use cautious frequency qualifiers (e.g., "tends to", "often").\\
- If all mutations improved fitness, set "contradictions" to [].\\
- Do not include claims unsupported by the cluster examples.\\
\\
Target property: \{target\_property\}\\
\\
Cluster data:\\
\{cluster\_examples\}
\end{promptbox6}
\newtcolorbox[use counter=figure]{promptbox13}{
  colback=red!4,
  colframe=black!70,
  boxrule=1.2pt,
  arc=3pt,
  left=10pt,
  right=10pt,
  top=10pt,
  bottom=10pt,
  width=\textwidth,
  fontupper=\ttfamily\fontsize{8}{9}\selectfont,
  fonttitle=\small,
  title={Figure~\thefigure: Prompt for Hypothesis Retrieval on Protein Sequence Engineering},
  label={app:fig:prompt_protein_engineering_hypothesis_retrieval},
  enhanced,
  breakable,
}

\begin{promptbox13}
You are selecting the most relevant episodic memories for a protein sequence optimization step.\\
\\
Query sequences:\\
\{query\_sequences\_text\}\\
\\
Target property: \{target\_property\}\\
Task Description: \{task\_description\}\\
\\
You will receive a list of candidate episodic memories. Each candidate includes:\\
- an index\\
- a parent sequence and child sequence\\
- the observed fitness change\\
- the inferred hypotheses for that mutation\\
Select the indices of the memories that are most relevant for guiding the next mutation or crossover of the query sequences.\\
\\
Selection priorities:\\
1. Hypotheses that are relevant to the query sequences\\
2. Mutations with actionable, reusable guidance\\
3. Mutations that improved fitness, unless a decreasing-fitness example is clearly useful as a warning\\
4. Avoid redundant memories that say essentially the same thing\\
\\
Return only valid JSON with this schema:\\
\{\{\\
  "indices": [0, 3, 5]\\
\}\}\\
\\
Requirements:\\
- Select at most \{top\_k\} indices.\\
- Indices must come only from the provided candidate list.\\
- If fewer than \{top\_k\} are useful, return fewer.\\
- Output JSON only.\\
\\
Candidate episodic memories:\\
\{candidate\_memories\_text\}
\end{promptbox13}
\newtcolorbox[use counter=figure]{promptbox16}{
  colback=red!4,
  colframe=black!70,
  boxrule=1.2pt,
  arc=3pt,
  left=10pt,
  right=10pt,
  top=10pt,
  bottom=10pt,
  width=\textwidth,
  fontupper=\ttfamily\fontsize{8}{9}\selectfont,
  fonttitle=\small,
  title={Figure~\thefigure: Prompt for Principle Retrieval on Protein Sequence Engineering},
  label={app:fig:prompt_protein_engineering_principle_retrieval},
  enhanced,
  breakable,
}

\begin{promptbox16}
You are selecting the most relevant semantic memories for a protein sequence optimization step.\\
\\
Query sequences:\\
\{query\_sequences\_text\}\\
\\
Target property: \{target\_property\}\\
Task Description: \{task\_description\}\\
\\
You will receive a list of candidate semantic memories. Each candidate is a cluster-level summary of recurring sequence patterns observed during optimization.\\
\\
Select the indices of the semantic memories that are most relevant for guiding the next mutation or crossover of the query sequences.\\
\\
Selection priorities:\\
1. Cluster summaries and patterns that best match the likely sequence/motif space of the query sequences\\
2. Memories with concrete positive/negative sequence-level guidance\\
3. Diverse but complementary guidance when selecting multiple memories\\
4. Avoid redundant clusters with overlapping advice\\
\\
Return only valid JSON with this schema:\\
\{\{"indices": [1, 2]\}\}\\
\\
Requirements:\\
- Select at most \{top\_k\} indices.\\
- Indices must come only from the provided candidate list.\\
- If fewer than \{top\_k\} are useful, return fewer.\\
- Output JSON only.\\
\\
Candidate semantic memories:\\
\{candidate\_memories\_text\}
\end{promptbox16}
\newtcolorbox[use counter=figure]{promptbox18}{
  colback=red!4,
  colframe=black!70,
  boxrule=1.2pt,
  arc=3pt,
  left=10pt,
  right=10pt,
  top=10pt,
  bottom=10pt,
  width=\textwidth,
  fontupper=\ttfamily\fontsize{8}{9}\selectfont,
  fonttitle=\small,
  title={Figure~\thefigure: Prompt for Hypothesis Revision on Protein Sequence Engineering},
  label={app:fig:prompt_protein_engineering_hypothesis_revision},
  enhanced,
  breakable,
}

\begin{promptbox18}
You are a computational biologist refining causal hypotheses about protein sequence optimization.\\
\\
During an evolutionary optimization process, a source sequence was mutated into a target sequence.\\
The hypotheses below were originally inferred to explain this mutation.\\
Now, additional mutation outcomes (observations) have been collected from other mutations.\\
Use these observations to refine the hypotheses.\\
\\
=== Edge Being Refined ===\\
Source sequence:\\
  Sequence: \{source\_sequence\}\\
  Fitness score: \{source\_fitness:.4f\}\\
\\
Target sequence:\\
  Sequence: \{target\_sequence\}
  Fitness score: \{target\_fitness:.4f\}

Target property: \{target\_property\}\\
Task Description: \{task\_description\}\\
\\
=== Current Hypotheses for This Edge ===\\
\{current\_hypotheses\_text\}\\
\\
=== Observations from Other Mutations ===\\
\{observations\_text\}\\
\\
Based on the observations above, please refine the current hypotheses for the source→target edge.\\
Consider the following:\\
- Which hypotheses are supported by the observed outcomes? Keep and strengthen them.\\
- Which hypotheses are contradicted by the outcomes? Revise or remove them.\\
- Are there any new patterns in the observations that suggest new hypotheses? Add them.\\
\\
Return the output as a valid JSON object with the following structure only:\\
\\
\{\{\\
    "hypotheses": [\\
        \{\{\\
            "causality": "A free-text mechanism description",\\
            "evidence\_features": ["feature1", ...]\\
        \}\}\\,
        ...\\
    ]\\
\}\}\\
\\
Requirements:\\
- "causality" must be a concise explanation of the refined causal mechanism.\\
- "evidence\_features" must contain short phrases naming the key sequence changes that support the hypothesis.\\
- Output only the JSON object and nothing else.
\end{promptbox18}
\newtcolorbox[use counter=figure]{promptbox10}{
  colback=red!4,
  colframe=black!70,
  boxrule=1.2pt,
  arc=3pt,
  left=10pt,
  right=10pt,
  top=10pt,
  bottom=10pt,
  width=\textwidth,
  fontupper=\ttfamily\fontsize{8}{9}\selectfont,
  fonttitle=\small,
  title={Figure~\thefigure: Prompt for Candidate Generation on Protein Sequence Engineering},
  label={app:fig:prompt_protein_engineering_candidate_generation},
  enhanced,
  breakable,
}

\begin{promptbox10}
You are a protein engineer tasked with creating variants by mutating parent amino-acid sequences.\\
\\
Given the following protein sequence:\\
Parent sequence: \{parent\_sequence\}\\
Parent \{target\_property\} score: \{parent\_score\}\\
\\
Target property: \{target\_property\}\\
Task Description: \{task\_description\}\\
Task Objective: \{task\_objective\}\\
\{in\_task\_experience\_memory\}\\
\\
Requirements:\\
1. Each variant must be a valid sequence over the 20 standard amino acids with the SAME length as the parent\\
2. Propose exactly \{num\_mutations\} mutated sequences\\
Output format:\\
Return only the amino-acid sequences, one per line in /box\{\{SEQUENCE\_HERE\}\}, without any additional text or explanations. YOU MUST FOLLOW THIS FORMAT.
\end{promptbox10}
\newtcolorbox[use counter=figure]{promptbox471}{
  colback=red!4,
  colframe=black!70,
  boxrule=1.2pt,
  arc=3pt,
  left=10pt,
  right=10pt,
  top=10pt,
  bottom=10pt,
  width=\textwidth,
  fontupper=\ttfamily\fontsize{8}{9}\selectfont,
  fonttitle=\small,
  title={Figure~\thefigure: Task Property, Description, and Objective on Protein Sequence Engineering},
  label={app:fig:prompt_protein_engineering_task_description},
  enhanced,
  breakable,
}

\begin{promptbox471}
Task Property: Syn-3bfo Potts Fitness\\
Task Description: I have protein sequences and their Syn-3bfo fitness scores. The score is a Potts-model fitness (the normalized negative energy) over a synthetic fitness landscape derived from PDB 3bfo, and a higher score is better.\\
Task Objective: Please propose a new protein sequence that has a higher Syn-3bfo fitness score. You can either mutate the given sequences or propose a new sequence based on your knowledge. Keep the same sequence length and use only the 20 standard amino acids.\\
\\
Task Property: GB1 Fitness\\
Task Description: I have protein sequences and their GB1 fitness scores. GB1 is the B1 domain of protein G; the fitness score measures the binding/stability of a short variable region, and a higher score is better. \\
Task Objective: Please propose a new protein sequence that has a higher GB1 fitness score. You can either mutate the given sequences or propose a new sequence based on your knowledge. Keep the same sequence length and use only the 20 standard amino acids.\\
\\
Task Property: AAV Capsid Fitness\\
Task Description: I have protein sequences and their AAV fitness scores. The score is predicted by a machine-learning model of AAV2 capsid protein viability over a short variable region, and a higher score is better.\\
Task Objective: Please propose a new protein sequence that has a higher AAV fitness score. You can either mutate the given sequences or propose a new sequence based on your knowledge. Keep the same sequence length and use only the 20 standard amino acids.\\
\\
Task Property: Syn-3bfo Potts fitness (Weighted Objective)\\
Task Description: I have protein sequences and their Syn-3bfo fitness scores. The score is a Potts-model fitness (the normalized negative energy) over a synthetic fitness landscape derived from PDB 3bfo, and a higher score is better. This is a multi-objective task: the score is a weighted sum (equal weights) of the raw fitness and a closeness term, 1 - (normalized Hamming distance) to a fixed reference sequence. A higher score means higher fitness while staying close to the reference (fewer edits from it).\\
Reference sequence: \{reference\_sequence\}\\
Task Objective: Please propose a new protein sequence that has a higher Syn-3bfo fitness score. You can either mutate the given sequences or propose a new sequence based on your knowledge. Keep the same sequence length and use only the 20 standard amino acids.
\end{promptbox471}

\newtcolorbox[use counter=figure]{promptbox20}{
  colback=green!4,
  colframe=black!70,
  boxrule=1.2pt,
  arc=3pt,
  left=10pt,
  right=10pt,
  top=10pt,
  bottom=10pt,
  width=\textwidth,
  fontupper=\ttfamily\fontsize{8}{9}\selectfont,
  fonttitle=\small,
  title={Figure~\thefigure: Prompt for Comparative Causal Hypothesis Generation on Gene Perturbation},
  label={app:fig:prompt_gene_perturbation_hypothesis_generation},
  enhanced,
  breakable,
}

\begin{promptbox20}
You are a computational biologist analyzing the results of a gene-perturbation screen.\\
\\
The following gene proposal was made during an evolutionary discovery process:\\
\\
Parent gene(s):\\
\{parent\_info\}
\\\\
Proposed gene:\\
Gene: \{child\_gene\}\\
Score: \{child\_fitness:.4f\}\\
\\
Research goal: \{target\_property\}\\
Task Description: \{task\_description\}\\
\\
Score change: \{fitness\_change\_description\}\\
\\
Based on the biological relationship between the parent gene(s) and the proposed gene, and the resulting score change, provide a concise causal hypothesis explaining why proposing this gene led to the observed score change.\\
\\
Return the output as a valid JSON object with the following structure only:\\
\\
\{\{\\
    "hypotheses": [\\
        \{\{\\
            "causality": "A free-text mechanism\\ description",\\
            "evidence\_features": ["feature1", ...]\\
        \}\},...]\}\}\\
\\
Requirements:\\
- "causality" must be a concise explanation of why proposing this gene likely caused the observed score change.\\
- Focus on biological relationships such as shared pathways, complexes, regulators, effectors, or interaction partners.\\
- "evidence\_features" must contain short phrases naming the key biological relationships that support the hypothesis.\\
- Output only the JSON object and nothing else.
\end{promptbox20}
\newtcolorbox[use counter=figure]{promptbox5}{
  colback=green!4,
  colframe=black!70,
  boxrule=1.2pt,
  arc=3pt,
  left=10pt,
  right=10pt,
  top=10pt,
  bottom=10pt,
  width=\textwidth,
  fontupper=\ttfamily\fontsize{8}{9}\selectfont,
  fonttitle=\small,
  title={Figure~\thefigure: Prompt for Principle Induction on Gene Perturbation},
  label={app:fig:prompt_gene_perturbation_principle_induction},
  enhanced,
  breakable,
}

\begin{promptbox5}
You are summarizing a gene cluster from an evolutionary gene-perturbation discovery run.\\
\\
You will be given multiple genes from the same cluster. For each gene, you will see:\\
- Gene symbol (HGNC)\\
- Fitness score ({target\_property})\\
- Hypotheses from parent→child transformations (if any)\\
\\
Your task is to extract a short, reusable, cluster-specific semantic memory that captures \
functional/biological patterns that are associated with fitness changes.\\
Internally do this before producing the final JSON:\\
1. Identify the biological themes that recur across the genes in this cluster \
(shared pathway, complex, functional family, or regulatory role).\\
2. Separate core shared function from peripheral gene-specific variation.\\
3. Check whether the hypotheses support those recurring functional themes.\\
4. Remove any statement that is too generic (e.g., "perturbing important genes improves the score").\\
5. Keep only patterns that are specific to this cluster.\\
\\
Priority order:\\
1. Shared pathway / protein complex / functional family\\
2. Recurring regulatory or mechanistic roles linked to fitness improvement\\
3. Perturbations that consistently increased or decreased fitness\\
4. Contradictions or inconsistencies across members\\
\\
Definition of a good cluster memory:\\
- Names the recurring pathway, complex, or functional theme\\
- Explains what kinds of genes improve vs. harm fitness\\
- Is specific enough to guide future gene proposals for similar genes\\
\\
Definition of a bad cluster memory:\\
- Mostly generic biology language without cluster-specific detail\\
- Ignores the actual recurring function shared by the genes\\
- Restates hypotheses verbatim without synthesis\\
- Explicitly lists the index of each gene as evidence\\
Return JSON:\\
\{\{\\
  "cluster\_summary": "...",\\
  "patterns": \{\{\\
    "positive": ["functional patterns linked to higher fitness..."],\\
    "negative": ["functional patterns linked to lower fitness..."]\\
  \}\},\\
  "contradictions": ["...", "..."]\\
\}\}\\
\\
Output constraints:\\
- Mention the actual recurring pathway/function in cluster\_summary.\\
- Use cautious frequency qualifiers (e.g., "tends to", "often").\\
- If all transformations improved fitness, set "contradictions" to [].\\
- Do not include claims unsupported by the cluster examples.\\
Target property: \{target\_property\}
\\
Cluster data:\\
\{cluster\_examples\}
\end{promptbox5}
\newtcolorbox[use counter=figure]{promptbox11}{
  colback=green!4,
  colframe=black!70,
  boxrule=1.2pt,
  arc=3pt,
  left=10pt,
  right=10pt,
  top=10pt,
  bottom=10pt,
  width=\textwidth,
  fontupper=\ttfamily\fontsize{8}{9}\selectfont,
  fonttitle=\small,
  title={Figure~\thefigure: Prompt for Hypothesis Retrieval on Gene Perturbation},
  label={app:fig:prompt_gene_perturbation_hypothesis_retrieval},
  enhanced,
  breakable,
}
\begin{promptbox11}
You are selecting the most relevant episodic memories for a gene-perturbation discovery step.\\
\\
Query genes:\\
\{query\_genes\_text\}\\
\\
Research goal: \{target\_property\}\\
Task Description:\{task\_description\}\\
\\
You will receive a list of candidate episodic memories. Each candidate includes:\\
- an index\\
- a parent gene and proposed gene\\
- the observed score change\\
- the inferred hypotheses for that proposal\\
\\
Select the indices of the memories that are most relevant for guiding the next gene proposal for the query genes.\\
\\
Selection priorities:\\
1. Hypotheses that are biologically relevant to the query genes\\
2. Proposals with actionable, reusable guidance\\
3. Proposals that improved the score, unless a decreasing-score example is clearly useful as a warning\\
4. Avoid redundant memories that say essentially the same thing\\
\\
Return only valid JSON with this schema:\\
\{\{\\
  "indices": [0, 3, 5]\\
\}\}\\
\\
Requirements:\\
- Select at most \{top\_k\} indices.\\
- Indices must come only from the provided candidate list.\\
- If fewer than \{top\_k\} are useful, return fewer.\\
- Output JSON only.\\
\\
Candidate episodic memories:\\
\{candidate\_memories\_text\}
\end{promptbox11}
\newtcolorbox[use counter=figure]{promptbox14}{
  colback=green!4,
  colframe=black!70,
  boxrule=1.2pt,
  arc=3pt,
  left=10pt,
  right=10pt,
  top=10pt,
  bottom=10pt,
  width=\textwidth,
  fontupper=\ttfamily\fontsize{8}{9}\selectfont,
  fonttitle=\small,
  title={Figure~\thefigure: Prompt for Principle Retrieval on Gene Perturbation},
  label={app:fig:prompt_gene_perturbation_principle_retrieval},
  enhanced,
  breakable,
}

\begin{promptbox14}
You are selecting the most relevant semantic memories for a gene-perturbation discovery step.\\
\\
Query genes:\\
\{query\_genes\_text\}\\
\\
Research goal: \{target\_property\}\\
Task Description: \{task\_description\}\\
\\
You will receive a list of candidate semantic memories. Each candidate is a cluster-level summary of recurring gene/pathway patterns observed during discovery.\\
\\
Select the indices of the semantic memories that are most relevant for guiding the next gene proposal for the query genes.\\
\\
Selection priorities:\\
1. Cluster summaries and patterns that best match the likely pathways or functions of the query genes\\
2. Memories with concrete positive/negative biological guidance\\
3. Diverse but complementary guidance when selecting multiple memories\\
4. Avoid redundant clusters with overlapping advice\\
\\
Return only valid JSON with this schema:\\
\{\{\\
  "indices": [1, 2]\\
\}\}\\
\\
Requirements:\\
- Select at most \{top\_k\} indices.\\
- Indices must come only from the provided candidate list.\\
- If fewer than \{top\_k\} are useful, return fewer.\\
- Output JSON only.\\
\\
Candidate semantic memories:\\
\{candidate\_memories\_text\}
\end{promptbox14}

\newtcolorbox[use counter=figure]{promptbox19}{
  colback=green!4,
  colframe=black!70,
  boxrule=1.2pt,
  arc=3pt,
  left=10pt,
  right=10pt,
  top=10pt,
  bottom=10pt,
  width=\textwidth,
  fontupper=\ttfamily\fontsize{8}{9}\selectfont,
  fonttitle=\small,
  title={Figure~\thefigure: Prompt for Hypothesis Revision on Gene Perturbation},
  label={app:fig:prompt_gene_perturbation_hypothesis_revision},
  enhanced,
  breakable,
}

\begin{promptbox19}
You are a computational biologist refining causal hypotheses about gene-perturbation discovery.\\
\\
During an evolutionary discovery process, a source gene led to proposing a target gene.\\
The hypotheses below were originally inferred to explain this relationship.\\
Now, additional proposal outcomes (observations) have been collected from other proposals.\\
Use these observations to refine the hypotheses.\\
\\
=== Edge Being Refined ===\\
Source gene:\\
  Gene: \{source\_gene\}\\
  Score: \{source\_fitness:.4f\}\\
\\
Target gene:\\
  Gene: \{target\_gene\}
  Score: \{target\_fitness:.4f\}

Research goal: \{target\_property\}
Task Description: \{task\_description\}
\\
=== Current Hypotheses for This Edge ===\\
\{current\_hypotheses\_text\}\\
\\
=== Observations from Other Proposals ===\\
\{observations\_text\}
\\
Based on the observations above, please refine the current hypotheses for the source->target edge.
Consider the following:\\
- Which hypotheses are supported by the observed outcomes? Keep and strengthen them.\\
- Which hypotheses are contradicted by the outcomes? Revise or remove them.\\
- Are there any new patterns in the observations that suggest new hypotheses? Add them.\\
\\
Return the output as a valid JSON object with the following structure only:\\
\\
\{\{\\
    "hypotheses": [\\
        \{\{\\
            "causality": "A free-text mechanism\\ description",\\
            "evidence\_features": ["feature1", ...]\\
        \}\},\\
        ...]\}\}\\
\\
Requirements:\\
- "causality" must be a concise explanation of the refined causal mechanism.\\
- "evidence\_features" must contain short phrases naming the key biological relationships that support the hypothesis.\\
- Output only the JSON object and nothing else.
\end{promptbox19}
\newtcolorbox[use counter=figure]{promptbox8}{
  colback=green!4,
  colframe=black!70,
  boxrule=1.2pt,
  arc=3pt,
  left=10pt,
  right=10pt,
  top=10pt,
  bottom=10pt,
  width=\textwidth,
  fontupper=\ttfamily\fontsize{8}{9}\selectfont,
  fonttitle=\small,
  title={Figure~\thefigure: Prompt for Candidate Generation on Gene Perturbation},
  label={app:fig:prompt_gene_perturbation_candidate_generation},
  enhanced,
  breakable,
}

\begin{promptbox8}
You are a geneticist proposing new candidate genes to perturb, informed by two parent genes and your biological knowledge.\\
Given the following two parent genes and their scores:\\
\\
Parent 1 gene: \{parent\_gene\_1\}\\
Parent 1 score: \{parent\_score\_1\}\\
Parent 2 gene: \{parent\_gene\_2\}\\
Parent 2 score: \{parent\_score\_2\}\\
Research goal: \{target\_property\}\\

Task Description: \{task\_description\}\\
Task Objective: \{task\_objective\}\\
\{in\_task\_experience\_memory\}\\
\\
You may propose genes related to either or both parents (shared pathways,
complexes, regulators, effectors, or interaction partners), or propose
entirely new genes informed by the parents and your knowledge of the
relevant biology. Prioritize proposals likely to improve upon the parents'
current scores.\\
Requirements:\\
1. Each gene must be a real human gene using its official HGNC symbol\\
2. Prefer novel candidate genes that are unlikely to have been proposed before, while remaining biologically valid and plausibly relevant to the research goal\\
3. Propose exactly \{num\_children\} new candidate genes\\
Output format:\\
Return only the gene symbols, one per line in /box\{\{GENE\_HERE\}\}, without any additional text or explanations. YOU MUST FOLLOW THIS FORMAT.\\
\\
STRICT EXCLUSION \\
— genes already proposed in earlier rounds:\\
The following genes have ALREADY been proposed/tested in previous rounds of this screen. You MUST NOT propose any of them again. Every gene you output must be different from all of the genes listed below.\\
\\
\{excluded\_genes\}
\end{promptbox8}

\newtcolorbox[use counter=figure]{promptbox3}{
  colback=green!4,
  colframe=black!70,
  boxrule=1.2pt,
  arc=3pt,
  left=10pt,
  right=10pt,
  top=10pt,
  bottom=10pt,
  width=\textwidth,
  fontupper=\ttfamily\fontsize{8}{9}\selectfont,
  fonttitle=\small,
  title={Figure~\thefigure: Task Property, Description, and Objective on Gene Perturbation},
  label={app:fig:prompt_gene_perturbation_task_description},
  enhanced,
  breakable,
}

\begin{promptbox3}
Task Property: identify genes mediating lysosomal choline recycling using an endolysosome-focused CRISPR-Cas9 screen\\
Task Description: I am running a genome-wide CRISPR screen to identify genes mediating lysosomal choline recycling using an endolysosome-focused CRISPR-Cas9 screen. For each gene I perturb, I can measure enrichment or depletion of targeting sgRNAs (indicated by a high score) in the culture medium lacking free choline, which will be referred to as the score; a higher score indicates a stronger hit. \\
Task Objective: Please propose new genes to perturb that are likely to maximize the score. Use official HGNC gene symbols. You can build on the genes and scores observed so far or propose new genes based on your knowledge of the relevant biology and pathways.\\
\\
Task Property: identify genes that regulate the production of Interferon-gamma (IFNG) \\
Task Description: I am running a genome-wide CRISPR screen to identify genes that regulate the production of Interferon-gamma (IFNG). For each gene I perturb, I can measure the log fold change in Interferon-gamma (IFNG) normalized read counts. The score reflects the magnitude of this effect, so a gene that strongly shifts the phenotype in EITHER direction (a strong increase or a strong decrease) receives a high score; a higher score indicates a stronger hit. \\
Task Objective: Please propose new genes to perturb that are likely to maximize the score, i.e. genes that produce the largest change in the phenotype regardless of direction (a strong increase OR a strong decrease). Use official HGNC gene symbols. You can build on the genes and scores observed so far or propose new genes based on your knowledge of the relevant biology and pathways.
\end{promptbox3}

\newtcolorbox[use counter=figure]{predictionprompthypothesisgeneration}{
  colback=blue!4,
  colframe=black!70,
  boxrule=1.2pt,
  arc=3pt,
  left=10pt,
  right=10pt,
  top=10pt,
  bottom=10pt,
  width=\textwidth,
  fontupper=\ttfamily\fontsize{8}{9}\selectfont,
  fonttitle=\small,
  title={Figure~\thefigure: Prompt for Comparative Causal Hypothesis Generation on Molecular Property Prediction},
  label={app:fig:prompt_molecular_property_prediction_hypothesis_generation},
  enhanced,
  breakable,
}

\begin{predictionprompthypothesisgeneration}
You are a medicinal chemistry expert. Given a target molecule (SMILES) with its observed label and ONE structurally similar reference molecule, generate plausible mechanistic hypotheses explaining why the target shows the observed activity, grounded in how it structurally differs from that reference.\\
\\
CRITICAL RULES:\\
1. Do NOT simply restate the label or predict the label\\
2. Every hypothesis must be grounded in a concrete structural difference between the target and the reference\\
3. If the reference carries the SAME label, the shared substructure is the likely driver — name it\\
4. If the reference carries a DIFFERENT label, the substructure that differs is the likely driver — name it\\
5. Generate 1-2 distinct hypotheses from this single comparison\\
6. Provide the relevant substructures as evidence features\\
\\
Task: \{task\_description\}\\
\\
Target molecule:\\
  SMILES: \{smiles\}\\
  Observed Label: \{label\}\\
\\
Reference molecule:\\
  SMILES: \{reference\_smiles\}\\
  Observed Label: \{reference\_label\} (label \{label\_agreement\})\\
  Tanimoto similarity to target: \{similarity\}\\
\\
Output your response as a JSON object with the following format:\\
\{\\
    "hypotheses": [\\
        \{\\
            "causality": "A free-text mechanism description",\\
            "evidence\_features": ["feature1", ...]\\
        \},\\
        ...\\
    ]\\
\}
\end{predictionprompthypothesisgeneration}

\newtcolorbox[use counter=figure]{predictionpromptprincipleinduction}{
  colback=blue!4,
  colframe=black!70,
  boxrule=1.2pt,
  arc=3pt,
  left=10pt,
  right=10pt,
  top=10pt,
  bottom=10pt,
  width=\textwidth,
  fontupper=\ttfamily\fontsize{8}{9}\selectfont,
  fonttitle=\small,
  title={Figure~\thefigure: Prompt for Principle Induction on Molecular Property Prediction},
  label={app:fig:prompt_molecular_property_prediction_principle_induction},
  enhanced,
  breakable,
}

\begin{predictionpromptprincipleinduction}
You are summarizing a molecular similarity cluster.\\
\\
You will be given multiple molecules from the same cluster.\\
For each molecule, you will see:\\
- SMILES\\
- label\\
- the comparisons it was hypothesised from: a structurally similar reference\\
  molecule, that reference's label, whether the label CHANGED or stayed\\
  UNCHANGED between the two, and the hypothesis explaining the difference\\
\\
Your task is to extract a short, reusable, cluster-specific semantic memory.\\
\\
Internally do this before producing the final JSON:\\
1. Identify the structural features that recur across the SMILES.\\
2. Separate core scaffold features from peripheral substituent variation.\\
3. Check whether the hypotheses support those recurring structural features.\\
4. Contrast the CHANGED comparisons against the UNCHANGED ones: the substructures that\\
   recur in CHANGED comparisons are label-determining, the ones that recur in UNCHANGED\\
   comparisons are tolerated variation.\\
5. Remove any statement that is too generic and would fit many unrelated molecules for the same endpoint.\\
6. Keep only features that are local to this cluster.\\
\\
Priority order\\
1. Shared scaffold / chemotype\\
2. Substructures that flip the label across the CHANGED comparisons\\
3. Substructures that the UNCHANGED comparisons show to be tolerated\\
4. Recurring substituent classes\\
5. Label-specific interpretation\\
6. Generic mechanism language only if still necessary\\
\\
Definition of a good cluster memory\\
- Names the recurring scaffold shape or motif\\
- Explains what is conserved versus variable\\
- Is specific enough to distinguish this cluster from other positive clusters for the same task\\
\\
Definition of a bad cluster memory\\
- Mostly generic pharmacophore language\\
- Mostly endpoint-mechanism language\\
- Statements that ignore the actual recurring scaffold in the SMILES\\
\\
Return JSON with the following format:\\
\{\\
  "cluster\_summary": "...",\\
  "patterns": \{\\
    "<label\_1>": ["...", "..."],\\
    "<label\_2>": ["...", "..."]\\
  \},\\
  "contradictions": ["...", "..."]\\
\}\\
\\
Output constraints\\
- Mention the actual recurring scaffold in cluster\_summary.\\
- Use cautious frequency qualifiers.\\
- Mention variable regions if present.\\
- Do not include claims unsupported by the cluster examples.\\
- Do not write generic statements unless they are made cluster-specific.\\
- If all molecules in the cluster share the same label (i.e., there is label consensus), set "contradictions" to an empty list [].\\
\\
Cluster data:\\
\{cluster\_examples\}
\end{predictionpromptprincipleinduction}

\newtcolorbox[use counter=figure]{predictionprompthypothesisretrieval}{
  colback=blue!4,
  colframe=black!70,
  boxrule=1.2pt,
  arc=3pt,
  left=10pt,
  right=10pt,
  top=10pt,
  bottom=10pt,
  width=\textwidth,
  fontupper=\ttfamily\fontsize{8}{9}\selectfont,
  fonttitle=\small,
  title={Figure~\thefigure: Prompt for Hypothesis Retrieval on Molecular Property Prediction},
  label={app:fig:prompt_molecular_property_prediction_hypothesis_retrieval},
  enhanced,
  breakable,
}

\begin{predictionprompthypothesisretrieval}
You are selecting which past molecular comparisons are most useful for predicting a query molecule's label.\\
\\
Task: \{task\_description\}\\
\\
Query molecule: \{query\_smiles\}\\
\\
Each candidate below is a comparison between a target training molecule and a reference training molecule, both with known labels, together with the hypotheses that explain their structural difference.\\
\\
\{candidate\_block\}\\
\\
Select the 3 candidates whose structural difference is most informative for the query molecule.\\
\\
Selection criteria:\\
- Prefer comparisons whose differing substructure is actually present in, or absent from, the query molecule.\\
- Keep a mix: comparisons where the label changes show what flips the outcome, comparisons where it stays the same show what is tolerated.\\
- Prefer specific, mechanistically grounded hypotheses over generic ones.\\
\\
Return JSON only:\\
\{\\
    "selected\_ids": [<candidate id>, ...]\\
\}
\end{predictionprompthypothesisretrieval}

\newtcolorbox[use counter=figure]{predictionpromptprincipleretrieval}{
  colback=blue!4,
  colframe=black!70,
  boxrule=1.2pt,
  arc=3pt,
  left=10pt,
  right=10pt,
  top=10pt,
  bottom=10pt,
  width=\textwidth,
  fontupper=\ttfamily\fontsize{8}{9}\selectfont,
  fonttitle=\small,
  title={Figure~\thefigure: Prompt for Principle Retrieval on Molecular Property Prediction},
  label={app:fig:prompt_molecular_property_prediction_principle_retrieval},
  enhanced,
  breakable,
}

\begin{predictionpromptprincipleretrieval}
You are selecting the most relevant semantic memories for a molecular property prediction.\\
\\
Task: \{task\_description\}\\
\\
Query molecule: \{query\_smiles\}\\
\\
You will receive a list of candidate semantic memories. Each candidate is a cluster-level\\
summary of structural patterns recurring among structurally similar training molecules.\\
\\
Select the semantic memories that are relevant for predicting the query molecule's label.\\
\\
Selection priorities:\\
1. Clusters whose patterns name a reactive or alerting functional group that is present in the query molecule, even when the overall scaffold differs\\
2. Memories with concrete label-discriminative structural guidance\\
3. Diverse but complementary guidance when selecting multiple memories\\
4. Avoid redundant clusters with overlapping advice\\
\\
Return only valid JSON with this schema:\\
\{\\
  "indices": [0, 1]\\
\}\\
\\
Requirements:\\
- Indices must come only from the provided candidate list.\\
- Output JSON only.\\
\\
Candidate semantic memories:\\
\{candidate\_memories\_text\}
\end{predictionpromptprincipleretrieval}

\newtcolorbox[use counter=figure]{predictionprompthypothesisrevision}{
  colback=blue!4,
  colframe=black!70,
  boxrule=1.2pt,
  arc=3pt,
  left=10pt,
  right=10pt,
  top=10pt,
  bottom=10pt,
  width=\textwidth,
  fontupper=\ttfamily\fontsize{8}{9}\selectfont,
  fonttitle=\small,
  title={Figure~\thefigure: Prompt for Hypothesis Revision on Molecular Property Prediction},
  label={app:fig:prompt_molecular_property_prediction_hypothesis_revision},
  enhanced,
  breakable,
}

\begin{predictionprompthypothesisrevision}
You are refining causal hypotheses about a molecular structure comparison.\\
\\
During evaluation, this comparison was retrieved as evidence for predicting other molecules' labels.\\
The hypotheses below were originally inferred to explain the comparison.\\
Now, additional prediction outcomes (observations) have been collected from the same evaluation round.\\
Use these observations to refine the hypotheses.\\
\\
Task: \{task\_description\}\\
\\
=== Comparison Being Refined ===\\
Target molecule:\\
  SMILES: \{entry\_smiles\}\\
  Observed label: \{entry\_label\}\\
\\
Reference molecule:\\
  SMILES: \{reference\_smiles\}\\
  Observed label: \{reference\_label\}\\
\\
Label Change: \{label\_change\}\\
\\
=== Current Hypotheses for This Comparison ===\\
\{current\_hypotheses\_text\}\\
\\
=== Observations from Other Predictions ===\\
\{observations\_text\}\\
\\
Based on the observations above, please refine the current hypotheses for the target-reference comparison.\\
Consider the following:\\
- Which hypotheses are supported by the observed outcomes? Keep and strengthen them.\\
- Which hypotheses are contradicted by the outcomes? Revise or remove them.\\
- Are there any new patterns in the observations that suggest new hypotheses? Add them.\\
\\
Return the output as a valid JSON object with the following structure only:\\
\\
\{\\
    "hypotheses": [\\
        \{\\
            "causality": "A free-text mechanism description",\\
            "evidence\_features": ["feature1", ...]\\
        \},\\
        ...\\
    ]\\
\}\\
\\
Requirements:\\
- "causality" must be a concise explanation of the refined causal mechanism.\\
- "evidence\_features" must contain short phrases naming the key structural differences that support the hypothesis.\\
- Output only the JSON object and nothing else.
\end{predictionprompthypothesisrevision}

\newtcolorbox[use counter=figure]{predictionpromptprediction}{
  colback=blue!4,
  colframe=black!70,
  boxrule=1.2pt,
  arc=3pt,
  left=10pt,
  right=10pt,
  top=10pt,
  bottom=10pt,
  width=\textwidth,
  fontupper=\ttfamily\fontsize{8}{9}\selectfont,
  fonttitle=\small,
  title={Figure~\thefigure: Prompt for Molecular Property Prediction},
  label={app:fig:prompt_molecular_property_prediction_prediction},
  enhanced,
  breakable,
}

\begin{predictionpromptprediction}
\{task\_description\}\\
\\
=== Retrieved Knowledge ===\\
\\
\#\# Semantic Knowledge:\\
\{retrieved\_principles\}\\
\\
\#\# Structural Comparisons and Their Hypotheses:\\
\{retrieved\_comparisons\}\\
\\
=== End Retrieved Knowledge ===\\
\\
\{task\_question\}\\
Molecule: \{query\_smiles\}\\
\\
Answer format: \{answer\_format\}
\end{predictionpromptprediction}

\newtcolorbox[use counter=figure]{predictionprompttaskdescription}{
  colback=blue!4,
  colframe=black!70,
  boxrule=1.2pt,
  arc=3pt,
  left=10pt,
  right=10pt,
  top=10pt,
  bottom=10pt,
  width=\textwidth,
  fontupper=\ttfamily\fontsize{8}{9}\selectfont,
  fonttitle=\small,
  title={Figure~\thefigure: Task Property, Description, and Objective on Molecular Property Prediction},
  label={app:fig:prompt_molecular_property_prediction_task_description},
  enhanced,
  breakable,
}

\begin{predictionprompttaskdescription}
Task Property: DILI\\
Task Description:\\
You are a drug discovery assistant tasked with predicting the DILI (Drug-Induced Liver Injury) of a molecule. Specifically, your role is to determine whether a molecule is likely to cause liver injury or not.\\
Task Objective / Question:\\
Determine whether the given molecule can cause liver injury or not.\\
Observed Labels: Can cause liver injury; Does not cause liver injury\\
Answer Format:\\
Your final answer should be formatted as either: 'Final answer: Can cause liver injury' or 'Final answer: Does not cause liver injury.'\\
\\
Task Property: hERG\\
Task Description:\\
You are a drug discovery assistant tasked with predicting the hERG (Human ether-a-go-go related gene (hERG)) inhibition of a molecule. Specifically, your role is to determine whether a molecule blocks or does not block the hERG.\\
Task Objective / Question:\\
Determine the hERG inhibition of the given molecule.\\
Observed Labels: Block; Not block\\
Answer Format:\\
Your final answer should be formatted as either: 'Final answer: Block' or 'Final answer: Not block'\\
\\
Task Property: Ames\\
Task Description:\\
You are a drug discovery assistant tasked with predicting the Ames mutagenicity of a molecule. Specifically, your role is to determine whether a molecule is mutagenic or not mutagenic.\\
Task Objective / Question:\\
Determine whether the given molecule is mutagenic or not mutagenic.\\
Observed Labels: Mutagenic; Not mutagenic\\
Answer Format:\\
Your final answer should be formatted as either: 'Final answer: Mutagenic' or 'Final answer: Not mutagenic.'\\
\\
Task Property: Skin Reaction\\
Task Description:\\
You are a drug discovery assistant tasked with predicting the skin reaction of a molecule. Specifically, your role is to determine whether a molecule induces an immune reaction in inherently susceptible individuals that leads to skin sensitization.\\
Task Objective / Question:\\
Determine whether the given molecule induces an immune reaction in inherently susceptible individuals that leads to skin sensitization.\\
Observed Labels: Cause skin reaction; Does not cause skin reaction\\
Answer Format:\\
Your final answer should be formatted as either: 'Final answer: Cause skin reaction' or 'Final answer: Does not cause skin reaction.'\\
\\
Task Property: CYP1A2\\
Task Description:\\
You are a drug discovery assistant tasked with predicting the inhibition of the cytochrome P450 1A2 (CYP1A2) enzyme by a molecule. Specifically, your role is to determine whether a molecule inhibits CYP1A2 or does not inhibit CYP1A2.\\
Task Objective / Question:\\
Determine whether the given molecule inhibits CYP1A2.\\
Observed Labels: Does not inhibit CYP1A2; Inhibits CYP1A2\\
Answer Format:\\
Your final answer should be formatted as either: 'Final answer: Inhibits CYP1A2' or 'Final answer: Does not inhibit CYP1A2.'\\
\\
Task Property: CYP2C9\\
Task Description:\\
You are a drug discovery assistant tasked with predicting the inhibition of the cytochrome P450 2C9 (CYP2C9) enzyme by a molecule. Specifically, your role is to determine whether a molecule inhibits CYP2C9 or does not inhibit CYP2C9.\\
Task Objective / Question:\\
Determine whether the given molecule inhibits CYP2C9.\\
Observed Labels: Does not inhibit CYP2C9; Inhibits CYP2C9\\
Answer Format:\\
Your final answer should be formatted as either: 'Final answer: Inhibits CYP2C9' or 'Final answer: Does not inhibit CYP2C9.'\\
\\
Task Property: CYP2D6\\
Task Description:\\
You are a drug discovery assistant tasked with predicting the inhibition of the cytochrome P450 2D6 (CYP2D6) enzyme by a molecule. Specifically, your role is to determine whether a molecule inhibits CYP2D6 or does not inhibit CYP2D6.\\
Task Objective / Question:\\
Determine whether the given molecule inhibits CYP2D6.\\
Observed Labels: Does not inhibit CYP2D6; Inhibits CYP2D6\\
Answer Format:\\
Your final answer should be formatted as either: 'Final answer: Inhibits CYP2D6' or 'Final answer: Does not inhibit CYP2D6.'\\
\\
Task Property: CYP3A4\\
Task Description:\\
You are a drug discovery assistant tasked with predicting the inhibition of the cytochrome P450 3A4 (CYP3A4) enzyme by a molecule. Specifically, your role is to determine whether a molecule inhibits CYP3A4 or does not inhibit CYP3A4.\\
Task Objective / Question:\\
Determine whether the given molecule inhibits CYP3A4.\\
Observed Labels: Does not inhibit CYP3A4; Inhibits CYP3A4\\
Answer Format:\\
Your final answer should be formatted as either: 'Final answer: Inhibits CYP3A4' or 'Final answer: Does not inhibit CYP3A4.'
\end{predictionprompttaskdescription}

\end{document}